\pdfoutput=1

\documentclass[11pt]{article}

\usepackage[final]{acl}

\usepackage{times}
\usepackage{latexsym}
\usepackage{booktabs}
\usepackage[table]{xcolor}
\usepackage{tcolorbox}
\IfFileExists{balance.sty}{\usepackage{balance}}{}

\usepackage[T1]{fontenc}

\usepackage[utf8]{inputenc}

\usepackage{microtype}

\usepackage{inconsolata}

\usepackage{graphicx}
\usepackage{multirow}
\usepackage{amssymb}
\usepackage{amsmath}
\usepackage{pifont}
\usepackage{weiwAlgorithm}
\usepackage{xspace}
\usepackage{diagbox}

\usepackage{graphics}
\usepackage{graphicx}
\usepackage{xspace}
\usepackage{subcaption}
\usepackage{amsmath}
\usepackage{amsthm}
\usepackage{mathrsfs}
\usepackage{array}
\usepackage{textcomp}
\usepackage{weiwAlgorithm}
\usepackage{url}
\usepackage{diagbox}
\usepackage{color}
\usepackage{booktabs}
\usepackage{listings}
\usepackage{setspace}
\usepackage{multirow}
\usepackage{booktabs}    
\usepackage{tabularx}    
\usepackage{array}       
\usepackage{makecell}    
\usepackage{enumitem}
\usepackage{lineno}
\usepackage{float}    
\usepackage{placeins} 
\usepackage{stfloats} 
\usepackage{tabularx}
\usepackage{array}
\usepackage{tabularx}
\usepackage{array}
\usepackage{booktabs}   
\usepackage{xcolor}     
\usepackage{amsmath}    
\usepackage{amssymb}    
\usepackage{tabularx}
\usepackage{array}
\usepackage{booktabs}
\usepackage{subcaption}   
\usepackage{xcolor}
\usepackage{amsmath}

\usepackage{graphicx}

\newcommand{\benchmark}{BioMedHop\xspace}
\newcommand{\bianque}{BioWeave\xspace}

\newcommand{\myparagraph}[1]{\noindent \textbf{#1}.}

\newcommand{\myparagraphunderline}[1]{\noindent \underline{#1.}}
\newcommand{\myparagraphquestion}[1]{\noindent \textbf{#1?}}

\newtheorem{definition}{Definition}
\definecolor{qblue}{RGB}{30,90,180}
\definecolor{trgreen}{RGB}{0,140,60}
\definecolor{dmpurple}{RGB}{120,70,170}
\definecolor{wmorange}{RGB}{210,110,20}
\definecolor{agent}{RGB}{0,130,130}
\newcommand{\Badge}[2][black]{%
  \begingroup
  \setlength{\fboxsep}{1.5pt}%
  \fcolorbox{#1}{white}{\textcolor{#1}{\scriptsize\sffamily\bfseries #2}}%
  \endgroup
}
\newcommand{\QTool}[1]{\ensuremath{\mathop{\text{\Badge[qblue]{#1}}}\nolimits}}
\newcommand{\WMTool}[1]{\ensuremath{\mathop{\text{\Badge[wmorange]{#1}}}\nolimits}}
\newcommand{\TreeTool}[1]{\ensuremath{\mathop{\text{\Badge[trgreen]{#1}}}\nolimits}}
\newcommand{\AgentTool}[1]{\ensuremath{\mathop{\text{\Badge[agent]{#1}}}\nolimits}}
\newcommand{\DMTool}[1]{\ensuremath{\mathop{\text{\Badge[dmpurple]{#1}}}\nolimits}}
\newcommand{\PlainTool}[1]{\ensuremath{\mathop{\text{\textcolor{black}{\scriptsize\sffamily\bfseries #1}}}\nolimits}}
\newcommand{\ie}{{i.e.,}\xspace}

\title{Weaving Multi-Source Evidence for Biomedical Reasoning: The BioMedHop Benchmark and BioWeave Framework}

\author{
 \textbf{Xingyu Tan\textsuperscript{1,2}},
 \textbf{Shiyuan Liu\textsuperscript{3}},
 \textbf{Xiaoyang Wang\textsuperscript{1}},
 \textbf{Qing Liu\textsuperscript{2}},
\\
 \textbf{Xiwei Xu\textsuperscript{2}},
 \textbf{Xin Yuan\textsuperscript{2,1}}, 
 \textbf{Liming Zhu\textsuperscript{2}},
 \textbf{Wenjie Zhang\textsuperscript{1}}
\\
 \textsuperscript{1}University of New South Wales, Australia, 
 \textsuperscript{2}CSIRO, Australia\\
 \textsuperscript{3}University of Technology Sydney, Australia
\\
   \texttt{\{xingyu.tan, xiaoyang.wang1, wenjie.zhang\}@unsw.edu.au}\\
   \texttt{\{q.liu, xiwei.xu, xin.yuan, liming.zhu\}@csiro.au}\\
   \texttt{shiyuan.liu-1@student.uts.edu.au}
}

\begin{document}
\maketitle

\begin{abstract}
Biomedical question answering (QA) increasingly requires reasoning over interacting entities, where supporting evidence is scattered across biomedical knowledge graphs, literature documents, and web-accessible resources.
However, existing biomedical QA benchmarks mainly focus on exam-style knowledge, literature comprehension, or short-range multi-hop inference, leaving source-conditioned graph reasoning and evidence topology construction underexplored.
To fill this gap, we introduce \textbf{BioMedHop}, a multi-source graph-grounded benchmark for evaluating biomedical reasoning over structured evidence topologies.
BioMedHop contains 10,045 instances across KG, document, web, and hybrid evidence settings, covering shared-neighbor matching, intersection reasoning, path-based reasoning, and counting, with option-based, open-ended, and numeric count renderings.
To support this benchmark, we further propose \textbf{BioWeave}, a source-aware reasoning framework that retrieves biomedical KG paths, gathers supporting clues from documents and web sources, assembles them into a unified evidence graph, and verifies answers through entity-level evidence support.
Comprehensive experiments\footnote{The project page is available at \url{https://stevetantan.github.io/BioWeave/}.} show that BioWeave achieves the best overall performance among compared methods on BioMedHop, outperforming the strong hybrid baseline ToG-2 by 10.5\% in the overall average.
Moreover, BioWeave consistently improves different LLM backbones and enables smaller models, such as Qwen3-4B, to achieve reasoning performance comparable to GPT-4-Turbo.
\end{abstract}

\section{Introduction}
\label{introduction}

\begin{figure*}[t!]
    \centering
    \includegraphics[width=0.8\textwidth]{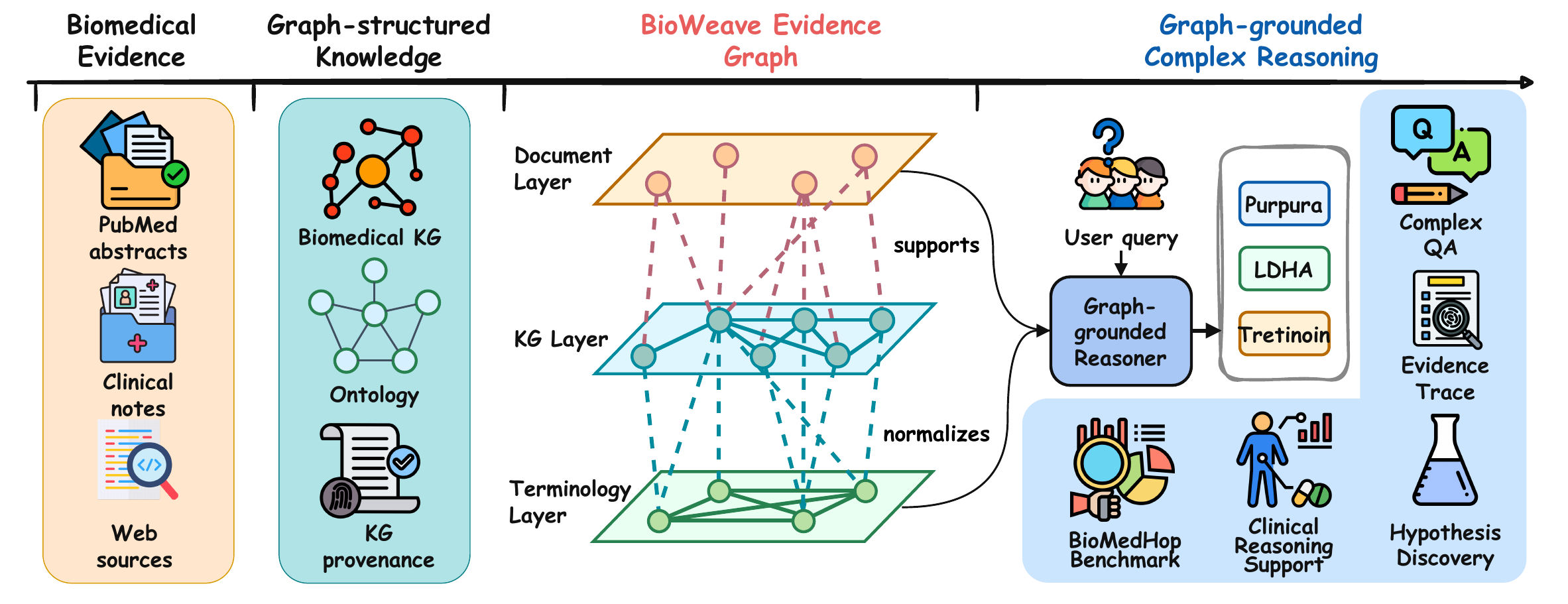}
\vspace{-3mm}
    \caption{Overview of \benchmark and \bianque. \benchmark controls KG, document, web, and hybrid evidence settings, while \bianque normalizes biomedical terminology and weaves KG paths, textual evidence, and optional web evidence into a unified evidence graph for complex biomedical reasoning.}
\vspace{-3mm}
    \label{fig:intro-bioweave}
\end{figure*}

Large language models (LLMs) have shown strong capability in clinical and biomedical question answering~\cite{singhal2023large,lievin2024can}.
However, complex biomedical reasoning remains challenging because the required evidence is specialized, fast-changing, and often distributed across heterogeneous sources.
A single question may require the model to connect genes, diseases, drugs, phenotypes, anatomical sites, cell types, and clinical outcomes, while also preserving the identity and relation of each biomedical concept.
This makes biomedical question answering (QA) not only a knowledge recall task, but also a source-conditioned reasoning problem.

Retrieval-augmented generation (RAG) is a natural way to ground LLMs with external evidence~\cite{lewis2020rag}.
However, in biomedical QA, retrieving relevant passages is not always sufficient.
The same biomedical concept can appear through abbreviations, synonyms, database identifiers, and ontology-specific names.
Resources such as UMLS and HPO are designed precisely to normalize these variants~\cite{bodenreider2004umls,robinson2008hpo}.
Moreover, many questions depend on typed relations, such as {gene--disease association} and {anatomy--cell-type localization}.
Text-only RAG therefore faces an entity-grounding limitation:
\textbf{L1: Surface-form retrieval cannot reliably preserve biomedical entity identity.}
It may retrieve locally relevant passages, but still fail to maintain a consistent entity-level reasoning chain.
Figure~\ref{fig:intro-bioweave} illustrates this need to connect documents, KG facts, and terminology-level aliases.

This limitation naturally motivates biomedical knowledge graphs (KGs), which explicitly represent biomedical entities and typed relations.
Biomedical KGs such as Monarch and SPOKE integrate genotype--phenotype, gene--disease, molecular, and clinical relations~\cite{mungall2017monarch,morris2023spoke}.
KG-based RAG can constrain reasoning to valid paths and make intermediate reasoning steps more explicit~\cite{tog1.0sun2023think,jiang2023structgpt,plan-on-graph}.
However, using the KG as the main evidence source shifts the bottleneck from lexical grounding to graph coverage.
Curated KGs may miss important edges, contain noisy links, encode ontology-specific semantics, and lag behind recent biomedical findings.
Gene--disease resources, e.g., DisGeNET, also show that biomedical evidence often combines expert curation, literature mining, and provenance scores~\cite{pinero2015disgenet}.
Thus, KG-RAG faces a coverage-and-freshness limitation:
\textbf{L2: Curated graph structure can guide reasoning, but cannot fully cover emerging biomedical evidence.}

Text and KGs therefore fail along complementary axes.
Text provides rich and updated evidence, but is weak at entity-level normalization and typed reasoning.
KGs provide structured relations and interpretable paths, but may be incomplete or outdated.
HybridRAG is a natural next step: KG paths can provide structure, while documents and web evidence can provide context, coverage, and freshness~\cite{li2023chaincok,sarmah2024hybridrag}.
However, simply retrieving from multiple sources is not the same as reasoning over aligned evidence.
Many hybrid pipelines still treat KG facts, document snippets, and web passages as separate evidence pools~\cite{li2023chaincok,tog2.0ma2024think}, leaving the generator to reconcile aliases, granularities, and provenance by itself.
This leads to an evidence-alignment limitation:
\textbf{L3: Multi-source retrieval does not by itself align heterogeneous biomedical evidence.}
For biomedical reasoning, evidence from KGs, documents, and the web must be connected, normalized, selected, and verified as part of the same reasoning chain.

These limitations also reveal an evaluation gap.
Existing biomedical QA benchmarks mainly focus on exam-style knowledge, literature comprehension, or limited multi-hop reasoning.
They rarely test whether a model can identify which source supports each reasoning step, align aliases across documents and KGs, verify KG paths against textual evidence, and answer both option-based and open-ended questions under the same graph-grounded reasoning structure.
Therefore, current evaluation is insufficient for diagnosing source-conditioned biomedical reasoning:
\textbf{L4: Existing biomedical QA benchmarks do not fully evaluate source-conditioned graph reasoning.}

\smallskip
\myparagraph{Contributions}
To fill the gap, we present \textbf{\benchmark} and \textbf{\bianque}: a source-conditioned biomedical reasoning benchmark and a framework for evidence topology construction and verification.
As shown in Figure~\ref{fig:intro-bioweave}, \benchmark evaluates whether models can reason over KG, document, web, and hybrid evidence settings, while \bianque connects heterogeneous biomedical evidence into a unified evidence graph for faithful multi-hop reasoning.
In summary, our contributions are as follows:

\noindent\raisebox{0.2ex}{\scriptsize$\bullet$}\enspace
\myparagraph{Source-conditioned biomedical benchmark}
To study complex biomedical reasoning beyond single-source QA, we construct \benchmark, a benchmark with 10,045 instances across KG, document, web, and hybrid evidence settings.
Each instance evaluates multi-hop and multi-entity reasoning under explicit evidence-source conditions, using option-based, open-ended, or numeric counting formats depending on the task family. 
This design enables systematic evaluation of a model's ability to identify, use, and combine evidence from different biomedical sources.

\noindent\raisebox{0.2ex}{\scriptsize$\bullet$}\enspace
\myparagraph{Evidence topology construction framework}
To address the limitation that KG facts and textual evidence are often retrieved as disconnected context, we propose \bianque, a retrieval and reasoning framework that represents biomedical KG paths, document evidence, and optional web evidence as a unified evidence graph.
\bianque uses KG paths to provide typed relational structure, while using documents and cautiously weighted web evidence to improve coverage and context.

\noindent\raisebox{0.2ex}{\scriptsize$\bullet$}\enspace
\myparagraph{Heterogeneous evidence weaving}
To align evidence across different biomedical sources, \bianque introduces unified evidence graph conversion, entity-level evidence alignment, cross-source evidence selection, and verification.
These modules normalize aliases, connect textual clues with KG relations, and verify whether each reasoning step is supported by source-compatible evidence.
This allows \bianque to build coherent reasoning chains rather than simply concatenating retrieved context.



\noindent\raisebox{0.2ex}{\scriptsize$\bullet$}\enspace
\myparagraph{Effectiveness and adaptability}
a) \bianque is a plug-and-play framework that can be applied to different LLMs, biomedical KGs, and text sources.
b) \bianque supports automatic evidence refresh.
It can incorporate new evidence through document and web retrieval without costly LLM fine-tuning.
c) \bianque achieves the best overall performance among compared methods on \benchmark, surpassing ToG-2 by 10.5\% and PoG by 20.5\% in Overall Avg., while enabling smaller models to achieve reasoning performance comparable to GPT-4-Turbo.

\section{Related Work}
\vspace{-1mm}
\label{relatedworks}
\myparagraph{RAG}
RAG grounds LLM outputs in external evidence and reduces unsupported generation~\cite{lewis2020rag,gao2023retrieval}.
In biomedical QA, however, evidence is distributed across papers, ontologies, terminologies, and biomedical KGs.
Text-only RAG and iterative retrieval-generation methods can improve evidence coverage~\cite{shao2023enhancing}, while hybrid RAG systems combine structured and unstructured evidence through source switching, document graphs, or KG-vector retrieval~\cite{sarmah2024hybridrag,tog2.0ma2024think}.
These methods remain limited when questions require entity normalization, relation typing, and reasoning over interacting biomedical entities.

\myparagraph{Biomedical KGs}
Biomedical KGs such as Hetionet, PrimeKG, Monarch, SPOKE, DisGeNET, UMLS, and HPO encode relations among diseases, genes, drugs, phenotypes, anatomy, and clinical concepts~\cite{himmelstein2017hetionet,chandak2023primekg,bodenreider2004umls}.
KG-based LLM reasoning methods retrieve triples or paths to constrain multi-hop inference~\cite{tog1.0sun2023think,plan-on-graph,debated-on-graph}.
These graphs provide interpretable structure but remain incomplete, noisy, and delayed relative to newly published evidence.

\myparagraph{Biomedical benchmarks}
Biomedical QA benchmarks such as PubMedQA, BioASQ-QA, MedQA, and MedMCQA mainly evaluate literature comprehension, medical exam knowledge, or answer selection~\cite{jin2019pubmedqa,tsatsaronis2015overview,nentidis2023bioasq,pal2022medmcqa}.
BioHopR evaluates multi-hop, multi-answer reasoning over PrimeKG~\cite{kim2025biohopr,chandak2023primekg}, while MedHopQA studies open-ended cross-document disease reasoning with ontology-grounded answer evaluation~\cite{islamaj2026medhopqa}.


More detailed related work is discussed in the Appendix~\ref{sec:detailed-related-work}.

\vspace{-1mm}
\section{Preliminary}
\vspace{-1mm}

\label{sec:prelim}
Consider a biomedical KG collection $\mathbb{G}=\{\mathcal{G}^{(s)}\}$, where each KG $\mathcal{G}^{(s)}=(\mathcal{E}^{(s)},\mathcal{R}^{(s)})$ contains entities, relations, and factual triples, \ie
$
\mathcal{G}^{(s)}=\{(e_h,r,e_t)\mid e_h,e_t\in\mathcal{E}^{(s)}, r\in\mathcal{R}^{(s)}\}.
$

\begin{definition}[Reasoning Path]
Given a KG $\mathcal{G}$, a reasoning path is a connected sequence of triples:
$
\operatorname{path}_{\mathcal{G}}(e_1,e_{l+1})
=\{(e_1,r_1,e_2),(e_2,r_2,e_3),\ldots,(e_l,r_l,e_{l+1})\},
$
where $l$ denotes the path length.
\end{definition}

\begin{definition}[Biomedical Evidence Source]
Given a question $q$ and topic entities $\operatorname{Topic}(q)$, we consider three evidence sources:
$
\mathbb{S}=\{\mathrm{KG},\mathrm{Doc},\mathrm{Web}\}.
$
KG evidence is $\mathcal{E}_{\mathrm{KG}}(q)=\{\mathcal{G}^{(s)}_q\mid \mathcal{G}^{(s)}\in\mathbb{G}\}$;\footnote{We use curated biomedical graph sources, including Monarch and SPOKE.}
document evidence is $\mathcal{E}_{\mathrm{Doc}}(q)=\operatorname{RetrieveMultiDoc}(\mathcal{I}_{doc},\operatorname{Topic}(q))$;\footnote{For each topic entity, we retrieve PubMed-centered literature using names, aliases, identifiers, and relation phrases.}
web evidence is $\mathcal{E}_{\mathrm{Web}}(q)=\operatorname{SearchWeb}(\mathcal{W},q)$.\footnote{Web retrieval uses clinical-trial records when applicable and Google Search API otherwise.}
\end{definition}

\begin{definition}[Source-conditioned Evidence]
Given a source condition $S_q\subseteq\mathbb{S}$, the evidence exposed for $q$ is
$
\mathcal{E}_{S_q}(q)=\bigcup_{s\in S_q}\mathcal{E}_{s}(q).
$
\end{definition}

\noindent \benchmark (Section \ref{benchmark}) controls $S_q$ during evaluation, while \bianque (Section \ref{methodology}) predicts and retrieves evidence from $S_q$ for answer generation.

\section{BioMedHop Benchmark}
\vspace{-1mm}
\label{benchmark}
\myparagraph{Overview}
\vspace{-1mm}
Most biomedical QA benchmarks isolate one evidence regime, such as exam-style medical knowledge, literature comprehension, KG reasoning, or multi-document synthesis.
\benchmark targets the missing setting of source-conditioned hybrid evidence reasoning, that is, each instance is grounded in typed biomedical entities and relations, while the available evidence channel is explicitly controlled over KG, document, web, and hybrid settings.
\benchmark varies along two axes: \textbf{source condition}, which specifies the exposed evidence channels, and \textbf{reasoning demand}, which covers shared-neighbor matching, clue intersection, deep path reasoning, and set/count aggregation.
It contains \textbf{10,045} source-conditioned instances with option-based, open-ended, and counting renderings, allowing the same latent reasoning structure to test answer recognition, direct generation, and evidence-grounded reasoning.
Table~\ref{tab:benchmark-comparison} compares \benchmark with representative biomedical QA benchmarks.
The key distinction is not only scale, but explicit control over evidence channels and multi-hop, multi-entity reasoning operations.

\begin{table*}[t]
\centering
\scriptsize
\renewcommand{\arraystretch}{1.15}
\setlength{\tabcolsep}{3.8pt}
\definecolor{benchsrc}{RGB}{0,82,170}
\definecolor{benchreas}{RGB}{0,128,75}
\definecolor{benchfmt}{RGB}{210,92,0}
\definecolor{benchmiss}{RGB}{175,175,175}
\newcommand{\HeavyBenchCheck}[1]{\textcolor{#1}{\large\ding{51}}}
\newcommand{\SrcYes}{\HeavyBenchCheck{benchsrc}}
\newcommand{\ReasYes}{\HeavyBenchCheck{benchreas}}
\newcommand{\FmtYes}{\HeavyBenchCheck{benchfmt}}
\newcommand{\BenchNo}{\textcolor{benchmiss}{--}}
\newcommand{\BenchGroup}[2]{\textcolor{#1}{\textbf{#2}}}
\caption{
Comparison with representative biomedical QA benchmarks.
\benchmark provides 10,045 source-conditioned cases over KG, document, web, and hybrid evidence for multi-hop, multi-entity QA.
}
\vspace{-2mm}
\label{tab:benchmark-comparison}
\resizebox{0.9\textwidth}{!}{%
\begin{tabular}{llcccc@{\hspace{9pt}}cc@{\hspace{9pt}}cc}
\toprule
\multirow{2}{*}{\textbf{Benchmark}} 
& \multirow{2}{*}{\textbf{Primary Scope}} 
& \multicolumn{4}{c}{\BenchGroup{benchsrc}{Evidence Source / Control}} 
& \multicolumn{2}{c}{\BenchGroup{benchreas}{Reasoning Requirement}} 
& \multicolumn{2}{c}{\BenchGroup{benchfmt}{Answer Format}} \\
\cmidrule(lr){3-6}
\cmidrule(lr){7-8}
\cmidrule(lr){9-10}
& 
& \textbf{KG} 
& \textbf{Docs} 
& \textbf{Web} 
& \textbf{Source-cond.} 
& \textbf{Multi-hop} 
& \textbf{Multi-entity} 
& \textbf{MCQ} 
& \textbf{Open} \\
\midrule
PubMedQA~\cite{jin2019pubmedqa} 
& Biomedical research QA 
& \BenchNo & \BenchNo & \BenchNo & \BenchNo 
& \BenchNo & \BenchNo 
& \BenchNo & \FmtYes \\

BioASQ-QA~\cite{tsatsaronis2015overview,nentidis2023bioasq} 
& Biomedical expert QA 
& \BenchNo & \SrcYes & \BenchNo & \BenchNo 
& \BenchNo & \BenchNo 
& \BenchNo & \FmtYes \\

MedQA~\cite{jin2021disease} 
& Medical exam QA 
& \BenchNo & \BenchNo & \BenchNo & \BenchNo 
& \BenchNo & \BenchNo 
& \FmtYes & \BenchNo \\

MedMCQA~\cite{pal2022medmcqa} 
& Medical exam QA 
& \BenchNo & \BenchNo & \BenchNo & \BenchNo 
& \BenchNo & \BenchNo 
& \FmtYes & \BenchNo \\

BioHopR~\cite{kim2025biohopr} 
& KG-grounded biomedical QA 
& \SrcYes & \BenchNo & \BenchNo & \BenchNo 
& \ReasYes & \ReasYes 
& \BenchNo & \FmtYes \\

MedHopQA~\cite{islamaj2026medhopqa} 
& Multi-document biomedical QA 
& \BenchNo & \SrcYes & \BenchNo & \BenchNo 
& \ReasYes & \ReasYes 
& \BenchNo & \FmtYes \\

\midrule
\textbf{\benchmark~(Ours)} 
& \textbf{Multi-source biomedical reasoning} 
& \SrcYes & \SrcYes & \SrcYes & \SrcYes 
& \ReasYes & \ReasYes 
& \FmtYes & \FmtYes \\
\bottomrule
\end{tabular}
}
\vspace{-2mm}
\end{table*}

\vspace{-1mm}
\subsection{Benchmark Construction}
\label{sec:benchmark-construction}
\vspace{-1mm}

\benchmark is constructed as a source-conditioned reasoning benchmark rather than a collection of independent question templates.
Formally, each instance is represented as
\[
x=(q, a, \mathcal{G}^{gold}_q, S_q, \mathcal{E}_{S_q}, o),
\]
where $q$ is the rendered question, $a$ is the gold answer, $\mathcal{G}^{gold}_q$ is the answer-determining biomedical graph pattern, $S_q\subseteq\{\mathrm{KG},\mathrm{Doc},\mathrm{Web}\}$ is the active source condition, $\mathcal{E}_{S_q}$ is the evidence exposed under that condition, and $o$ denotes the reasoning operator.
This representation separates the latent reasoning structure from the surface evidence channel: changing the source condition does not change the gold graph or answer, but changes what evidence the model can use to recover it.

Algorithm~\ref{alg:benchmark_construction} of Appendix \ref{appendix:alg} summarizes the construction controller.
The pipeline has three stages.
First, \textbf{graph-pattern instantiation} uses \TreeTool{InstantiateGraphPattern} to sample typed biomedical neighborhoods and metapaths from curated KGs, with \DMTool{ValidateGraphAnswer} enforcing entity-type compatibility, relation constraints, answer uniqueness for single-answer tasks, and closed answer sets for counting tasks.
Second, \textbf{source-conditioned evidence alignment} uses \WMTool{BuildHopQueries} to convert each graph hop into entity, alias, identifier, and relation-phrase queries, then retrieves multi-document and web evidence through \WMTool{RetrieveMultiDoc} and \AgentTool{SearchWeb}.
The retrieved snippets are normalized to canonical KG entities before \DMTool{ComposeSourceEvidence} assembles the active source condition.
Third, \textbf{question rendering} uses \QTool{RenderMCQ} and \QTool{RenderOpenQuestion} for the first three task families, and \QTool{RenderCountQuestion} for Path-based Counting.
Distractors are type-compatible and lexically close to the answer, while open-ended evaluation uses normalized aliases and identifiers.
This pipeline gives \benchmark a controlled way to test whether models can recover the same biomedical reasoning structure from structured graph evidence, textual evidence, web evidence, or hybrid evidence.
This design trades some ecological breadth for diagnostic control: the latent graph is fixed while the available evidence topology varies, making source effects easier to isolate.
The question rewriting prompt used is provided in Appendix~\ref{prompt:question-rewriting}.

\definecolor{bhtaskblue}{RGB}{45,108,176}
\definecolor{bhtaskgreen}{RGB}{48,145,109}
\definecolor{bhtaskorange}{RGB}{214,126,49}
\definecolor{bhtaskpurple}{RGB}{130,91,180}
\newcommand{\BioHopBar}[2]{\textcolor{#1}{\rule{#2}{1.05ex}}}
\begin{table}[t]
\centering
\scriptsize
\renewcommand{\arraystretch}{1.18}
\caption{Task taxonomy of \benchmark.}
\vspace{-2mm}
\label{tab:benchmark-taxonomy}
\resizebox{\columnwidth}{!}{%
\begin{tabular}{@{}llcl@{}}
\toprule
\textbf{Task Family} & \textbf{Evidence Motif} & \textbf{Output} & \textbf{Instances} \\
\midrule
\BioHopBar{bhtaskblue}{1.1ex} Entity Pair Matching & 2-edge shared neighbor & MCQ/Open & \BioHopBar{bhtaskblue}{10.4mm}~2,142 (21.3\%) \\
\BioHopBar{bhtaskgreen}{1.1ex} Intersection Reasoning & 3-edge intersection graph & MCQ/Open & \BioHopBar{bhtaskgreen}{18.0mm}~3,682 (36.7\%) \\
\BioHopBar{bhtaskorange}{1.1ex} Path-based Reasoning & 4-hop typed metapath & MCQ/Open & \BioHopBar{bhtaskorange}{12.9mm}~2,653 (26.4\%) \\
\BioHopBar{bhtaskpurple}{1.1ex} Path-based Counting & 4-hop set aggregation & Count & \BioHopBar{bhtaskpurple}{7.7mm}~1,568 (15.6\%) \\
\bottomrule
\end{tabular}
}
\end{table}

\begin{table}[t]
\centering
\scriptsize
\renewcommand{\arraystretch}{1.18}
\vspace{-2mm}
\caption{Answer-centered evidence-graph distribution in \benchmark.}
\vspace{-2mm}
\label{tab:structure-distribution}
\resizebox{\columnwidth}{!}{%
\begin{tabular}{@{}llcl@{}}
\toprule
\textbf{Evidence Graph} & \textbf{Reasoning Focus} & \textbf{Count} & \textbf{Share} \\
\midrule
\BioHopBar{bhtaskgreen}{1.1ex} 3-edge intersection graph & three answer-clue edges & 3,682 & \BioHopBar{bhtaskgreen}{19.2mm}~36.7\% \\
\BioHopBar{bhtaskblue}{1.1ex} 2-edge shared neighbor & two anchor-answer edges & 2,142 & \BioHopBar{bhtaskblue}{11.2mm}~21.3\% \\
\BioHopBar{bhtaskorange}{1.1ex} 4-hop chain & metapath / count aggregation & 4,221 & \BioHopBar{bhtaskorange}{22.0mm}~42.0\% \\
\bottomrule
\end{tabular}
}
\vspace{-5mm}
\end{table}

\vspace{-1mm}
\subsection{Benchmark Taxonomy}
\vspace{-1mm}

\benchmark separates reasoning structure from source condition.
Each question first defines a graph-grounded reasoning target, while source condition controls which evidence channels are exposed or required during evaluation.
Table~\ref{tab:benchmark-taxonomy} summarizes the four task families in the \textbf{10,045} questions.
Table~\ref{tab:structure-distribution} further summarizes the answer-centered evidence-graph composition.
Notably, 42.0\% of \benchmark requires four-hop path reasoning or set/count aggregation, so the benchmark is not dominated by local entity matching.


The first two task families evaluate shallow but non-trivial multi-hop reasoning.
\textbf{Entity Pair Matching} asks the model to identify the biomedical node connecting two query entities.
For example, two diseases may share a gene, symptom, or anatomical entity.
We construct these questions from Monarch relation families and sample type-compatible distractors with similar lexical length, so the task cannot be solved by answer-type mismatch or obvious surface cues.
\textbf{Intersection Reasoning} instead starts from disease-centered neighborhoods.
The model receives typed clues, such as a gene, symptom, compound, organism, protein, or location, and must recover the disease connected to all of them.
This evaluates conjunctive entity aggregation rather than single-hop lookup.

The latter two task families evaluate deeper metapath reasoning over heterogeneous biomedical relations.
\textbf{Path-based Reasoning} instantiates typed metapaths from SPOKE, such as \textit{Anatomy $\rightarrow$ Gene $\rightarrow$ Disease $\rightarrow$ Compound $\rightarrow$ Blend}, and hides the disease node.
The question states endpoints and relation descriptions, requiring the model to traverse the path and preserve relation direction.
\textbf{Path-based Counting} turns the same path structure into set-valued reasoning: the model must count all distinct diseases satisfying the typed pattern.
This is intentionally harder than selecting one plausible entity because aliases, duplicate mentions, and multiple valid candidates must be resolved before producing a number, stressing ontology-grounded aggregation rather than semantic plausibility alone.
Detailed construction rules, distributions, templates, and examples are provided in Appendix~\ref{sec:appendix}.

\vspace{-1mm}
\subsection{Source-conditioned Evidence}
\vspace{-1mm}
For each item, the KG path or neighborhood defines the gold reasoning structure, while documents and web snippets provide natural-language evidence aligned with entities, aliases, and typed relation phrases.
The source condition changes only the evidence exposed to the model, not the underlying answer.
This enables controlled comparison among prompt-only, KG-only, document-only, web-only, and hybrid settings, and tests whether models can align textual clues with graph-grounded biomedical entities rather than merely retrieve fluent context.


\myparagraph{Hop-aware retrieval}
For each graph path $e_1 \stackrel{r_1}{\rightarrow} e_2 \cdots \stackrel{r_k}{\rightarrow} e_{k+1}$, we use \WMTool{BuildHopQueries} to construct both hop-level queries $(e_i,r_i,e_{i+1})$ and partial-chain queries $(e_i,r_i,\ldots,e_j)$.
This design allows evidence to be distributed across multiple documents or web snippets: one source may support a gene--disease association, another may support a contraindication relation, and another may provide phenotype or anatomical context.
As a result, the benchmark evaluates multi-document and multi-source synthesis rather than single-passage lookup.

\myparagraph{Entity grounding and alias control}
For entity grounding, \bianque applies \AgentTool{NormalizeMentions} to link retrieved evidence back to KG entities through exact names, aliases, identifiers, and type-compatible mentions.
Since biomedical concepts may appear as abbreviations, ontology labels, database identifiers, or clinical synonyms, the evidence layer records both the textual mention and the normalized entity.
This allows evaluation to check whether a model connects surface-form evidence to graph-grounded biomedical entities.

\myparagraph{Source-aware provenance}
For hybrid evidence settings, \benchmark records provenance at the hop level.
\WMTool{StoreMetadata} links evidence snippets to retrieval queries, source type, supporting hop, normalized entities, relation phrases, and provenance scores when available.
This lets evaluation distinguish answer correctness from faithful evidence use: a model should recover the right answer and support it with evidence consistent with the graph-defined reasoning structure.
The current metadata is source-type and hop aware; finer evidence-quality hierarchies, such as primary study versus review article versus weakly mined association, are left to future extensions.
Appendix~\ref{appendix:benchmark-source-evidence} provides additional implementation details.

\vspace{-2mm}
\subsection{Evaluation Metrics}
\vspace{-2mm}
We evaluate \benchmark using option accuracy, alias-aware open-answer accuracy, and CountEx for counting questions.
Automatic metrics are the primary signal, while an LLM verifier separately reports semantic answer correctness, evidence support, and reasoning completeness.
Detailed metric definitions, verifier schema, and judge protocols are provided in Appendix~\ref{appendix:benchmark-metrics}.

\section{\bianque}
\vspace{-1mm}

\label{methodology}
\myparagraph{Overview}
\bianque treats hybrid biomedical RAG as evidence topology construction rather than context concatenation.
The overview of \bianque is presented in Figure \ref{fig:method} of Appendix \ref{appendix:workflow_diagram}.
Given $q$ and the source space in Section~\ref{sec:prelim}, it builds a unified evidence graph $\mathcal{E}_q=(\mathcal{V}_q,\mathcal{R}_q)$ over normalized biomedical entities, KG paths, text-derived evidence units, and provenance links.
It proceeds through question initialization, multi-source evidence exploration, evidence pruning and verification, and verified answer generation.
Algorithm~\ref{alg:bioweave_query} of Appendix \ref{appendix:alg} summarizes this controller, and Appendix~\ref{appendix:prompts} lists the prompt templates.



\vspace{-1mm}
\subsection{Question Initialization}
\label{sec:question-initialization}
\vspace{-1mm}

Given a question $q$, \QTool{AbstractQuestion} first maps it to structured retrieval controls, including biomedical mentions, candidate entities $\mathcal{C}_q$, the expected answer type, and the reasoning operator $o(q)$, such as shared-neighbor matching, intersection reasoning, chain traversal, aggregation, or counting.
\AgentTool{InitSourcePlan} then selects evidence sources $S_q \subseteq \{\mathrm{KG}, \mathrm{Doc}, \mathrm{Web}\}$, restricted by the active benchmark source condition.

\vspace{-1mm}

\subsection{Multi-source Evidence Exploration}
\vspace{-1mm}

\noindent \bianque explores selected sources in parallel, handling structured KG evidence and unstructured text/web evidence separately.
Structured retrieval extracts explicit KG traces, while document and web retrieval recover implicit or recently updated evidence.
Algorithms~\ref{alg:structured_retrieval}--\ref{alg:bioweave_explore} give the pseudo-code.

\myparagraph{Structured retrieval}
\bianque retrieves KG evidence through two phases.

\myparagraphunderline{Evidence subgraph detection}
Given candidate entities $\mathcal{C}_{q}$, operator $o(q)$, missing-hop set $\mathcal{H}_{miss}$ (uncovered or uncertain graph hops), and maximum depth $D_{\max}$, \TreeTool{DetectEvidenceSubgraph} constructs a biomedical evidence subgraph $\mathcal{G}_{q}$ from $\mathbb{G}$.
\TreeTool{PruneSubgraph} then removes relation types and nodes that violate entity-type compatibility or are too frequent to be discriminative.

\myparagraphunderline{Operator-guided path retrieval}
Starting from $\mathcal{G}_{q}$, \TreeTool{ExpandTypedPaths} explores typed paths for the predicted operator, including shared-neighbor matching, intersection, chain traversal, aggregation, and counting.
\TreeTool{RankPrunePaths} ranks candidate paths by entity coverage, relation compatibility, missing-hop coverage, and operator consistency, retaining the top-$W_1$ paths as $\mathcal{P}_{kg}$, where $W_1$ is the first-stage retrieval budget.
These paths act as constrained reasoning traces with source KG, relation labels, and provenance.

\myparagraph{Unstructured retrieval}
For document evidence, 
\bianque uses \WMTool{BuildHopQueries} to construct hop-aware queries from $q$, normalized entity names, aliases, relation phrases, retrieved KG paths, and $\mathcal{H}_{miss}$.
It then calls \WMTool{RetrieveMultiDoc} over the PubMed-centered index $\mathcal{I}_{doc}$, followed by \WMTool{EntityAwareFilter} to keep spans with required entities and relation cues under budget $W_{doc}$.

\myparagraphunderline{Web document retrieval}
When $\mathrm{Web}\in S_q$, \AgentTool{BuildTargetedQueries} creates targeted searches for missing entities, uncertain hops, clinical-trial records, drug labels, recent guideline relations, or other freshness-sensitive evidence.
\AgentTool{SearchWeb} queries the web interface $\mathcal{W}$, after which \WMTool{EntityAwareFilter} keeps source-conditioned spans under budget $W_{web}$.
Because web evidence is less stable than curated KG or indexed literature, \DMTool{AssignLowerSourcePrior} marks it as lower-prior evidence before verification.

\myparagraph{Unified evidence graph conversion}
For text evidence, \bianque uses \AgentTool{ConvertEvidenceGraph} to convert selected document and web spans into graph-compatible candidate units.
Each unit contains normalized entities, relation predicates, hop labels, original text spans, and provenance.
Then, \WMTool{UpdateEvidenceMemory} stores KG paths and converted text units in a shared evidence memory.
Importantly, this conversion does not assert new KG facts; it only makes text evidence comparable with KG traces before verification.
The unified candidate pool is:
$
\mathcal{U}_q=\mathcal{P}_{kg}\cup\mathcal{U}_{doc}\cup\mathcal{U}_{web}.
$
\begin{table*}[!t]
  \centering
  \scriptsize
  \renewcommand{\arraystretch}{1.12}
  \caption{Main automatic evaluation on \benchmark with GPT-4-Turbo. Entity-answer tasks report MCQ accuracy, open-answer accuracy, and their macro average; Path-based Counting reports CountEx. Overall Avg. summarizes format-specific accuracy scores. Best results are in bold and second-best results are underlined.}
  \label{tab:main-results}
\vspace{-2mm}
  
  \resizebox{0.9\textwidth}{!}{%
  \begin{tabular}{@{}llccccccccccc@{}}
    \toprule
    \multirow{2}{*}{\textbf{Type}} & \multirow{2}{*}{\textbf{Method}}
    & \multicolumn{3}{c}{\textbf{Entity Pair}} & \multicolumn{3}{c}{\textbf{Intersection}} & \multicolumn{3}{c}{\textbf{Path Reason.}} & \multicolumn{1}{c}{\textbf{Path Counting}} & \multirow{2}{*}{\textbf{Overall Avg.}} \\
    \cmidrule(lr){3-5}\cmidrule(lr){6-8}\cmidrule(lr){9-11}\cmidrule(lr){12-12}
    & & \textbf{MCQ} & \textbf{Open} & \textbf{Avg.} & \textbf{MCQ} & \textbf{Open} & \textbf{Avg.} & \textbf{MCQ} & \textbf{Open} & \textbf{Avg.} & \textbf{CountEx} & \\
    \midrule
    \multirow{2}{*}{LLM-only}
    & IO & 23.0 & 2.4 & 12.7 & 34.0 & 9.3 & 21.6 & 26.7 & 25.5 & 26.1 & 12.5 & 18.2 \\
    & CoT~\cite{wei2022cot} & 31.0 & 2.4 & 16.7 & 41.0 & 7.0 & 24.0 & 33.3 & 21.8 & 27.5 & 15.2 & 20.9 \\
    \midrule
    \multirow{2}{*}{Vanilla}
    & Naive Doc~\cite{lewis2020rag} & 57.6 & 2.4 & 30.0 & 63.2 & 14.0 & 38.6 & 37.8 & 29.1 & 33.5 & 18.4 & 30.1 \\
    & Naive Web~\cite{lewis2020rag} & 61.0 & 0.0 & 30.5 & 57.9 & 9.3 & 33.6 & 37.8 & 32.7 & 35.2 & 17.5 & 29.2 \\
    \midrule
    \multirow{2}{*}{KG-based RAG}
    & ToG~\cite{tog1.0sun2023think} & 68.5 & 8.4 & 38.5 & 52.4 & 4.2 & 28.3 & 48.5 & 49.2 & 48.9 & 18.5 & 33.5 \\
    & PoG~\cite{plan-on-graph} & \underline{76.5} & 12.3 & 44.4 & 61.2 & 5.5 & 33.4 & \underline{65.4} & 46.8 & 56.1 & 23.4 & 39.3 \\
    \midrule
    \multirow{2}{*}{Hybrid RAG}
    & CoK~\cite{li2023chaincok} & 65.4 & 25.1 & 45.2 & 65.8 & 18.2 & 42.0 & 55.4 & \underline{64.5} & 60.0 & 21.2 & 42.1 \\
    & ToG-2~\cite{tog2.0ma2024think} & 74.2 & \underline{38.5} & \underline{56.4} & \underline{69.5} & \underline{22.4} & \underline{46.0} & 63.1 & 61.8 & \underline{62.5} & \underline{32.5} & \underline{49.3} \\
    \midrule
    Proposed & \textbf{\bianque} & \textbf{80.4} & \textbf{50.6} & \textbf{65.5} & \textbf{80.2} & \textbf{34.6} & \textbf{57.4} & \textbf{73.2} & \textbf{68.6} & \textbf{70.9} & \textbf{45.4} & \textbf{59.8} \\
    \bottomrule
  \end{tabular}
  }
  \vspace{-2mm}
\end{table*}

\vspace{-1mm}

\subsection{Evidence Pruning and Verification}
\vspace{-1mm}

\label{sec:method:pruning}

\myparagraph{Verification-guided pruning}
\bianque prunes heterogeneous evidence with both relevance and source-aware verification, since KG paths, PubMed snippets, and web evidence differ in reliability, granularity, and entity-normalization quality.
The verification term up-weights units grounded to normalized biomedical entities and supported by reliable or cross-source evidence, while down-weighting isolated low-prior claims.
In Algorithm~\ref{alg:bioweave_query}, this stage is invoked by \DMTool{CrossVerify}; Algorithm~\ref{alg:bioweave_prune} details the internal scoring and selection.

Formally, let $\mathcal{U}_q=\{u_i\}_{i=1}^{N}$ be the candidate evidence units for question $q$, where each unit has source type $\operatorname{src}(u_i)\in\{\mathrm{KG},\mathrm{Doc},\mathrm{Web}\}$. 
Given linked topic entities $\operatorname{Topic}(q)$, we compute the relevance score as:
\vspace{-1mm}
\[
\begin{aligned}
s^{\mathrm{rel}}_i
&=
\lambda_{\mathrm{sem}}\cos(\mathbf{h}(q),\mathbf{h}(u_i))\\
&\quad+
\lambda_{\mathrm{ent}}\mathrm{Jaccard}(\operatorname{Topic}(q),\operatorname{Ent}(u_i)),
\end{aligned}
\]
\vspace{-1mm}

\noindent where $\lambda_{\mathrm{sem}}+\lambda_{\mathrm{ent}}=1$, $\mathbf{h}(\cdot)$ is implemented with SentenceBERT~\cite{reimers-gurevych-2019-sentence}, and $\operatorname{Ent}(u_i)$ returns normalized biomedical entities in $u_i$. 
The top-$W_1$ units form a preliminary pool $\widetilde{\mathcal{U}}_q$ for the following source-aware verification.

\myparagraph{Source-aware verification}
For each $u_i\in\widetilde{\mathcal{U}}_q$, we estimate its reliability using three features: source prior, cross-source support, and KG grounding. 
The supporting source set is:
$
\operatorname{Supp}(u_i)=
\{\operatorname{src}(u_j)\mid u_j\in\widetilde{\mathcal{U}}_q,\ j\neq i,\ \operatorname{Sim}(u_i,u_j)\ge\gamma\},
$
where $\gamma$ is a similarity threshold. 
The three verification features are defined as:
\vspace{-2mm}
\[
\begin{aligned}
r_i^{\mathrm{prior}}
&=
\rho_{\operatorname{src}(u_i)},\,\,\,\,\,\, \,
r_i^{\mathrm{sup}}
=
\frac{\min(|\operatorname{Supp}(u_i)|, W_{\mathrm{src}})}{W_{\mathrm{src}}},\\
\end{aligned}
\]
\vspace{-3mm}
\[
\begin{aligned}
r_i^{\mathrm{kg}}
&=
\frac{
|\operatorname{Ent}(u_i)\cap V(\mathcal{G}_q)|
}{
\max\{1,|\operatorname{Ent}(u_i)|\}
}.
\end{aligned}
\]
\vspace{-3mm}

\noindent The verification score is computed as:
$
s_i^{\mathrm{ver}}
=
\sum_{m\in\{\mathrm{prior},\mathrm{sup},\mathrm{kg}\}}\alpha_m r_i^m,
$
where $\alpha_m\ge 0$ and $\sum_m\alpha_m=1$.

\myparagraph{Reasoning-aware selection}
Each candidate unit is then ranked by a cross-score:
$
s_i^{\mathrm{cross}}
=
\beta s_i^{\mathrm{rel}}
+
(1-\beta)s_i^{\mathrm{ver}}.
$
The top-$W_2$ units, where $W_2$ is the verification-stage budget, are passed to an LLM-aware selector, which checks source-condition satisfaction, answer-type compatibility, relation consistency, and support for the required reasoning operator. 
Selected units are finally compacted by \DMTool{RefineEvidenceGraph}, which collapses aliases, deduplicates entities, and preserves provenance for verified answer generation.

\vspace{-1mm}
\subsection{Verified Answer Generation}
\vspace{-1mm}

In the final stage, \QTool{GenFinalAnswer} uses the refined evidence graph as the basis for prediction.
For MCQ tasks, \bianque selects the option that satisfies both graph constraints and textual support; for open questions, it returns the normalized entity or entity set; and for counting, it counts distinct answer entities after alias resolution.
If \DMTool{SatisfyOperator} fails on the required relation pattern, \DMTool{BacktraceMissingHop} identifies the uncertain hop and reissues KG, document, or web retrieval for the missing link.

\vspace{-3mm}
\section{Experiments}
\vspace{-3mm}

\label{experiment}


In this section, we evaluate \bianque on \benchmark. The detailed experimental settings, including datasets, baselines, and implementations, can be found in Appendix \ref{sec:experiment-details}.

\vspace{-1mm}
\subsection{Main Results}
\vspace{-2mm}
\label{sec:main_results}

Since BioWeave leverages external biomedical knowledge, we first compare it against other RAG-based methods. As shown in Table~\ref{tab:main-results}, BioWeave achieves the best overall performance among compared methods across all task families, outperforming the strongest baseline by 10.5\% in Overall Avg. Compared with ToG-2, a strong hybrid RAG baseline, BioWeave improves Entity Pair, Intersection, and Path Reasoning by 9.1\%, 11.4\%, and 8.4\%, respectively. It also improves Path Counting by 12.9\% in CountEx, showing stronger capability in recovering complete biomedical evidence paths.
Compared with KG-based RAG methods, BioWeave improves the Overall Avg. over PoG by 20.5\%, indicating that KG paths alone are insufficient when biomedical evidence is incomplete or distributed across heterogeneous sources. Additionally, compared to vanilla document/web-based RAG methods, BioWeave achieves average gains of 30.2\%. Notably, vanilla RAG remains weak on open-answer questions, suggesting simply adding text context cannot reliably preserve biomedical entity identity and relation-level evidence.

When compared to methods without external knowledge (IO, CoT), BioWeave improves Overall Avg. by an average of 40.3\%.
These findings demonstrate that BioWeave is effective for complex biomedical reasoning. By weaving KG paths, document evidence, and web evidence into a unified source-conditioned evidence graph, BioWeave enhances the reasoning capability of LLMs and achieves superior performance on BioMedHop.

\begin{figure*}[t]
  \centering
  \includegraphics[width=0.9\textwidth]{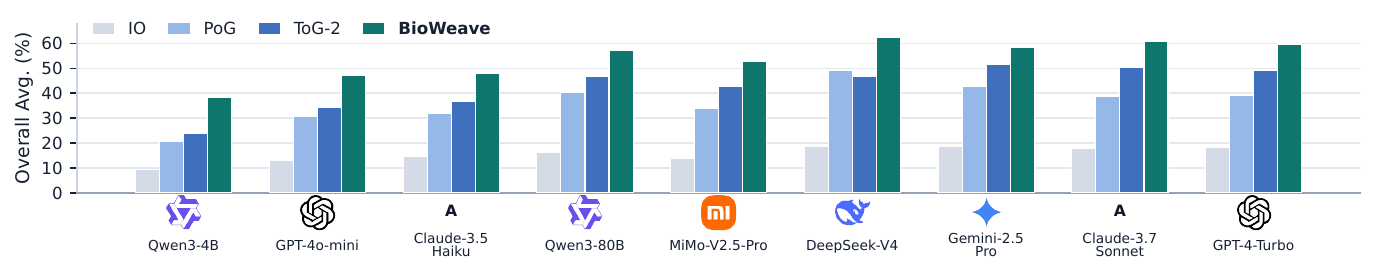}
\vspace{-4mm}
  \caption{Backbone robustness on \benchmark. Bars show Overall Avg. scores for IO, PoG, ToG-2, and \bianque across nine LLM backbones. Table~\ref{tab:full-backbone-sweep} reports the full task-family score table.}
\vspace{-2mm}
  \label{fig:backbone-robustness}
\end{figure*}
\vspace{-4mm}
\begin{table*}[!t]
  \centering
  \scriptsize
  \setlength{\tabcolsep}{2.4pt}
  \renewcommand{\arraystretch}{0.92}
  \caption{LLM-verifier effectiveness on \benchmark. Ans., Evid., and Reason. denote semantic answer correctness, evidence support, and reasoning completeness; best results are in bold and second-best results are underlined.}
\vspace{-2mm}
  \label{tab:main-llm-verifier}
  \resizebox{0.8\textwidth}{!}{%
  \begin{tabular}{@{}llcccccccccccc@{}}
    \toprule
    \multirow{2}{*}{\textbf{Type}} & \multirow{2}{*}{\textbf{Method}}
    & \multicolumn{3}{c}{\textbf{Entity Pair}} & \multicolumn{3}{c}{\textbf{Intersection}} & \multicolumn{3}{c}{\textbf{Path Reason.}} & \multicolumn{3}{c}{\textbf{Path Counting}} \\
    \cmidrule(lr){3-5}\cmidrule(lr){6-8}\cmidrule(lr){9-11}\cmidrule(lr){12-14}
    & & \textbf{Ans.} & \textbf{Evid.} & \textbf{Reason.} & \textbf{Ans.} & \textbf{Evid.} & \textbf{Reason.} & \textbf{Ans.} & \textbf{Evid.} & \textbf{Reason.} & \textbf{Ans.} & \textbf{Evid.} & \textbf{Reason.} \\
    \midrule
    \multirow{2}{*}{LLM-only}
    & IO & 13.3 & 58.0 & 42.0 & 23.3 & 52.0 & 45.0 & 28.0 & 40.0 & 32.0 & 14.5 & 16.0 & 12.0 \\
    & CoT~\cite{wei2022cot} & 17.8 & 56.0 & 41.0 & 25.2 & 55.0 & 48.0 & 29.5 & 36.0 & 28.0 & 17.0 & 18.0 & 18.0 \\
    \midrule
    \multirow{2}{*}{KG-based RAG}
    & ToG~\cite{tog1.0sun2023think} & 40.0 & 78.0 & 55.0 & 30.0 & 72.0 & 42.0 & 50.0 & 85.0 & 65.0 & 20.0 & 48.0 & 38.0 \\
    & PoG~\cite{plan-on-graph} & 46.5 & \textbf{91.0} & 61.0 & 35.0 & 76.0 & 48.0 & 58.5 & \underline{88.0} & 76.0 & 25.5 & 54.0 & 45.0 \\
    \midrule
    \multirow{2}{*}{Hybrid RAG}
    & CoK~\cite{li2023chaincok} & 47.0 & 75.0 & 58.0 & 44.0 & \underline{78.0} & 56.0 & 62.0 & 74.0 & 66.0 & 23.0 & 58.0 & 50.0 \\
    & ToG-2~\cite{tog2.0ma2024think} & \underline{58.5} & 85.0 & \underline{71.0} & \underline{48.5} & \underline{78.0} & \underline{64.0} & \underline{64.5} & 84.0 & \underline{78.0} & \underline{34.0} & \underline{72.0} & \underline{62.0} \\
    \midrule
    Proposed & \textbf{\bianque} & \textbf{67.5} & \textbf{91.0} & \textbf{80.0} & \textbf{60.5} & \textbf{86.0} & \textbf{76.0} & \textbf{73.0} & \textbf{92.0} & \textbf{86.0} & \textbf{48.0} & \textbf{85.0} & \textbf{78.0} \\
    \bottomrule
  \end{tabular}
  }
\vspace{-4mm}
\end{table*}

\vspace{2mm}
\subsection{Ablation Study}
\vspace{-1mm}

\myparagraphquestion{How does the effectiveness of \bianque vary with different LLM capabilities}
We evaluated \bianque with nine representative LLM backbones. As shown in Figure~\ref{fig:backbone-robustness}, \bianque improves performance across all backbones by an average of 38.4\% over IO, with the largest absolute gain of 43.8\% on DeepSeek-V4. The gains are also clear on smaller backbones, such as Qwen3-4B and GPT-4o-mini, bringing weaker models close to or even beyond the IO prompting performance of stronger backbones. Stronger models also benefit from \bianque. On GPT-4-Turbo, Claude-3.7-Sonnet, Gemini-2.5-Pro, and DeepSeek-V4, \bianque still consistently outperforms IO, PoG, and ToG-2. Compared with ToG-2, \bianque achieves an average Overall Avg. gain of 11.5\% across all backbones. 
Overall,  \bianque enables deeper knowledge retrieval and more reliable and interpretable reasoning across LLMs of varying strength, rather than relying solely on their inherent knowledge.
The full task-family breakdown is reported in Table~\ref{tab:full-backbone-sweep}.
\vspace{-1mm}
\subsection{Effectiveness Evaluation}
\vspace{-1mm}

\myparagraphquestion{Does \bianque produce verifiable evidence chains}
We further evaluate \bianque with LLM-verifier grounding metrics, including answer correctness, evidence support, and reasoning completeness. 
As shown in Table~\ref{tab:main-llm-verifier}, \bianque achieves the best overall performance across all tasks, outperforming ToG-2 by 10.3\% on average and PoG by 18.2\%. 
The gains are especially clear on Intersection and Path Counting, where models must recover bridge entities, preserve relation constraints, and deduplicate biomedical entities. 
These results indicate that \bianque improves not only answer accuracy, but also the faithfulness and completeness of evidence chains, supporting the effectiveness of its source-conditioned evidence topology.

\textit{To further evaluate the performance, we conduct additional experiments, including ablation studies on source-conditioned evidence, evidence alignment, and evidence verification in Appendix~\ref{appendix:additional-ablation-analysis}; additional benchmark evaluation on reasoning-structure difficulty, MCQ-Open gaps, exact-vs-grounding disagreement, and LLM backbone robustness in Appendix~\ref{appendix:additional-benchmark-analysis}; reasoning faithfulness analysis in Appendix~\ref{appendix:faithfulness-analysis}; error and judge reliability analysis in Appendix~\ref{appendix:error-judge-analysis}; efficiency analysis in Appendix~\ref{appendix:effiency_analysis}; and case studies on cross-verified interpretable reasoning in Appendix~\ref{appendix:case-study}. A detailed outline is shown in \hyperref[outline]{Appendix Outline}.}

\vspace{-2mm}
\section{Conclusions}
\vspace{-3mm}

\label{conclusions}

In this work, we introduce \benchmark and \bianque for faithful biomedical reasoning. 
\benchmark is a multi-source graph-grounded benchmark that evaluates reasoning across KG, document, web, and hybrid evidence settings. 
\bianque is a source-aware framework that links KG paths and textual evidence into a unified evidence graph. 
It answers complex biomedical questions through multi-source evidence exploration, entity-level alignment, and verification-guided pruning. 
Extensive experiments show that \bianque outperforms existing baselines, demonstrating stronger biomedical reasoning capability and more faithful evidence support.

\section{Ethics Statement}
In this work, all benchmark construction and evaluation are based on publicly available biomedical knowledge sources, including biomedical KGs, literature-derived documents, and online evidence sources. 
We do not use private patient records or personally identifiable clinical data. 
We use LLMs to support source planning, evidence selection, and evidence-grounded answer generation, rather than relying on unconstrained open-ended generation alone. 
By grounding model outputs in KG, document, and web evidence, the ethical risks associated with hallucinated biomedical answers are expected to be reduced. 
However, LLMs and retrieved biomedical sources may still contain biases from pretraining data, literature coverage, or source curation. 
In future work, we plan to systematically investigate how such biases appear in multi-source biomedical reasoning and how evidence verification can further mitigate them.

\section{Limitations}
The primary limitation of \bianque is that it focuses on text-based biomedical knowledge sources, including KGs, documents, and web evidence. 
It does not incorporate external biomedical modalities such as medical images, molecular structures, or videos, which may also contain useful factual information. 
Integrating multimodal biomedical evidence alongside textual sources remains an important direction for future work and could further enhance the reasoning capability of the framework.

\bibliography{custom}

@article{singhal2023large,
  title={Large language models encode clinical knowledge},
  author={Singhal, Karan and Azizi, Shekoofeh and Tu, Tao and Mahdavi, S Sara and Wei, Jason and Chung, Hyung Won and Scales, Nathan and Tanwani, Ajay and Cole-Lewis, Heather and Pfohl, Stephen and others},
  journal={Nature},
  volume={620},
  number={7972},
  pages={172--180},
  year={2023},
  publisher={Nature Publishing Group UK London}
}

@article{lievin2024can,
  title={Can large language models reason about medical questions?},
  author={Li{\'e}vin, Valentin and Hother, Christoffer Egeberg and Motzfeldt, Andreas Geert and Winther, Ole},
  journal={Patterns},
  volume={5},
  number={3},
  year={2024},
  publisher={Elsevier}
}

@article{thirunavukarasu2023large,
  title={Large language models in medicine},
  author={Thirunavukarasu, Arun James and Ting, Darren Shu Jeng and Elangovan, Kabilan and Gutierrez, Laura and Tan, Ting Fang and Ting, Daniel Shu Wei},
  journal={Nature medicine},
  volume={29},
  number={8},
  pages={1930--1940},
  year={2023},
  publisher={Nature Publishing Group US New York}
}

@inproceedings{jin2019pubmedqa,
  title={Pubmedqa: A dataset for biomedical research question answering},
  author={Jin, Qiao and Dhingra, Bhuwan and Liu, Zhengping and Cohen, William and Lu, Xinghua},
  booktitle={Proceedings of the 2019 conference on empirical methods in natural language processing and the 9th international joint conference on natural language processing (EMNLP-IJCNLP)},
  pages={2567--2577},
  year={2019}
}

@article{tsatsaronis2015overview,
  title={An overview of the BIOASQ large-scale biomedical semantic indexing and question answering competition},
  author={Tsatsaronis, George and Balikas, Georgios and Malakasiotis, Prodromos and Partalas, Ioannis and Zschunke, Matthias and Alvers, Michael R and Weissenborn, Dirk and Krithara, Anastasia and Petridis, Sergios and Polychronopoulos, Dimitris and others},
  journal={BMC bioinformatics},
  volume={16},
  number={1},
  pages={138},
  year={2015},
  publisher={Springer}
}

@article{nentidis2023bioasq,
  title={BioASQ-QA: A manually curated corpus for Biomedical Question Answering},
  author={Krithara, Anastasia and Nentidis, Anastasios and Bougiatiotis, Konstantinos and Paliouras, Georgios},
  journal={Scientific data},
  volume={10},
  number={1},
  pages={170},
  year={2023},
  publisher={Nature Publishing Group UK London}
}

@inproceedings{jin2021disease,
  title={Infusing disease knowledge into BERT for health question answering, medical inference and disease name recognition},
  author={He, Yun and Zhu, Ziwei and Zhang, Yin and Chen, Qin and Caverlee, James},
  booktitle={Proceedings of the 2020 conference on empirical methods in natural language processing (emnlp)},
  pages={4604--4614},
  year={2020}
}

@inproceedings{pal2022medmcqa,
  title={Medmcqa: A large-scale multi-subject multi-choice dataset for medical domain question answering},
  author={Pal, Ankit and Umapathi, Logesh Kumar and Sankarasubbu, Malaikannan},
  booktitle={Conference on health, inference, and learning},
  pages={248--260},
  year={2022},
  organization={PMLR}
}

@inproceedings{kim2025biohopr,
  title={Biohopr: A benchmark for multi-hop, multi-answer reasoning in biomedical domain},
  author={Kim, Yunsoo and Abdulle, Yusuf and Wu, Honghan},
  booktitle={Findings of the Association for Computational Linguistics: ACL 2025},
  pages={12894--12908},
  year={2025}
}

@article{islamaj2026medhopqa,
  title={MedHopQA: A Disease-Centered Multi-Hop Reasoning Benchmark and Evaluation Framework for LLM-Based Biomedical Question Answering},
  author={Islamaj, Rezarta and Leaman, Robert and Chan, Joey and Wan, Nicholas and Jin, Qiao and Xie, Natalie and Wilbur, John and Tian, Shubo and Yeganova, Lana and Lai, Po-Ting and others},
  journal={arXiv preprint arXiv:2605.12361},
  year={2026}
}

@article{chandak2023primekg,
  title={Building a knowledge graph to enable precision medicine},
  author={Chandak, Payal and Huang, Kexin and Zitnik, Marinka},
  journal={Scientific data},
  volume={10},
  number={1},
  pages={67},
  year={2023},
  publisher={Nature Publishing Group UK London}
}

@article{himmelstein2017hetionet,
  title={Systematic integration of biomedical knowledge prioritizes drugs for repurposing},
  author={Himmelstein, Daniel Scott and Lizee, Antoine and Hessler, Christine and Brueggeman, Leo and Chen, Sabrina L and Hadley, Dexter and Green, Ari and Khankhanian, Pouya and Baranzini, Sergio E},
  journal={elife},
  volume={6},
  pages={e26726},
  year={2017},
  publisher={eLife Sciences Publications, Ltd}
}

@inproceedings{kim2024medexqa,
  title={MedExQA: Medical question answering benchmark with multiple explanations},
  author={Kim, Yunsoo and Wu, Jinge and Abdulle, Yusuf and Wu, Honghan},
  booktitle={Proceedings of the 23rd Workshop on biomedical natural language processing},
  pages={167--181},
  year={2024}
}

@article{lewis2020rag,
  title={Retrieval-augmented generation for knowledge-intensive nlp tasks},
  author={Lewis, Patrick and Perez, Ethan and Piktus, Aleksandra and Petroni, Fabio and Karpukhin, Vladimir and Goyal, Naman and K{\"u}ttler, Heinrich and Lewis, Mike and Yih, Wen-tau and Rockt{\"a}schel, Tim and others},
  journal={Advances in neural information processing systems},
  volume={33},
  pages={9459--9474},
  year={2020}
}

@article{bodenreider2004umls,
  title={The unified medical language system (UMLS): integrating biomedical terminology},
  author={Bodenreider, Olivier},
  journal={Nucleic acids research},
  volume={32},
  number={suppl\_1},
  pages={D267--D270},
  year={2004},
  publisher={Oxford University Press}
}

@article{robinson2008hpo,
  title={The Human Phenotype Ontology: a tool for annotating and analyzing human hereditary disease},
  author={Robinson, Peter N and K{\"o}hler, Sebastian and Bauer, Sebastian and Seelow, Dominik and Horn, Denise and Mundlos, Stefan},
  journal={The American Journal of Human Genetics},
  volume={83},
  number={5},
  pages={610--615},
  year={2008},
  publisher={Elsevier}
}

@article{mungall2017monarch,
  title={The Monarch Initiative: an integrative data and analytic platform connecting phenotypes to genotypes across species},
  author={Mungall, Christopher J and McMurry, Julie A and K{\"o}hler, Sebastian and Balhoff, James P and Borromeo, Charles and Brush, Matthew and Carbon, Seth and Conlin, Tom and Dunn, Nathan and Engelstad, Mark and others},
  journal={Nucleic acids research},
  volume={45},
  number={D1},
  pages={D712--D722},
  year={2017},
  publisher={Oxford University Press}
}

@article{morris2023spoke,
  title={The scalable precision medicine open knowledge engine (SPOKE): a massive knowledge graph of biomedical information},
  author={Morris, John H and Soman, Karthik and Akbas, Rabia E and Zhou, Xiaoyuan and Smith, Brett and Meng, Elaine C and Huang, Conrad C and Cerono, Gabriel and Schenk, Gundolf and Rizk-Jackson, Angela and others},
  journal={Bioinformatics},
  volume={39},
  number={2},
  pages={btad080},
  year={2023},
  publisher={Oxford University Press}
}

@article{pinero2015disgenet,
  title={DisGeNET: a discovery platform for the dynamical exploration of human diseases and their genes},
  author={Pi{\~n}ero, Janet and Queralt-Rosinach, N{\'u}ria and Bravo, Alex and Deu-Pons, Jordi and Bauer-Mehren, Anna and Baron, Martin and Sanz, Ferran and Furlong, Laura I},
  journal={Database},
  volume={2015},
  pages={bav028},
  year={2015},
  publisher={Oxford University Press}
}

@article{gao2023retrieval,
  title={Retrieval-augmented generation for large language models: A survey},
  author={Gao, Yunfan and Xiong, Yun and Gao, Xinyu and Jia, Kangxiang and Pan, Jinliu and Bi, Yuxi and Dai, Yixin and Sun, Jiawei and Wang, Haofen and Wang, Haofen and others},
  journal={arXiv preprint arXiv:2312.10997},
  volume={2},
  number={1},
  pages={32},
  year={2023}
}

@inproceedings{shao2023enhancing,
  title={Enhancing retrieval-augmented large language models with iterative retrieval-generation synergy},
  author={Shao, Zhihong and Gong, Yeyun and Shen, Yelong and Huang, Minlie and Duan, Nan and Chen, Weizhu},
  booktitle={Findings of the Association for Computational Linguistics: EMNLP 2023},
  pages={9248--9274},
  year={2023}
}

@article{wei2022cot,
  title={Chain-of-thought prompting elicits reasoning in large language models},
  author={Wei, Jason and Wang, Xuezhi and Schuurmans, Dale and Bosma, Maarten and Xia, Fei and Chi, Ed and Le, Quoc V and Zhou, Denny and others},
  journal={Advances in neural information processing systems},
  volume={35},
  pages={24824--24837},
  year={2022}
}

@inproceedings{tog1.0sun2023think,
  title={Think-on-graph: Deep and responsible reasoning of large language model on knowledge graph},
  author={Sun, Jiashuo and Xu, Chengjin and Tang, Lumingyuan and Wang, Saizhuo and Lin, Chen and Gong, Yeyun and Ni, Lionel and Shum, Heung-Yeung and Guo, Jian},
  booktitle={International Conference on Learning Representations},
  volume={2024},
  pages={3868--3898},
  year={2024}
}

@inproceedings{pogtan2025paths,
  title={Paths-over-graph: Knowledge graph empowered large language model reasoning},
  author={Tan, Xingyu and Wang, Xiaoyang and Liu, Qing and Xu, Xiwei and Yuan, Xin and Zhang, Wenjie},
  booktitle={Proceedings of the ACM on Web Conference 2025},
  pages={3505--3522},
  year={2025}
}

@article{tog2.0ma2024think,
  title={Think-on-graph 2.0: Deep and interpretable large language model reasoning with knowledge graph-guided retrieval},
  author={Ma, Shengjie and Xu, Chengjin and Jiang, Xuhui and Li, Muzhi and Qu, Huaren and Guo, Jian},
  journal={arXiv preprint arXiv:2407.10805},
  year={2024}
}

@inproceedings{li2023chaincok,
  title={Chain-of-knowledge: Grounding large language models via dynamic knowledge adapting over heterogeneous sources},
  author={Li, Xingxuan and Zhao, Ruochen and Chia, Yew Ken and Ding, Bosheng and others},
  booktitle={ICLR},
  year={2024}
}

@article{graphragmicrosoft,
  title={From local to global: A graph rag approach to query-focused summarization},
  author={Edge, Darren and Trinh, Ha and Cheng, Newman and Bradley, Joshua and Chao, Alex and Mody, Apurva and Truitt, Steven and Metropolitansky, Dasha and Ness, Robert Osazuwa and Larson, Jonathan},
  journal={arXiv preprint arXiv:2404.16130},
  year={2024}
}

@inproceedings{sarmah2024hybridrag,
  author       = {Bhaskarjit Sarmah and
                  Dhagash Mehta and
                  Benika Hall and
                  Rohan Rao and
                  Sunil Patel and
                  Stefano Pasquali},
  title        = {HybridRAG: Integrating Knowledge Graphs and Vector Retrieval Augmented
                  Generation for Efficient Information Extraction},
  booktitle    = {Proceedings of the 5th {ACM} International Conference on {AI} in Finance,
                  {ICAIF} 2024, Brooklyn, NY, USA, November 14-17, 2024},
  pages        = {608--616},
  publisher    = {{ACM}},
  year         = {2024},
  bibsource    = {dblp computer science bibliography, https://dblp.org}
}

@inproceedings{yao2022react,
  title={React: Synergizing reasoning and acting in language models},
  author={Yao, Shunyu and Zhao, Jeffrey and Yu, Dian and Du, Nan and Shafran, Izhak and Narasimhan, Karthik and Cao, Yuan},
  booktitle={ICLR},
  year={2023}
}

@inproceedings{jiang2023structgpt,
  title={StructGPT: A General Framework for Large Language Model to Reason over Structured Data},
  author={Jiang, Jinhao and Zhou, Kun and Dong, Zican and Ye, Keming and Zhao, Wayne Xin and Wen, Ji-Rong},
  booktitle={Proceedings of the 2023 Conference on Empirical Methods in Natural Language Processing},
  year={2023}
}

@inproceedings{plan-on-graph,
  title={Plan-on-Graph: Self-Correcting Adaptive Planning of Large Language Model on Knowledge Graphs},
  author={Chen, Liyi and Tong, Panrong and Jin, Zhongming and Sun, Ying and Ye, Jieping and Xiong, Hui},
  booktitle={NeurIPS},
  year={2024}
}

@inproceedings{debated-on-graph,
  title={Debate on Graph: A Flexible and Reliable Reasoning Framework for Large Language Models},
  author={Ma, Jie and Gao, Zhitao and Chai, Qi and Sun, Wangchun and Wang, Pinghui and Pei, Hongbin and others},
  booktitle={AAAI},
  pages={24768--24776},
  year={2025}
}

@inproceedings{reimers-gurevych-2019-sentence,
  author       = {Nils Reimers and
                  Iryna Gurevych},
  editor       = {Kentaro Inui and
                  Jing Jiang and
                  Vincent Ng and
                  Xiaojun Wan},
  title        = {Sentence-BERT: Sentence Embeddings using Siamese BERT-Networks},
  booktitle    = {Proceedings of the 2019 Conference on Empirical Methods in Natural
                  Language Processing and the 9th International Joint Conference on
                  Natural Language Processing, {EMNLP-IJCNLP} 2019, Hong Kong, China,
                  November 3-7, 2019},
  pages        = {3980--3990},
  publisher    = {Association for Computational Linguistics},
  year         = {2019},
  bibsource    = {dblp computer science bibliography, https://dblp.org}
}

\appendix

\onecolumn
\section*{Appendix Outline}
\label{outline}
\hypersetup{hidelinks}
\newcommand{\OutlineItem}[3]{\hspace*{#1}\hyperref[#2]{#3}\dotfill\pageref{#2}\\}
\newcommand{\OutlineItemGap}[3]{\hspace*{#1}\hyperref[#2]{#3}\dotfill\pageref{#2}\\[3pt]}
\OutlineItem{0pt}{appendix:alg}{A.\,Algorithm}
\OutlineItem{1.5em}{alg:benchmark_construction}{A.1 \benchmark Construction Pipeline}
\OutlineItem{1.5em}{alg:bioweave_query}{A.2 \bianque Framework Overview}
\OutlineItem{1.5em}{alg:structured_retrieval}{A.3 Structured Retrieval}
\OutlineItem{1.5em}{alg:unstructured_retrieval}{A.4 Unstructured Retrieval}
\OutlineItem{1.5em}{alg:bioweave_explore}{A.5 Multi-source Evidence Exploration}
\OutlineItem{1.5em}{alg:bioweave_prune}{A.6 Evidence Pruning and Verification}
\OutlineItemGap{1.5em}{tab:algorithm-notation}{A.7 Algorithm Notation}
\OutlineItemGap{0pt}{appendix:workflow_diagram}{B.\,Workflow Diagram}
\OutlineItem{0pt}{appendix:additional-experiments}{C.\,Additional Experiments}
\OutlineItem{1.5em}{appendix:additional-ablation-analysis}{C.1 Ablation Study}
\OutlineItem{3em}{exp:source-conditioned-evidence}{How does source-conditioned evidence affect \bianque?}
\OutlineItem{3em}{exp:evidence-alignment}{How does evidence alignment affect \bianque?}
\OutlineItem{3em}{exp:evidence-verification}{How does evidence verification affect \bianque?}
\OutlineItem{1.5em}{appendix:additional-benchmark-analysis}{C.2 Additional Benchmark Evaluation and Analysis}
\OutlineItem{3em}{exp:reasoning-structure-difficulty}{How difficult are different reasoning structures in \benchmark?}
\OutlineItem{3em}{exp:mcq-open-gap}{Does the multiple-choice format overestimate biomedical reasoning?}
\OutlineItem{3em}{exp:grounding-exactness-gap}{Where does evidence grounding differ from exact answering?}
\OutlineItem{3em}{exp:backbone-diagnostic}{Does \benchmark remain diagnostic across LLM backbones?}
\OutlineItem{1.5em}{appendix:faithfulness-analysis}{C.3 Faithfulness Analysis}
\OutlineItem{3em}{exp:evidence-source-distribution}{Evidence sources of correct answers}
\OutlineItem{3em}{sec:exp:incomplete_kg}{Robustness against incomplete KGs and source routing}
\OutlineItem{3em}{appendix:evidence-path-overlap}{Overlap between retrieved KG paths and ground-truth paths}
\OutlineItem{3em}{exp:evidence-support-heatmap}{Evidence support across task families}
\OutlineItem{1.5em}{appendix:error-judge-analysis}{C.4 Error and Judge Analysis}
\OutlineItem{3em}{sec:exp:error_analysis}{Error Analysis}
\OutlineItem{3em}{appendix:judge-reliability}{Judge reliability analysis}
\OutlineItem{1.5em}{appendix:effiency_analysis}{C.5 Efficiency Analysis}
\OutlineItem{3em}{exp:llm-calls-cost}{LLM calls cost analysis}
\OutlineItemGap{3em}{exp:efficiency-analysis-biomedhop}{Efficiency analysis on \benchmark}
\OutlineItem{0pt}{sec:appendix}{D.\,Benchmark Details}
\OutlineItem{1.5em}{appendix:benchmark-task-construction}{D.1 Task Construction Details}
\OutlineItem{3em}{appendix:benchmark-task-construction}{Task Family}
\OutlineItem{4.5em}{fig:entity-pair-example}{Entity Pair Matching}
\OutlineItem{4.5em}{fig:intersection-reasoning-example}{Intersection Reasoning}
\OutlineItem{4.5em}{fig:path-reasoning-example}{Path-based Reasoning}
\OutlineItem{4.5em}{fig:path-counting-example}{Path-based Counting}
\OutlineItem{3em}{tab:path-templates}{Path-based reasoning templates}
\OutlineItem{3em}{tab:dataset-statistics}{Dataset statistics}
\OutlineItem{1.5em}{appendix:benchmark-source-evidence}{D.2 Source-conditioned Evidence Construction}
\OutlineItemGap{1.5em}{appendix:benchmark-metrics}{D.3 Benchmark Metrics}
\OutlineItem{0pt}{sec:experiment-details}{E.\,Experiment Details}
\OutlineItem{1.5em}{appendix:evaluation-setup}{E.1 Experiment datasets}
\OutlineItem{1.5em}{appendix:compared-methods}{E.2 Experiment baselines}
\OutlineItem{1.5em}{appendix:backbone-models}{E.3 Backbone models}
\OutlineItemGap{1.5em}{appendix:implementation-details}{E.4 Experiment implementation}
\OutlineItem{0pt}{sec:detailed-related-work}{F.\,Detailed Related Work}
\OutlineItem{1.5em}{relwork:rag}{Retrieval-augmented LLMs}
\OutlineItem{1.5em}{relwork:kg-rag}{KG-based RAG}
\OutlineItem{1.5em}{relwork:hybrid-rag}{Hybrid RAG}
\OutlineItem{1.5em}{relwork:biomedical-multihop-benchmarks}{Biomedical Multi-hop Benchmarks}
\OutlineItem{1.5em}{relwork:biomedical-kg-reasoning}{Biomedical Knowledge Graph Reasoning}
\OutlineItemGap{1.5em}{relwork:biomedical-qa}{Biomedical question answering}
\OutlineItem{0pt}{appendix:case-study}{G.\,Case Study: Multi-Source Cross-Verified Interpretable Reasoning}
\OutlineItem{1.5em}{tab:case-study-kg-doc-hlab}{Case study example on KG--Doc cross-verified reasoning}
\OutlineItem{1.5em}{tab:case-study-kg-web-egfr}{Case study example on KG--Web cross-verified reasoning}
\OutlineItem{1.5em}{tab:case-study-doc-web-olaparib}{Case study example on Doc--Web cross-verified reasoning}
\OutlineItem{1.5em}{tab:case-study-all-cml}{Case study example on all-source cross-verified reasoning}
\OutlineItem{0pt}{appendix:prompts}{H.\,Prompts}
\OutlineItem{1.5em}{prompt:question-rewriting}{Benchmark Question Rewriting}
\OutlineItem{1.5em}{prompt:question-initialization}{Question Initialization and Source Planning}
\OutlineItem{1.5em}{prompt:retrieval-query-construction}{Retrieval Query Construction}
\OutlineItem{1.5em}{prompt:bioweave-source-repair}{\bianque Source Repair}
\OutlineItem{1.5em}{prompt:unified-evidence-graph-conversion}{Unified Evidence Graph Conversion}
\OutlineItem{1.5em}{prompt:path-selection}{LLM-aware Evidence Selection}
\OutlineItem{1.5em}{prompt:bioweave-graph-refinement}{\bianque Evidence Graph Refinement}
\OutlineItem{1.5em}{prompt:source-conditioned-answering}{Source-conditioned Answering}
\OutlineItem{1.5em}{prompt:bioweave-verified-answer}{\bianque Verified Answer Generation}
\OutlineItem{1.5em}{prompt:llm-verifier}{LLM Verifier}

\clearpage
\twocolumn

\section{Algorithm}
\label{appendix:alg}
\definecolor{qblue}{RGB}{30,90,180}
\definecolor{trgreen}{RGB}{0,140,60}
\definecolor{dmpurple}{RGB}{120,70,170}
\definecolor{wmorange}{RGB}{210,110,20}
\definecolor{agent}{RGB}{0,130,130}

\providecommand{\Badge}[2][black]{%
  \begingroup
  \setlength{\fboxsep}{1.5pt}%
  \fcolorbox{#1}{white}{\textcolor{#1}{\scriptsize\sffamily\bfseries #2}}%
  \endgroup
}

\providecommand{\QTool}[1]{%
  \ensuremath{\mathop{\text{\Badge[qblue]{#1}}}\nolimits}%
}

\providecommand{\WMTool}[1]{%
  \ensuremath{\mathop{\text{\Badge[wmorange]{#1}}}\nolimits}%
}

\providecommand{\TreeTool}[1]{%
  \ensuremath{\mathop{\text{\Badge[trgreen]{#1}}}\nolimits}%
}

\providecommand{\AgentTool}[1]{%
  \ensuremath{\mathop{\text{\Badge[agent]{#1}}}\nolimits}%
}

\providecommand{\DMTool}[1]{%
  \ensuremath{\mathop{\text{\Badge[dmpurple]{#1}}}\nolimits}%
}

\providecommand{\PlainTool}[1]{%
  \ensuremath{
    \mathop{
      \text{
        \textcolor{black}{
          \scriptsize\sffamily\bfseries #1
        }
      }
    }\nolimits
  }%
}

\subsection{Benchmark Construction}\label{appendix:alg:benchmark}
Algorithm~\ref{alg:benchmark_construction} summarizes the \benchmark construction pipeline.
The algorithm uses the colored tool  below:
\QTool{Blue} denotes question rendering and answer validation,
\TreeTool{Green} denotes KG pattern construction,
\WMTool{Orange} denotes evidence retrieval and storage,
\AgentTool{Teal} denotes LLM-assisted text normalization, and
\DMTool{Purple} denotes source-condition and quality-control decisions.
\begin{algorithm}[H]
\small
\SetVline
\caption{\small BenchmarkConstruction}\label{alg:benchmark_construction}
\Input{Biomedical KGs $\mathbb{G}$, document index $\mathcal{I}_{doc}$, web interface $\mathcal{W}$, task specs $\mathcal{T}$, source conditions $\mathcal{C}_{src}$, target size $N$}
\Output{Benchmark dataset $\mathcal{D}$}

\State{$\mathcal{D}\gets \emptyset$}
\State{$\mathcal{C}_{src}\gets \operatorname{SingleSrc}\cup\operatorname{HybridSrc}$}

\While{$|\mathcal{D}|<N$}{
  \ForEach{$\tau\in\mathcal{T}$}{
    \State{$\mathcal{G}^{gold}_{q},~a,~o(q)\gets \TreeTool{InstantiateGraphPattern}(\mathbb{G},\tau)$}
    \If{$\neg\DMTool{ValidateGraphAnswer}(\mathcal{G}^{gold}_{q},a,o(q))$}{
      \textbf{continue}
    }

    \State{$\mathcal{Q}_{hop}\gets \WMTool{BuildHopQueries}(\mathcal{G}^{gold}_{q})$}
    \State{$\mathcal{E}_{kg}\gets \TreeTool{SerializeKGEvidence}(\mathcal{G}^{gold}_{q})$}
    \State{$\mathcal{E}_{doc}\gets \WMTool{RetrieveMultiDoc}(\mathcal{I}_{doc},\mathcal{Q}_{hop})$}
    \State{$\mathcal{Q}_{web}\gets \AgentTool{BuildTargetedQueries}(\mathcal{G}^{gold}_{q},\mathcal{Q}_{hop})$}
    \State{$\mathcal{R}_{web}\gets \AgentTool{SearchWeb}(\mathcal{W},\mathcal{Q}_{web})$}
    \State{$\mathcal{E}_{web}\gets \WMTool{EntityAwareFilter}(\mathcal{R}_{web},\mathcal{G}^{gold}_{q})$}
    \State{$\mathcal{E}_{doc},\mathcal{E}_{web}\gets \AgentTool{NormalizeMentions}(\mathcal{E}_{doc},\mathcal{E}_{web},\mathcal{G}^{gold}_{q})$}
    \State{$\mathcal{B}\gets \{\mathcal{E}_{kg},\mathcal{E}_{doc},\mathcal{E}_{web}\}$}

    \ForEach{$S\in\mathcal{C}_{src}$}{
      \State{$\mathcal{E}_{S}\gets \DMTool{ComposeSourceEvidence}(S,\mathcal{B})$}
      \If{$\neg\DMTool{ValidateSourceCoverage}(\mathcal{E}_{S},\mathcal{G}^{gold}_{q},S)$}{
        \textbf{continue}
      }
      \If{$o(q)=\text{counting}$}{
        \State{$q_{count}\gets \QTool{RenderCountQuestion}(\mathcal{G}^{gold}_{q},a,o(q))$}
        \State{$I_{qa}\gets \QTool{PackCountQA}(q_{count},a)$}
        \State{$\mathcal{O}\gets \emptyset$}
      }
      \Else{
        \State{$q_{mcq},~\mathcal{O}\gets \QTool{RenderMCQ}(\mathcal{G}^{gold}_{q},a,o(q))$}
        \State{$q_{open}\gets \QTool{RenderOpenQuestion}(\mathcal{G}^{gold}_{q},a,o(q))$}
        \State{$I_{qa}\gets \QTool{PackQA}(q_{mcq},q_{open},a,\mathcal{O})$}
      }
      \State{$m\gets \WMTool{StoreMetadata}(\mathcal{G}^{gold}_{q},S,\mathcal{E}_{S},o(q),\mathcal{O})$}
      \State{$I_{ev}\gets \WMTool{PackEvidence}(\mathcal{G}^{gold}_{q},S,\mathcal{E}_{S},m)$}
      \State{$\mathcal{D}\gets \mathcal{D}\cup\{(I_{qa},I_{ev})\}$}
      \If{$|\mathcal{D}|\ge N$}{
        \textbf{break}
      }
    }
  }
}
\Return $\mathcal{D}$\;
\end{algorithm}

\subsection{\bianque Framework Overview}\label{appendix:alg:query}
We summarize the overall \bianque query procedure in Algorithm~\ref{alg:bioweave_query}.
The controller first abstracts the biomedical question, initializes a source plan, and then repeatedly invokes an evidence agent.
If the selected evidence graph does not satisfy the predicted reasoning operator, the controller backtraces the missing hop and updates the source plan.
Colored badges indicate module types:
\QTool{Blue} denotes question analysis and answer generation,
\WMTool{Orange} denotes evidence working-memory operations,
\TreeTool{Green} denotes graph-structured KG retrieval,
\AgentTool{Teal} denotes LLM/controller tool calls, and
\DMTool{Purple} denotes decision, verification, and evidence-graph management.

\begin{algorithm}[H]
\small
\SetVline
\caption{\small FrameworkOverview}\label{alg:bioweave_query}
\Input{Question $q$, biomedical KGs $\mathbb{G}$, document index $\mathcal{I}_{doc}$, web interface $\mathcal{W}$, max repair steps $T$}
\Output{Final answer $a$, refined evidence graph $\mathcal{E}^{*}_{q}$}

\State{$E_q,~\tau(q),~o(q),~\mathcal{C}_q \gets \QTool{AbstractQuestion}(q)$}
\State{$S_q \gets \AgentTool{InitSourcePlan}(q,E_q,\tau(q),o(q),\mathcal{C}_q)$}
\State{$\mathcal{M} \gets \WMTool{InitEvidenceMemory}()$}
\State{$\mathcal{H}_{miss} \gets \emptyset$}
\State{$\mathcal{F} \gets \{\QTool{InitPlan}(q,S_q,\mathcal{C}_q,o(q),\mathcal{H}_{miss})\}$}

\While{$\mathcal{F} \neq \emptyset$ \textbf{and} $T>0$}{
  \State{$R \gets \mathcal{F}.pop()$}
	  \State{$\Theta_R\gets(R,\mathbb{G},\mathcal{I}_{doc},\mathcal{W},\mathcal{M})$}
	  \State{$\mathcal{U}_q,~\mathcal{G}_q,~\mathcal{M} \gets \AgentTool{MultiSourceEvidenceExploration}(\Theta_R)$}
	  \State{$\mathcal{E}_{q} \gets \DMTool{CrossVerify}(\mathcal{U}_q,\mathcal{G}_q,q,o(q))$}
  \State{$\mathcal{E}^{*}_{q} \gets \DMTool{RefineEvidenceGraph}(\mathcal{E}_{q},o(q))$}
  \If{$\DMTool{SatisfyOperator}(\mathcal{E}^{*}_{q},o(q))$}{
    \State{$a \gets \QTool{GenFinalAnswer}(q,\mathcal{E}^{*}_{q},\tau(q))$}
    \Return $a,\mathcal{E}^{*}_{q}$\;
  }
  \State{$\mathcal{H}_{miss} \gets \DMTool{BacktraceMissingHop}(\mathcal{E}^{*}_{q},o(q))$}
  \State{$S_q \gets \AgentTool{UpdateSourcePlan}(S_q,\mathcal{H}_{miss})$}
  \State{$\mathcal{F}.push(\QTool{InitPlan}(q,S_q,\mathcal{C}_q,o(q),\mathcal{H}_{miss}))$}
  \State{$T \gets T-1$}
}

\State{$a \gets \QTool{GenFinalAnswer}(q,\mathcal{E}^{*}_{q},\tau(q))$}
\Return $a,\mathcal{E}^{*}_{q}$\;
\end{algorithm}

\newpage
\subsection{Structured Retrieval}\label{appendix:alg:structured}
Algorithm~\ref{alg:structured_retrieval} details the KG-side retrieval procedure.
It first constructs or reuses a biomedical evidence subgraph around the current entity candidates and missing hops, then retrieves operator-compatible KG paths as constrained reasoning traces.

\begin{algorithm}[H]
\small
\SetVline
\caption{\small StructuredRetrieval}\label{alg:structured_retrieval}
\Input{Biomedical KGs $\mathbb{G}$, KG subgraph $\mathcal{G}_{q}$, source plan $S_q$, candidate entities $\mathcal{C}_{q}$, operator $o(q)$, missing hops $\mathcal{H}_{miss}$, max depth $D_{\max}$, path budget $W_1$}
\Output{KG paths $\mathcal{P}_{kg}$, updated KG subgraph $\mathcal{G}_{q}$}

\State{$\mathcal{P}_{kg}\gets \emptyset$}
\If{$\mathrm{KG}\notin S_q$}{
  \Return $\mathcal{P}_{kg},\mathcal{G}_{q}$\;
}
\If{$\mathcal{G}_{q}=\emptyset$}{
  \State{$\Theta_{sub}\gets(\mathbb{G},\mathcal{C}_{q},o(q),\mathcal{H}_{miss},D_{\max})$}
  \State{$\mathcal{G}_{q}\gets \TreeTool{DetectEvidenceSubgraph}(\Theta_{sub})$}
  \State{$\mathcal{G}_{q}\gets \TreeTool{PruneSubgraph}(\mathcal{G}_{q},\mathcal{C}_{q},o(q))$}
}
\State{$\mathcal{P}_{kg}\gets \TreeTool{ExpandTypedPaths}(\mathcal{G}_{q},\mathcal{C}_{q},o(q),\mathcal{H}_{miss},D_{\max},W_1)$}
\State{$\mathcal{P}_{kg}\gets \TreeTool{AttachKGProvenance}(\mathcal{P}_{kg},\mathcal{G}_{q})$}
\Return $\mathcal{P}_{kg},\mathcal{G}_{q}$\;

\vspace{2mm}
{\textbf{Procedure} \TreeTool{ExpandTypedPaths}$(\mathcal{G}_{q},\mathcal{C}_{q},o(q),\mathcal{H}_{miss},D_{\max},W_1)$\\
            \SetVline
            \small
\State{$d\gets 1$;~ $\mathcal{P}\gets\emptyset$;~ $\mathcal{F}_{ent}\gets \mathcal{C}_{q}$}
\While{$d\le D_{\max}$}{
  \State{$\mathcal{P}^{d}\gets \emptyset$;~ $\mathcal{F}'_{ent}\gets \emptyset$}
  \ForEach{$e\in \mathcal{F}_{ent}$}{
    \State{$\mathcal{P}_{e},~\mathcal{N}_{e}\gets \TreeTool{ExpandOneHop}(\mathcal{G}_{q},e,o(q),\mathcal{H}_{miss})$}
    \State{$\mathcal{P}^{d}\gets \mathcal{P}^{d}\cup \mathcal{P}_{e}$}
    \State{$\mathcal{F}'_{ent}\gets \mathcal{F}'_{ent}\cup \mathcal{N}_{e}$}
  }
  \State{$\mathcal{P}^{d}\gets \TreeTool{FilterOperatorPaths}(\mathcal{P}^{d},\mathcal{C}_{q},o(q),d)$}
  \State{$\mathcal{P}\gets \mathcal{P}\cup\mathcal{P}^{d}$}
  \If{$|\mathcal{P}|>W_1$}{
    \State{$\mathcal{P}\gets \TreeTool{RankPrunePaths}(\mathcal{P},\mathcal{C}_{q},o(q),W_1)$}
  }
  \State{$\mathcal{F}_{ent}\gets \TreeTool{UpdatePathFrontier}(\mathcal{F}'_{ent},\mathcal{P},\mathcal{C}_{q},o(q))$}
  \If{$\mathcal{F}_{ent}=\emptyset$}{
    \textbf{break}
  }
  \State{$d\gets d+1$}
}
\Return $\mathcal{P}$\;
}
\end{algorithm}

\newpage
\subsection{Unstructured Retrieval}\label{appendix:alg:unstructured}
Algorithm~\ref{alg:unstructured_retrieval} describes document and web retrieval.
It builds hop-aware retrieval queries from entities, aliases, relation phrases, KG paths, and missing-hop signals, then converts selected text spans into graph-compatible evidence units.

\begin{algorithm}[H]
\small
\SetVline
\caption{\small UnstructuredRetrieval}\label{alg:unstructured_retrieval}
\Input{Source plan $S_q$, document index $\mathcal{I}_{doc}$, web interface $\mathcal{W}$, question $q$, candidate entities $\mathcal{C}_{q}$, KG paths $\mathcal{P}_{kg}$, missing hops $\mathcal{H}_{miss}$, budgets $W_{doc},W_{web}$}
\Output{Document evidence units $\mathcal{U}_{doc}$, web evidence units $\mathcal{U}_{web}$}

\State{$\mathcal{U}_{doc},~\mathcal{U}_{web}\gets \emptyset$}
\If{$\mathrm{Doc}\in S_q$}{
  \State{$\mathcal{Q}_{doc}\gets \WMTool{BuildHopQueries}(q,\mathcal{C}_{q},\mathcal{P}_{kg},\mathcal{H}_{miss})$}
  \State{$\mathcal{D}_{q}\gets \WMTool{RetrieveMultiDoc}(\mathcal{I}_{doc},\mathcal{Q}_{doc})$}
  \State{$\mathcal{S}_{doc}\gets \WMTool{EntityAwareFilter}(\mathcal{D}_{q},\mathcal{C}_{q},W_{doc})$}
  \State{$\mathcal{U}_{doc}\gets \AgentTool{ConvertEvidenceGraph}(\mathcal{S}_{doc},q,\mathcal{C}_{q})$}
}
\If{$\mathrm{Web}\in S_q$}{
  \State{$\mathcal{Q}_{web}\gets \AgentTool{BuildTargetedQueries}(q,\mathcal{C}_{q},\mathcal{H}_{miss})$}
  \State{$\mathcal{R}_{web}\gets \AgentTool{SearchWeb}(\mathcal{W},\mathcal{Q}_{web})$}
  \State{$\mathcal{S}_{web}\gets \WMTool{EntityAwareFilter}(\mathcal{R}_{web},\mathcal{C}_{q},W_{web})$}
  \State{$\mathcal{U}_{web}\gets \AgentTool{ConvertEvidenceGraph}(\mathcal{S}_{web},q,\mathcal{C}_{q})$}
  \State{$\mathcal{U}_{web}\gets \DMTool{AssignLowerSourcePrior}(\mathcal{U}_{web})$}
}
\Return $\mathcal{U}_{doc},\mathcal{U}_{web}$\;
\end{algorithm}

\subsection{Multi-source Evidence Exploration}\label{appendix:alg:exploration}\label{appendix:sec:explora}
Algorithm~\ref{alg:bioweave_explore} shows how the evidence agent combines structured and unstructured retrieval.
The agent updates working memory after each channel and returns a unified candidate evidence pool for verification.

\begin{algorithm}[H]
\small
\SetVline
\caption{\small MultiSourceEvidenceExploration}\label{alg:bioweave_explore}
\Input{Retrieval plan $R=(q,S_q,\mathcal{C}_q,o(q),\mathcal{H}_{miss})$, biomedical KGs $\mathbb{G}$, document index $\mathcal{I}_{doc}$, web interface $\mathcal{W}$, evidence memory $\mathcal{M}$, budgets $D_{\max},W_1,W_{doc},W_{web}$}
\Output{Candidate evidence units $\mathcal{U}_{q}$, KG subgraph $\mathcal{G}_{q}$, updated memory $\mathcal{M}$}

\State{$\mathcal{G}_{q}\gets \WMTool{GetCachedSubgraph}(\mathcal{M},q)$}
\State{$\Theta_{kg}\gets(\mathbb{G},\mathcal{G}_{q},S_q,\mathcal{C}_{q},o(q),\mathcal{H}_{miss},D_{\max},W_1)$}
\State{$\mathcal{P}_{kg},\mathcal{G}_{q}\gets \TreeTool{StructuredRetrieval}(\Theta_{kg})$}
\State{$\mathcal{M}\gets \WMTool{UpdateEvidenceMemory}(\mathcal{M},\mathcal{P}_{kg})$}
\State{$\Theta_{txt}\gets(S_q,\mathcal{I}_{doc},\mathcal{W},q,\mathcal{C}_{q},\mathcal{P}_{kg},\mathcal{H}_{miss},W_{doc},W_{web})$}
\State{$\mathcal{U}_{doc},\mathcal{U}_{web}\gets \AgentTool{UnstructuredRetrieval}(\Theta_{txt})$}
\State{$\mathcal{M}\gets \WMTool{UpdateEvidenceMemory}(\mathcal{M},\mathcal{U}_{doc}\cup\mathcal{U}_{web})$}
\State{$\mathcal{U}_{q}\gets \mathcal{P}_{kg}\cup\mathcal{U}_{doc}\cup\mathcal{U}_{web}$}
\Return $\mathcal{U}_{q},\mathcal{G}_{q},\mathcal{M}$\;
\end{algorithm}

\newpage
\subsection{Evidence Pruning and Verification}\label{appendix:alg:verification}
Algorithm~\ref{alg:bioweave_prune} describes the decision module used after evidence exploration.
It first builds a relevance-filtered candidate pool, then performs source-aware verification with cross-source support signals and KG grounding before LLM-aware selection.

\begin{algorithm}[H]
\small
\SetVline
\caption{\small EvidencePruningAndVerification}\label{alg:bioweave_prune}
\Input{Candidate evidence units $\mathcal{U}_{q}$, KG subgraph $\mathcal{G}_{q}$, question $q$, reasoning operator $o(q)$, budgets $W_1,W_2,K$, threshold $\gamma$}
\Output{Source-aware evidence graph $\mathcal{E}_{q}$}

\ForEach{$u_i\in\mathcal{U}_{q}$}{
  \State{$r_i\gets \operatorname{cos}(\mathbf{h}(q),\mathbf{h}(u_i))$}
  \State{$c_i\gets \operatorname{Jaccard}(\operatorname{Topic}(q),\operatorname{Ent}(u_i))$}
  \State{$s_i^{rel}\gets \lambda_{sem}r_i+\lambda_{ent}c_i$}
}
\State{$\widetilde{\mathcal{U}}_{q}\gets \DMTool{TopK}(\mathcal{U}_{q},s^{rel},W_1)$}

\ForEach{$u_i\in\widetilde{\mathcal{U}}_{q}$}{
  \State{$\operatorname{Supp}(u_i)\gets \DMTool{FindSourceSupport}(u_i,\widetilde{\mathcal{U}}_{q},\gamma)$}
  \State{$f_1(u_i)\gets \rho_{\operatorname{src}(u_i)}$}
  \State{$f_2(u_i)\gets \min(|\operatorname{Supp}(u_i)|,W_{src})/W_{src}$}
  \State{$f_3(u_i)\gets |\operatorname{Ent}(u_i)\cap V(\mathcal{G}_{q})|/\max(1,|\operatorname{Ent}(u_i)|)$}
  \State{$s_i^{ver}\gets \alpha_1f_1(u_i)+\alpha_2f_2(u_i)+\alpha_3f_3(u_i)$}
  \State{$\operatorname{CrossScore}(u_i)\gets \beta s_i^{rel}+(1-\beta)s_i^{ver}$}
}
\State{$\mathcal{U}_{top}\gets \DMTool{TopK}(\widetilde{\mathcal{U}}_{q},\operatorname{CrossScore},W_2)$}
\State{$\mathcal{U}_{sel}\gets \AgentTool{LLMSelect}(\mathcal{U}_{top},q,o(q),K)$}
\State{$\mathcal{E}_{q}\gets \DMTool{MergeAliasProvenance}(\mathcal{U}_{sel})$}
\Return $\mathcal{E}_{q}$\;
\end{algorithm}

\newpage
\subsection{Algorithm Notation}\label{appendix:alg:notation}
Table~\ref{tab:algorithm-notation} summarizes the notation used in the appendix algorithms.
\begin{table}[H]
\scriptsize
\centering
\renewcommand{\arraystretch}{0.88}
\vspace{-3mm}
\caption{AlgorithmNotation}
\vspace{-3mm}
\label{tab:algorithm-notation}
\begin{tabular}{@{}p{0.37\linewidth}p{0.55\linewidth}@{}}
\hline
\textbf{Symbol} & \textbf{Meaning} \\
\hline
$q$ & input biomedical question \\
$a$ & final answer \\
$E_q$ & biomedical mentions in $q$ \\
$\tau(q)$ & expected answer format/type \\
$o(q)$ & reasoning operator \\
$\mathcal{C}_q$ & candidate normalized entities \\
$S_q$ & selected source plan \\
$R$ & current retrieval plan \\
$\mathbb{G}$ & biomedical KG collection \\
$\mathcal{G}^{gold}_{q}$ & answer-determining graph pattern used in benchmark construction \\
$\mathcal{I}_{doc}$ & biomedical document index \\
$\mathcal{W}$ & web search interface \\
$\mathcal{M}$ & evidence working memory \\
$\mathcal{F}$ & retrieval-plan frontier \\
$D_{\max}$ & maximum KG retrieval depth \\
$\mathcal{P}_{kg}$ & KG-derived paths \\
$\mathcal{F}_{ent}$ & entity frontier for KG path expansion \\
$\mathcal{F}'_{ent}$ & next-hop entity frontier \\
$\mathcal{P}^{d}$ & KG paths retrieved at depth $d$ \\
$\mathcal{P}_{e}$ & one-hop paths expanded from entity $e$ \\
$\mathcal{N}_{e}$ & neighbor entities reached from entity $e$ \\
$\mathcal{U}_{doc}$ & document-derived evidence units \\
$\mathcal{U}_{web}$ & web-derived evidence units \\
$\mathcal{Q}_{doc}$ & hop-aware document retrieval queries \\
$\mathcal{Q}_{web}$ & targeted web retrieval queries \\
$\mathcal{U}_{q}$ & unified candidate evidence-unit pool \\
$\widetilde{\mathcal{U}}_{q}$ & relevance-filtered candidate evidence-unit pool \\
$\mathcal{G}_{q}$ & retrieved KG subgraph \\
$\mathcal{E}_{q}$ & source-aware evidence graph \\
$\mathcal{E}^{*}_{q}$ & refined evidence graph \\
$\mathcal{H}_{miss}$ & missing or uncertain hops \\
$\operatorname{prov}(u)$ & provenance record of evidence unit $u$ \\
$\operatorname{hop}(u)$ & reasoning hop supported by evidence unit $u$ \\
$s_i^{rel}$ & semantic and entity relevance score for evidence unit $u_i$ \\
$s_i^{ver}$ & source-aware verification score for evidence unit $u_i$ \\
$\operatorname{Supp}(u_i)$ & compatible source channels supporting evidence unit $u_i$ \\
$\lambda_{\mathrm{sem}},\lambda_{\mathrm{ent}}$ & semantic and entity-overlap weights in relevance scoring \\
$\rho_{\operatorname{src}(u)}$ & source prior for an evidence unit from KG, document, or web evidence \\
$\alpha_{\mathrm{prior}},\alpha_{\mathrm{sup}},\alpha_{\mathrm{kg}}$ & verification-feature weights for source prior, cross-source support, and KG grounding \\
$\beta$ & cross-score weight balancing relevance and verification \\
$W_{\mathrm{src}}$ & cap used to normalize cross-source support count \\
$W_1,W_2,K$ & path, verification, and final selection budgets \\
$W_{doc},W_{web}$ & document and web retrieval budgets \\
$\gamma$ & similarity threshold for cross-source support \\
\texttt{\scriptsize DetectEvidenceSubgraph} & construct a $D_{\max}$-hop biomedical KG subgraph around candidate entities and missing hops \\
\texttt{\scriptsize ExpandTypedPaths} & expand operator-compatible KG reasoning paths \\
\texttt{\scriptsize ExpandOneHop} & expand one-hop KG edges from a frontier entity under operator and missing-hop constraints \\
\texttt{\scriptsize FilterOperatorPaths} & keep paths compatible with the predicted reasoning operator and current depth \\
\texttt{\scriptsize RankPrunePaths} & rank and prune KG paths by entity, relation, and operator compatibility \\
\texttt{\scriptsize UpdatePathFrontier} & update the entity frontier after path pruning and operator matching \\
\texttt{\scriptsize AttachKGProvenance} & attach KG source and relation provenance to retained paths \\
\texttt{\scriptsize BuildHopQueries} & build document queries from entities, aliases, relation phrases, KG paths, and missing hops \\
\texttt{\scriptsize RetrieveMultiDoc} & retrieve PubMed-centered documents or snippets from $\mathcal{I}_{doc}$ \\
\texttt{\scriptsize BuildTargetedQueries} & build web queries for missing, uncertain, or freshness-sensitive hops \\
\texttt{\scriptsize SearchWeb} & retrieve clinical-trial or Google Search API results through $\mathcal{W}$ \\
\texttt{\scriptsize ConvertEvidenceGraph} & convert selected text spans into graph-compatible candidate evidence units \\
\hline
\end{tabular}
\end{table}

\clearpage
\section{Workflow Diagram}
\label{appendix:Workflow}
\label{appendix:workflow_diagram}
\begin{figure*}
    \centering
    \includegraphics[width=0.95\linewidth]{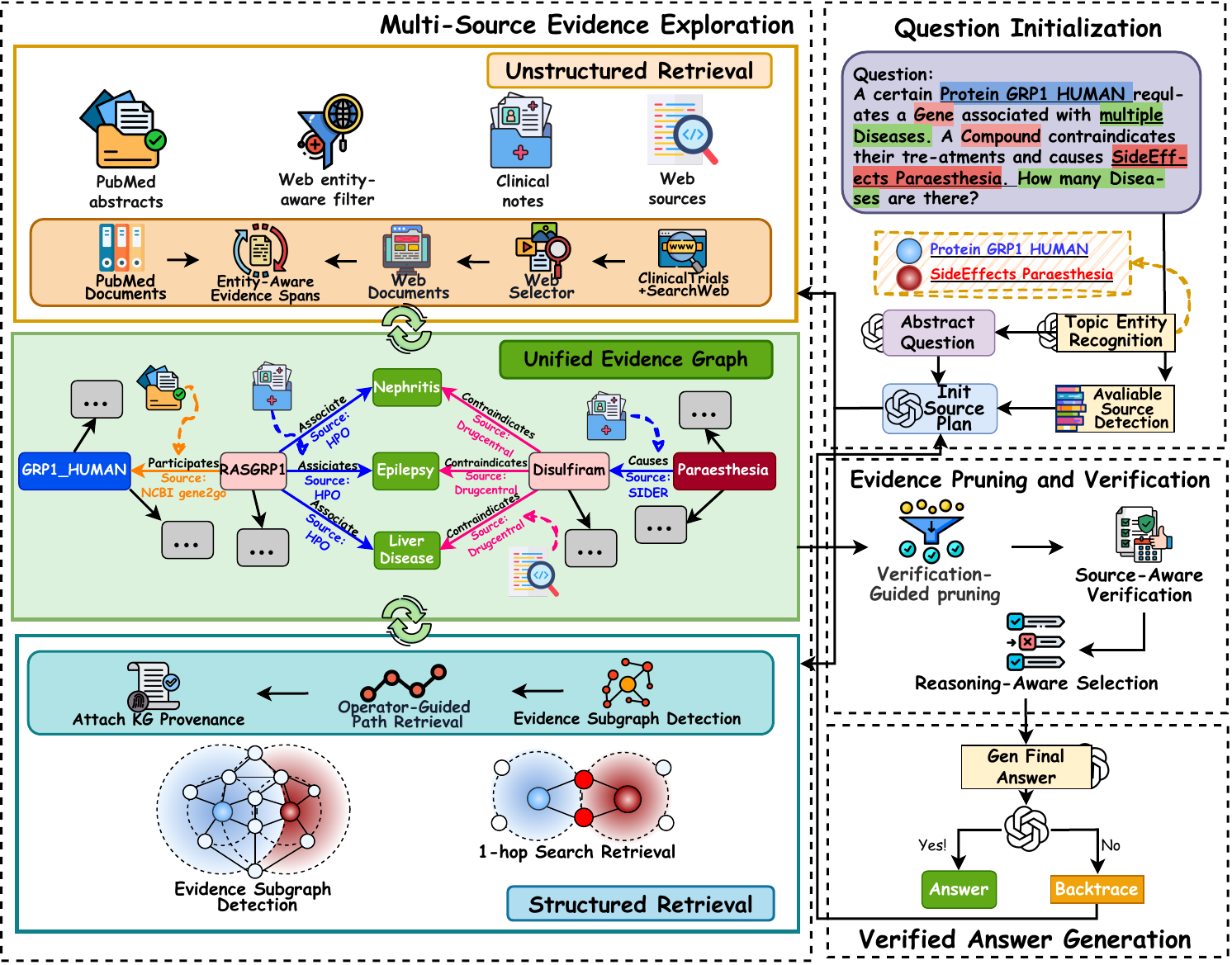}
    \caption{
    Overview of the \bianque architecture.
    Question Initialization abstracts the input question into biomedical entities, answer type, reasoning operator, and a source plan.
    Multi-source Evidence Exploration retrieves structured KG paths and unstructured evidence from PubMed documents, clinical-trial records, and web sources, then converts them into a unified evidence graph.
    Evidence Pruning and Verification ranks evidence by relevance, source-aware verification, and reasoning compatibility.
    Verified Answer Generation produces the final answer from the refined evidence graph, and triggers missing-hop backtracking when the evidence does not satisfy the required reasoning operator.
    }
    \label{fig:method}
\end{figure*}

Figure~\ref{fig:method} provides an overview of the \bianque workflow, which consists of four main components.

\myparagraph{Question Initialization}
The process begins with question initialization, where \bianque abstracts the input question into normalized biomedical mentions, candidate entities, an expected answer type, and a reasoning operator. 
The source planner then determines the active source condition $S_q$, such as KG, document, web, or hybrid evidence, and passes these controls to the retrieval modules.

\myparagraph{Multi-source Evidence Exploration}
During evidence exploration, \bianque retrieves evidence from structured and unstructured sources in parallel.
Structured retrieval detects a biomedical evidence subgraph and performs operator-guided path retrieval over KG relations, producing constrained reasoning traces with provenance.
Unstructured retrieval collects PubMed-centered document evidence, clinical-trial records, and web evidence, then applies entity-aware filtering to retain spans that preserve the required biomedical entities and relation cues.
The selected KG paths and text-derived evidence units are converted into a unified evidence graph, where entities, relations, source provenance, and hop labels are aligned under the same topology.

\myparagraph{Evidence Pruning and Verification}
The unified evidence graph is then passed to evidence pruning and verification.
\bianque first scores evidence by semantic relevance and entity overlap, then applies source-aware verification to favor evidence units supported by reliable or cross-source signals.
A reasoning-aware selector further checks whether the remaining evidence satisfies the required operator, such as shared-neighbor matching, intersection reasoning, path traversal, or counting.

\myparagraph{Verified Answer Generation}
Finally, \bianque generates the answer from the refined evidence graph.
For MCQ, open-answer, and counting questions, the model answers under the verified evidence constraints rather than relying only on parametric knowledge.
If the current graph cannot satisfy the reasoning operator, \bianque backtraces the missing hop and reissues targeted KG, document, or web retrieval before attempting answer generation again.

\section{Additional Experiments}
\label{appendix:additional-experiments}

\subsection{Ablation Study}
\label{appendix:additional-ablation-analysis}

\begin{table*}[t]
  \centering
  \scriptsize
  \renewcommand{\arraystretch}{1.10}
  \caption{Ablation over source-conditioned evidence. We report format-specific automatic scores for each task family and an Overall Avg. over accuracy-style metrics.}
  \label{tab:source-ablation}
  \resizebox{0.8\textwidth}{!}{%
  \begin{tabular}{@{}lcccccccc@{}}
    \toprule
    \multirow{2}{*}{\textbf{Variant}}
    & \multicolumn{2}{c}{\textbf{Entity Pair}}
    & \multicolumn{2}{c}{\textbf{Intersection}}
    & \multicolumn{2}{c}{\textbf{Path Reason.}}
    & \multicolumn{1}{c}{\textbf{Path Counting}}
    & \multirow{2}{*}{\textbf{Overall Avg.}} \\
    \cmidrule(lr){2-3}\cmidrule(lr){4-5}\cmidrule(lr){6-7}\cmidrule(lr){8-8}
    & \textbf{MCQ} & \textbf{Open} & \textbf{MCQ} & \textbf{Open} & \textbf{MCQ} & \textbf{Open} & \textbf{CountEx} & \\
    \midrule
    \multicolumn{9}{@{}l}{\textit{Single-source evidence}} \\
    KG only & 70.3 & 9.5 & 53.6 & 4.8 & 55.4 & 15.2 & 22.5 & 31.7 \\
    Document only & 56.5 & 24.3 & 43.2 & 12.6 & 41.5 & 28.4 & 14.8 & 29.5 \\
    Web only & 57.2 & 23.5 & 41.8 & 11.5 & 42.0 & 30.1 & 15.1 & 29.5 \\
    \midrule
    \multicolumn{9}{@{}l}{\textit{Dual-source evidence}} \\
    Document with web evidence & 60.1 & 31.2 & 46.5 & 16.8 & 44.8 & 34.5 & 17.5 & 33.6 \\
    KG with document evidence & 79.4 & 43.6 & 72.5 & 28.4 & 68.2 & 64.3 & 38.6 & 54.2 \\
    KG with web evidence & 78.8 & 42.1 & 71.3 & 27.5 & 67.5 & 65.1 & 37.2 & 53.3 \\
    \midrule
    \multicolumn{9}{@{}l}{\textit{Full hybrid}} \\
    Full hybrid evidence & 82.5 & 49.2 & 78.4 & 33.5 & 72.8 & 71.5 & 44.2 & 59.5 \\
    \bottomrule
  \end{tabular}
  }
\end{table*}
\myparagraphquestion{How does source-conditioned evidence affect the performance of \bianque}
\label{exp:source-conditioned-evidence}
\label{appendix:source-conditioned-ablation}
To evaluate the impact of different evidence sources, we conduct ablation experiments under single-source, dual-source, and full hybrid evidence settings. As shown in Table~\ref{tab:source-ablation}, KG-only evidence achieves the strongest performance among single-source settings, reaching 31.7 Overall Avg., while document-only and web-only evidence obtain similar scores of 29.5. This shows that structured KG evidence is important for biomedical reasoning, but relying on a single source is still insufficient, especially for open-answer and counting questions where entity aliases, missing edges, and evidence coverage become critical.
Adding textual evidence to KG substantially improves performance. KG with document evidence reaches 54.2 Overall Avg., and KG with web evidence reaches 53.3, both clearly outperforming document with web evidence. This indicates that KG paths provide the main reasoning structure, while documents and web sources complement the KG by providing richer context and missing evidence. Notably, the full hybrid setting achieves the best performance across all task families, with 59.5 Overall Avg., 71.5 on Path Reasoning Open, and 44.2 CountEx. These results demonstrate that \bianque benefits from the complementary strengths of structured and unstructured evidence. By combining KG paths, document evidence, and web evidence in a unified evidence graph, \bianque can better support source-conditioned biomedical reasoning than using any single source or simple dual-source combination.

\begin{table*}[t]
  \centering
  \scriptsize
  \renewcommand{\arraystretch}{1.12}
  \caption{Ablation over evidence alignment. MCQ, Open, and CountEx denote format-specific automatic metrics; verifier columns are reported separately rather than collapsed into one score.}
  \label{tab:alignment-ablation}
  \resizebox{0.78\textwidth}{!}{%
  \begin{tabular}{@{}lcccccccc@{}}
    \toprule
    \textbf{Variant} & \textbf{MCQ} & \textbf{Open} & \textbf{CountEx} & \textbf{Ans.} & \textbf{Evid.} & \textbf{Reason.} & \textbf{Tok.} $\downarrow$ & \textbf{Calls} $\downarrow$ \\
    \midrule
    Full \bianque & 77.9 & 51.3 & 45.4 & 69.8 & 90.4 & 83.6 & 4.2k & 4.5 \\
    w/o entity normalization & 73.5 & 31.5 & 22.1 & 58.2 & 88.5 & 76.4 & 4.1k & 4.5 \\
    w/o unified evidence graph conversion & 68.2 & 44.1 & 35.4 & 61.3 & 75.6 & 62.4 & 5.8k & 3.2 \\
    \bottomrule
  \end{tabular}
  }
\end{table*}
\myparagraphquestion{How does evidence alignment affect the performance of \bianque}
\label{exp:evidence-alignment}
\label{appendix:alignment-verification-ablation}
To evaluate the impact of evidence alignment, we remove two core alignment modules: entity normalization and unified evidence graph conversion. As shown in Table~\ref{tab:alignment-ablation}, removing entity normalization sharply reduces Open accuracy from 51.3 to 31.5 and CountEx from 45.4 to 22.1. This shows that biomedical aliases, identifiers, and surface forms must be normalized before answer generation, especially for open-answer and counting tasks where the model needs to merge equivalent biomedical entities.
Removing unified evidence graph conversion also causes clear performance drops, reducing MCQ from 77.9 to 68.2 and reasoning completeness from 83.6 to 62.4. This indicates that simply retrieving KG paths and textual snippets is not enough. \bianque needs to convert heterogeneous evidence into a shared graph structure so that KG relations, document clues, and web evidence can be jointly organized and reasoned over. Overall, these results demonstrate that evidence alignment is essential for faithful biomedical reasoning, especially when the answer depends on normalized entities and relation-level evidence.

\begin{table*}[t]
  \centering
  \scriptsize
  \renewcommand{\arraystretch}{1.12}
  \caption{Ablation over evidence verification. MCQ, Open, and CountEx denote format-specific automatic metrics; verifier columns are reported separately rather than collapsed into one score.}
  \label{tab:verification-ablation}
  \resizebox{0.78\textwidth}{!}{%
  \begin{tabular}{@{}lcccccccc@{}}
    \toprule
    \textbf{Variant} & \textbf{MCQ} & \textbf{Open} & \textbf{CountEx} & \textbf{Ans.} & \textbf{Evid.} & \textbf{Reason.} & \textbf{Tok.} $\downarrow$ & \textbf{Calls} $\downarrow$ \\
    \midrule
    Full \bianque & 77.9 & 51.3 & 45.4 & 69.8 & 90.4 & 83.6 & 4.2k & 4.5 \\
    w/o cross-source verification & 72.1 & 46.5 & 38.2 & 55.4 & 71.5 & 70.1 & 4.8k & 2.1 \\
    relevance-only pruning & 69.5 & 42.3 & 34.5 & 52.6 & 68.3 & 65.1 & 6.5k & 3.5 \\
    \bottomrule
  \end{tabular}
  }
\end{table*}

\newpage
\myparagraphquestion{How does evidence verification affect the performance of \bianque}
\label{exp:evidence-verification}
To evaluate the impact of evidence verification, we compare the full \bianque with two variants that remove cross-source verification or replace it with relevance-only pruning. As shown in Table~\ref{tab:verification-ablation}, removing cross-source verification reduces answer correctness from 69.8 to 55.4, evidence support from 90.4 to 71.5, and reasoning completeness from 83.6 to 70.1. This suggests that source-aware verification is important for filtering unsupported evidence and preserving reliable reasoning chains.
Relevance-only pruning performs even worse, dropping MCQ, Open, and CountEx to 69.5, 42.3, and 34.5, respectively. It also increases token cost from 4.2k to 6.5k, indicating that relevance alone tends to retain noisy or duplicated evidence. These results show that \bianque does not benefit simply from retrieving more context. Instead, cross-source verification helps select evidence that is not only relevant to the question, but also better supported by source-compatible biomedical evidence.

\subsection{Additional Benchmark Evaluation and Analysis}
\label{appendix:additional-benchmark-analysis}

This section provides additional benchmark-level analyses.
We use these analyses to examine whether \benchmark remains informative across different model backbones, reasoning structures, source conditions, and output formats.
Together, these results help characterize where current systems fail, including multi-clue aggregation, open-ended entity generation, and count-based biomedical reasoning.

\phantomsection\label{appendix:difficulty-source-slice}\label{appendix:reasoning-completeness-hop}
\begin{table*}[t]
  \centering
  \scriptsize
  \renewcommand{\arraystretch}{1.08}
  \caption{Benchmark difficulty analysis by reasoning structure. Best baseline denotes the strongest non-\bianque method under each task-family metric. Gain denotes the absolute improvement of \bianque over the best baseline.}
  \label{tab:benchmark-difficulty-analysis}
  \resizebox{0.92\textwidth}{!}{%
  \begin{tabular}{@{}lccccc@{}}
    \toprule
    \textbf{Task Family} & \textbf{Evidence Motif} & \textbf{Best Baseline} & \textbf{\bianque} & \textbf{Gain} & \textbf{Main Challenge} \\
    \midrule
    Entity Pair & 2-edge shared neighbor & 56.4 & 65.5 & +9.1 & shared-neighbor matching \\
    Intersection & 3-edge intersection graph & 46.0 & 57.4 & +11.4 & multi-clue aggregation \\
    Path Reasoning & 4-hop typed metapath & 62.5 & 70.9 & +8.4 & relation-direction reasoning \\
    Path Counting & 4-hop set aggregation & 32.5 & 45.4 & +12.9 & entity deduplication and counting \\
    \bottomrule
  \end{tabular}
  }
\end{table*}
\myparagraphquestion{How difficult are different reasoning structures in \benchmark}
\label{exp:reasoning-structure-difficulty}
To better understand the benchmark difficulty, we analyze performance by task family rather than only reporting the overall score. As shown in Table~\ref{tab:benchmark-difficulty-analysis}, Path Counting is the most challenging setting for all methods. Even the strongest baseline ToG-2 only achieves 32.5 CountEx, while \bianque improves it to 45.4. This suggests that the main difficulty is not only finding a plausible biomedical entity, but also recovering all valid entities and deduplicating aliases before producing the final count.
Interestingly, task difficulty is not strictly monotonic with hop length. Although Path Reasoning uses 4-hop typed metapaths, \bianque achieves 70.9 on this task, higher than 57.4 on Intersection. This indicates that typed path constraints can provide useful structural guidance, while Intersection requires aggregating multiple biomedical clues into a shared disease node and is more sensitive to entity ambiguity. Overall, these results show that \benchmark evaluates multiple forms of complexity, including entity matching, multi-clue aggregation, relation-direction reasoning, and set-level counting.

\phantomsection\label{appendix:shortcut-analysis}
\begin{figure}[t]
  \centering  \includegraphics[width=0.88\linewidth]{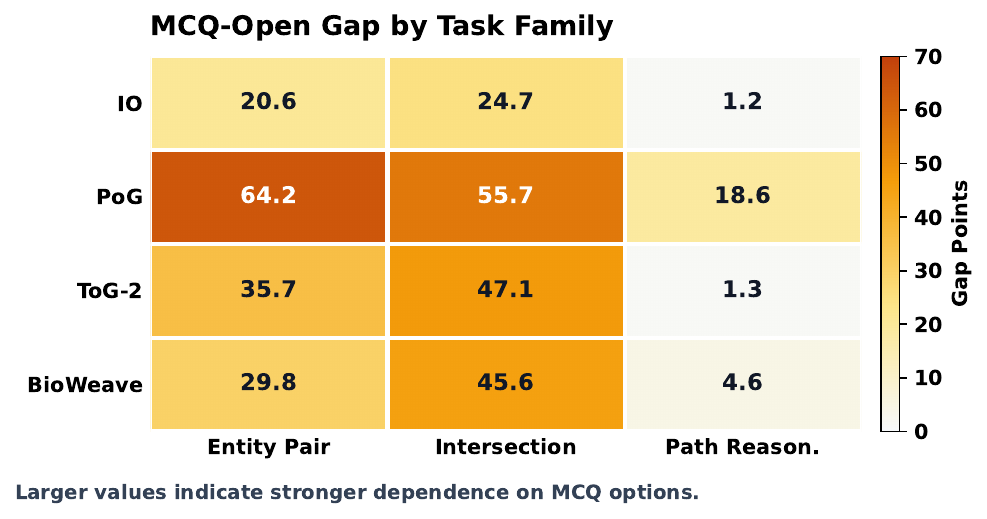}
  \caption{MCQ--Open gap across task families. Larger values indicate stronger dependence on answer options and weaker open-ended entity grounding.}
  \label{fig:mcq-open-gap}
  \vspace{-6mm}
\end{figure}

\myparagraphquestion{Does the multiple-choice format overestimate biomedical reasoning}
\label{exp:mcq-open-gap}
We further analyze the gap between MCQ and open-answer evaluation. As shown in Figure~\ref{fig:mcq-open-gap}, Intersection shows the largest MCQ--Open gap across strong methods. For example, \bianque reaches 80.2 MCQ accuracy but only 34.6 open-answer accuracy on Intersection, yielding a 45.6\% gap. This indicates that multiple-choice options provide strong answer-type and lexical constraints, while open-answer evaluation requires the model to generate the correct normalized biomedical entity without option guidance.
Compared with the baselines, \bianque shows a more controlled MCQ--Open gap. The difference is especially clear against PoG, whose average gap is the largest among all methods, suggesting that PoG benefits substantially from answer-option exposure but struggles more in open-ended entity generation. \bianque still has a non-trivial gap on Intersection, but its average gap is much lower than PoG and slightly lower than ToG-2, while preserving strong open-answer performance. On Path Reasoning, \bianque has only a 4.6\% difference between MCQ and open-answer accuracy, suggesting that typed metapath constraints can better guide answer generation once the reasoning chain is recovered. These results highlight the importance of including both MCQ and open-answer formats in \benchmark: MCQ tests answer recognition, while open-answer evaluation more directly tests entity grounding and normalization.

\phantomsection\label{appendix:grounding-breakdown}
\begin{table}[t]
  \centering
  \scriptsize
  \renewcommand{\arraystretch}{1.08}
  \caption{Grounding--exactness gap of \bianque across task families.}
  \label{tab:grounding-exactness-gap}
  \resizebox{0.92\linewidth}{!}{%
  \begin{tabular}{@{}lccccc@{}}
    \toprule
    \textbf{Task Family} & \textbf{Acc.} & \textbf{Ans.} & \textbf{Evid.} & \textbf{Reason.} & \textbf{Evid.-Acc.} \\
    \midrule
    Entity Pair & 65.5 & 67.5 & 91.0 & 80.0 & 25.5 \\
    Intersection & 57.4 & 60.5 & 86.0 & 76.0 & 28.6 \\
    Path Reasoning & 70.9 & 73.0 & 92.0 & 86.0 & 21.1 \\
    Path Counting & 45.4 & 48.0 & 85.0 & 78.0 & 39.6 \\
    \bottomrule
  \end{tabular}
  }
\end{table}

\myparagraphquestion{Where does evidence grounding still differ from exact answering}
\label{exp:grounding-exactness-gap}
We further compare automatic scores with verifier-based grounding scores. As shown in Table~\ref{tab:grounding-exactness-gap}, \bianque obtains high evidence support across all task families, but exact automatic scores remain lower, especially on Path Counting. For Path Counting, the evidence support score reaches 85.0, while CountEx is 45.4, resulting in a 39.6\% grounding--exactness gap. This suggests that the model often retrieves or constructs useful evidence, but exact set aggregation and numerical output remain difficult.
Intersection also shows a large gap, with 86.0 evidence support but 57.4 automatic score. This indicates that evidence may support the right biomedical neighborhood, while exact disease generation is still affected by alias normalization and multi-clue ambiguity. In contrast, Path Reasoning has the smallest gap among the four tasks, suggesting that typed metapath constraints make the final answer easier to recover once the evidence chain is found. These results show that \benchmark can separate evidence grounding from final answer exactness, which is important for diagnosing biomedical reasoning systems.

\myparagraphquestion{Does \benchmark remain diagnostic across different LLM backbones}
\label{exp:backbone-diagnostic}
To examine whether \benchmark can still distinguish model capability under different LLM backbones, we conduct a full backbone sweep over nine representative models. As shown in Table~\ref{tab:full-backbone-sweep}, stronger backbones improve IO prompting, but the improvement is limited. The average Overall Avg. of IO prompting is only 15.5, and even the strongest IO result reaches only 18.6. This indicates that \benchmark cannot be solved by backbone scaling alone, since biomedical reasoning still requires evidence grounding, entity alignment, and source-aware path construction.

\definecolor{bwrow}{RGB}{232,246,243}
\definecolor{bwgreen}{RGB}{0,105,92}
\phantomsection\label{appendix:additional-model-results}
\begin{table*}[t]
  \centering
  \scriptsize
  \renewcommand{\arraystretch}{0.98}
  \caption{Full backbone sweep on \benchmark. We report task-family average scores for IO, PoG, ToG-2, and \bianque across representative LLM backbones. $\Delta_{\mathrm{IO}}$ compares \bianque against IO on Overall Avg.; best results are in bold and second-best results are underlined.}
  \label{tab:full-backbone-sweep}
  \resizebox{\textwidth}{!}{%
  \begin{tabular}{@{}llcccccc@{}}
    \toprule
    \textbf{LLM Backbone} & \textbf{Method} & \textbf{Entity Pair} & \textbf{Intersection} & \textbf{Path Reason.} & \textbf{Path Counting} & \textbf{Overall Avg.} & \textbf{$\Delta_{\mathrm{IO}}$} \\
    & & \textbf{Avg.} & \textbf{Avg.} & \textbf{Avg.} & \textbf{CountEx} & & \\
    \midrule
    \multirow{4}{*}{Qwen3-4B}
      & IO & 5.9 & 9.5 & 16.0 & 6.0 & 9.3 & -- \\
      & PoG & 26.5 & 18.6 & 27.4 & 9.5 & 20.5 & -- \\
      & ToG-2 & \underline{28.5} & \underline{22.0} & \underline{30.5} & \underline{15.0} & \underline{24.0} & -- \\
      \rowcolor{bwrow}
      & \textbf{\textcolor{bwgreen}{\bianque}} & \textbf{42.5} & \textbf{36.4} & \textbf{46.5} & \textbf{28.6} & \textbf{38.5} & \textbf{+29.2 {\scriptsize(+314\%)}} \\
    \midrule
    \multirow{4}{*}{GPT-4o-mini}
      & IO & 7.8 & 16.7 & 19.8 & 8.5 & 13.2 & -- \\
      & PoG & \underline{38.2} & 25.9 & \underline{44.5} & 15.2 & 30.9 & -- \\
      & ToG-2 & 35.5 & \underline{35.2} & 42.1 & \underline{24.0} & \underline{34.2} & -- \\
      \rowcolor{bwrow}
      & \textbf{\textcolor{bwgreen}{\bianque}} & \textbf{49.8} & \textbf{47.5} & \textbf{55.4} & \textbf{36.5} & \textbf{47.3} & \textbf{+34.1 {\scriptsize(+258\%)}} \\
    \midrule
    \multirow{4}{*}{Claude-3.5-Haiku}
      & IO & 8.9 & 17.7 & 21.4 & 10.2 & 14.5 & -- \\
      & PoG & 34.5 & 28.5 & \underline{47.1} & 18.5 & 32.1 & -- \\
      & ToG-2 & \underline{45.2} & \underline{32.8} & 46.2 & \underline{22.4} & \underline{36.6} & -- \\
      \rowcolor{bwrow}
      & \textbf{\textcolor{bwgreen}{\bianque}} & \textbf{54.5} & \textbf{44.2} & \textbf{58.6} & \textbf{33.8} & \textbf{47.8} & \textbf{+33.3 {\scriptsize(+230\%)}} \\
    \midrule
    \multirow{4}{*}{Qwen3-80B}
      & IO & 10.8 & 19.1 & 23.5 & 11.0 & 16.1 & -- \\
      & PoG & \underline{55.4} & 31.3 & 53.2 & 21.4 & 40.3 & -- \\
      & ToG-2 & 53.4 & \underline{42.8} & \underline{60.5} & \underline{30.5} & \underline{46.8} & -- \\
      \rowcolor{bwrow}
      & \textbf{\textcolor{bwgreen}{\bianque}} & \textbf{63.5} & \textbf{54.2} & \textbf{68.4} & \textbf{43.1} & \textbf{57.3} & \textbf{+41.2 {\scriptsize(+256\%)}} \\
    \midrule
    \multirow{4}{*}{MiMo-V2.5-Pro}
      & IO & 9.4 & 16.2 & 19.5 & 9.5 & 13.7 & -- \\
      & PoG & 39.2 & 28.6 & 49.1 & 18.5 & 33.9 & -- \\
      & ToG-2 & \underline{50.5} & \underline{38.4} & \underline{55.2} & \underline{26.4} & \underline{42.6} & -- \\
      \rowcolor{bwrow}
      & \textbf{\textcolor{bwgreen}{\bianque}} & \textbf{60.2} & \textbf{50.5} & \textbf{62.5} & \textbf{38.5} & \textbf{52.9} & \textbf{+39.2 {\scriptsize(+286\%)}} \\
    \midrule
    \multirow{4}{*}{DeepSeek-V4}
      & IO & 10.8 & 20.5 & 28.2 & 14.5 & 18.5 & -- \\
      & PoG & \underline{56.5} & 32.6 & \textbf{79.9} & 28.2 & \underline{49.3} & -- \\
      & ToG-2 & 53.4 & \underline{41.5} & 58.2 & \underline{34.1} & 46.8 & -- \\
      \rowcolor{bwrow}
      & \textbf{\textcolor{bwgreen}{\bianque}} & \textbf{63.5} & \textbf{58.2} & \underline{76.8} & \textbf{50.7} & \textbf{62.3} & \textbf{+43.8 {\scriptsize(+237\%)}} \\
    \midrule
    \multirow{4}{*}{Gemini-2.5-Pro}
      & IO & 13.3 & 22.2 & 25.4 & 13.5 & 18.6 & -- \\
      & PoG & \underline{60.5} & 34.2 & 54.0 & 22.5 & 42.8 & -- \\
      & ToG-2 & 59.2 & \underline{50.4} & \underline{60.5} & \underline{35.9} & \underline{51.5} & -- \\
      \rowcolor{bwrow}
      & \textbf{\textcolor{bwgreen}{\bianque}} & \textbf{67.2} & \textbf{60.5} & \textbf{62.4} & \textbf{44.3} & \textbf{58.6} & \textbf{+40.0 {\scriptsize(+215\%)}} \\
    \midrule
    \multirow{4}{*}{Claude-3.7-Sonnet}
      & IO & 12.4 & 21.1 & 25.5 & 12.0 & 17.8 & -- \\
      & PoG & 43.5 & 34.1 & 55.2 & 22.8 & 38.9 & -- \\
      & ToG-2 & \underline{57.5} & \underline{46.8} & \underline{62.4} & \underline{34.1} & \underline{50.2} & -- \\
      \rowcolor{bwrow}
      & \textbf{\textcolor{bwgreen}{\bianque}} & \textbf{68.4} & \textbf{55.6} & \textbf{72.5} & \textbf{47.1} & \textbf{60.9} & \textbf{+43.1 {\scriptsize(+242\%)}} \\
    \midrule
    \multirow{4}{*}{GPT-4-Turbo}
      & IO & 12.7 & 21.6 & 26.1 & 12.5 & 18.2 & -- \\
      & PoG & 44.4 & 33.4 & 56.1 & 23.4 & 39.3 & -- \\
      & ToG-2 & \underline{56.4} & \underline{46.0} & \underline{62.5} & \underline{32.5} & \underline{49.3} & -- \\
      \rowcolor{bwrow}
      & \textbf{\textcolor{bwgreen}{\bianque}} & \textbf{65.5} & \textbf{57.4} & \textbf{70.9} & \textbf{45.4} & \textbf{59.8} & \textbf{+41.6 {\scriptsize(+229\%)}} \\
    \bottomrule
  \end{tabular}
  }
\end{table*}


Retrieval-based methods improve over IO prompting, but the gap between different reasoning frameworks remains clear across all backbones. On average, \bianque reaches 53.9 Overall Avg., compared with 36.4 for PoG and 42.5 for ToG-2. This corresponds to average improvements of 17.5\% over PoG and 11.5\% over ToG-2. The improvement is also consistent across task families, with average gains over ToG-2 of 10.6\% on Entity Pair, 12.1\% on Intersection, 10.7\% on Path Reasoning, and 12.6\% on Path Counting. These results show that \benchmark can distinguish not only stronger and weaker LLMs, but also different evidence-reasoning designs under the same backbone.
Notably, even with \bianque, Path Counting remains the hardest task across backbones, with an average CountEx of 40.9, while Path Reasoning reaches 63.8 on average. This further supports our previous observation that set-level aggregation and entity deduplication introduce a different type of difficulty from typed path traversal. Overall, the backbone sweep shows that \benchmark is not saturated by current LLMs and remains diagnostic across model scales, retrieval strategies, and reasoning structures.

\subsection{Faithfulness Analysis}
\label{appendix:faithfulness-analysis}
\label{appendix:evidence-faithfulness}

This group focuses on whether the model's answer is supported by the active evidence and whether the explored evidence topology matches the intended answer-centered path.

\begin{figure}[t]
    \centering
    \includegraphics[width=1\linewidth]{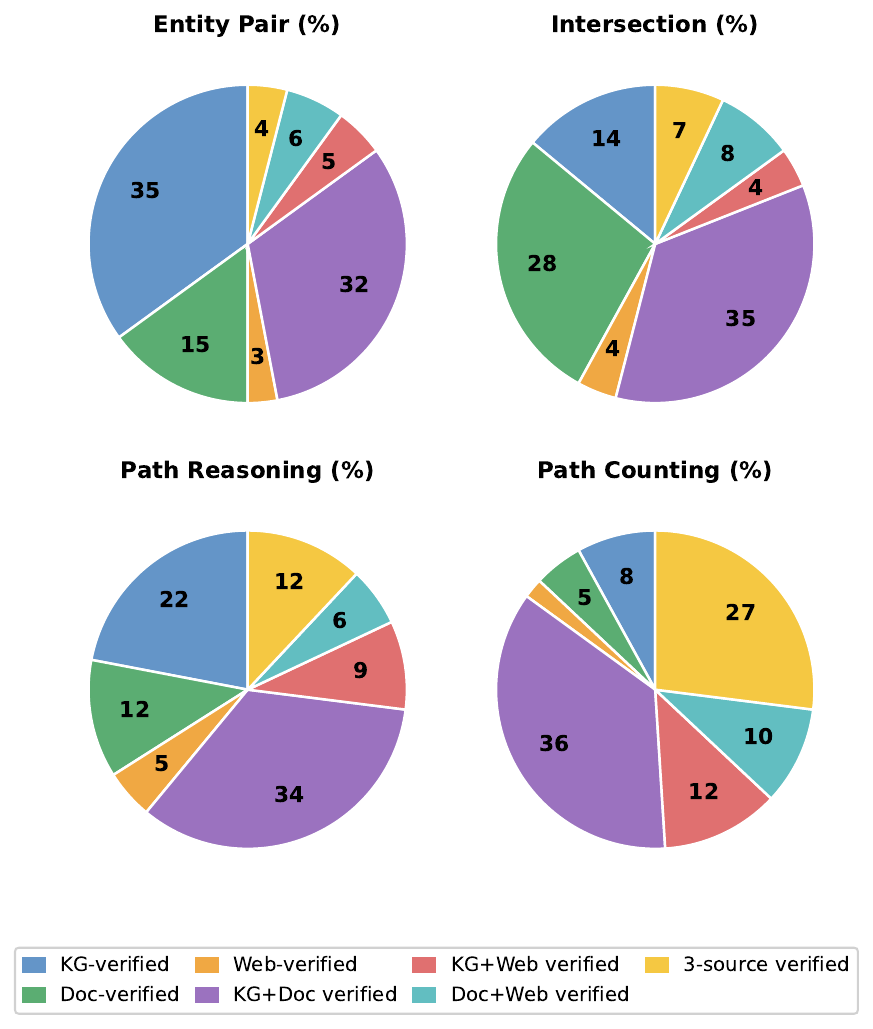}
    \caption{Evidence source distribution of correct answers across task families.}\label{fig:evidence-source-distribution}
\end{figure}

\myparagraph{Evidence sources of correct answers}
\label{exp:evidence-source-distribution}
We analyze the evidence sources supporting correct answers across four task families to assess how \bianque uses heterogeneous biomedical knowledge, as shown in Figure~\ref{fig:evidence-source-distribution}. Specifically, all correct answers are classified based on their verification sources: KG-verified, document-verified, web-verified, as well as cross-source combinations such as KG-Doc, KG-Web, Doc-Web, and 3-source verified answers. The analysis shows a clear shift from single-source support to multi-source verification as the reasoning structure becomes more complex.

For Entity Pair, KG evidence dominates the support distribution. KG-verified answers account for 35\%, the largest single-source category, and KG-involved evidence accounts for 76\% in total. This indicates structured KG relations are highly effective for shallow shared-entity matching, where clean graph edges can often directly support the answer. In contrast, Intersection shows a stronger dependence on document evidence: document-verified answers increase to 28\%, and document-involved support reaches 78\%. This shows biomedical documents often describe diseases together with their related genes, symptoms, or anatomical terms, making documents useful for aggregating multiple clues into the same target entity.

For deeper reasoning tasks, cross-source verification becomes increasingly important. In Path Reasoning, KG-Doc verified answers account for 34\%, showing that KG structure and textual evidence need to be aligned to complete longer reasoning chains. In Path Counting, single-source evidence is greatly reduced: KG-only, document-only, and web-only support account for only 8\%, 5\%, and 2\%, respectively. Instead, multi-source verification accounts for 85\% of correct answers, and 3-source verification reaches 27\%, the highest among all task families. This demonstrates that counting requires not only finding valid paths, but also resolving aliases, removing duplicates, and verifying entity sets across different sources.

Overall, these results show that \bianque does not use all evidence sources uniformly. KG evidence provides reliable structure for shallow entity matching, document evidence strengthens multi-clue aggregation, and web evidence becomes more useful as complementary support in harder path and counting tasks. The increasing ratio of multi-source verification from Entity Pair to Path Counting further confirms the need for unified evidence graph construction and source-aware verification in complex biomedical reasoning.

\begin{figure}[t]
    \centering
    \includegraphics[width=0.86\linewidth]{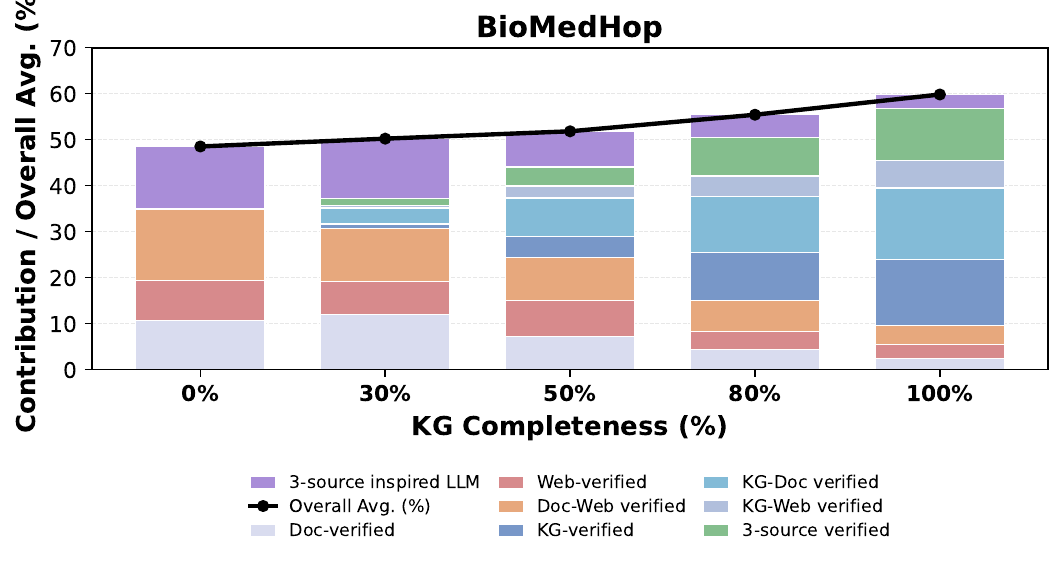}
    \caption{Robustness against incomplete KGs and non-linear source routing. Stacked bars decompose the achieved Overall Avg. by the answer sources, while the black line reports Overall Avg. under different KG completeness levels.}
    \label{fig:kg-completeness}
\end{figure}
\myparagraph{Robustness against incomplete KGs and non-linear source routing}
\label{sec:exp:incomplete_kg}
To measure how \bianque handles incomplete biomedical KGs, we simulate KG incompleteness by randomly retaining different proportions of KG triples (0\%, 30\%, 50\%, 80\%, and 100\%) while keeping the document and web evidence collections intact. 
Figure~\ref{fig:kg-completeness} reports Overall Avg. together with a proportional decomposition of the answer sources; the stacked bars are scaled by the accuracy at each KG completeness level.
The results show graceful degradation rather than collapse. 
Even at 0\% KG completeness, \bianque obtains 48.5\% Overall Avg., supported by document, web, Doc-Web, and 3-source inspired LLM evidence.
As the KG becomes more complete, performance rises to 59.8\%, and the evidence composition shifts toward KG-involved verification.
This shift is not strictly linear.
At 30\% KG completeness, fragmented KG edges do not yet form reliable multi-hop bridges, so \bianque still routes many answers to Doc-verified and 3-source inspired LLM evidence.
Once KG completeness reaches 50--80\%, KG-verified and 3-source verified answers increase more clearly, indicating that restored graph connectivity makes structured verification more useful.
Overall, the trend suggests that \bianque does not simply concatenate all available sources: source-aware verification routes reasoning among text-only evidence, LLM-inspired cross-source cues, and strict graph-supported paths according to the structural integrity of the available KG.

   \phantomsection\label{appendix:evidence-path-overlap}
   \myparagraph{Overlap between \bianque's retrieved KG paths and ground-truth paths}
  We analyze correctly answered samples from \bianque to examine how well its
  retrieved KG subgraph covers the ground-truth biomedical paths.
  For each question, the overlap ratio is defined as the proportion of ground-truth
  path edges present in the retrieved SPOKE subgraph:
  \[
    \text{Ratio}(P) = \frac{|\text{Edges}(P) \cap \text{Edges}(P_G)|}{|\text{Edges}(P_G)|}
  \]
  where $P_G$ denotes the ground-truth KG path and $P$ is the entity-level subgraph
  retrieved by \bianque's BiBFS.

\begin{figure}[t]
\includegraphics[width=1\linewidth]{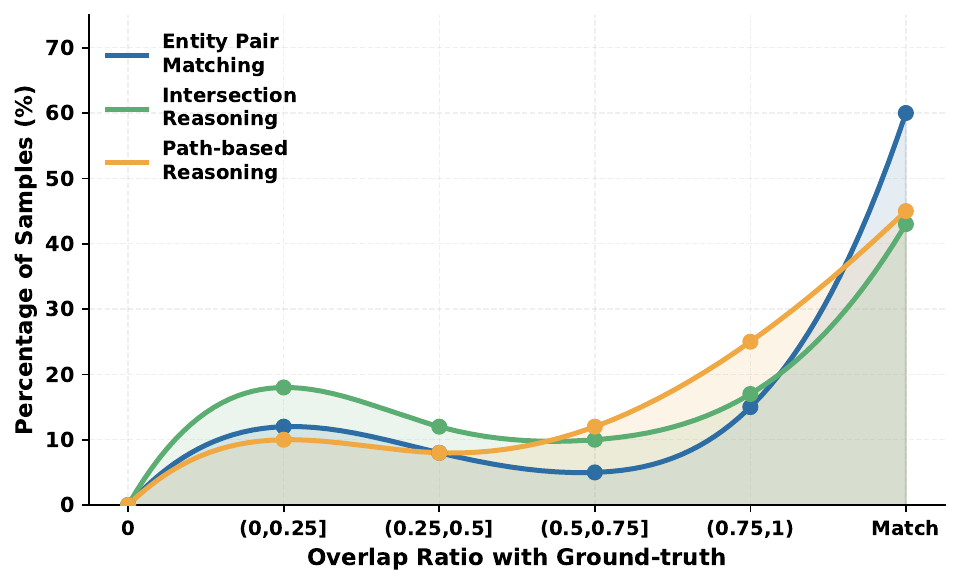}      
\caption{ The path overlap ratio of \bianque.}\label{fig:exp:overlap}
    
\end{figure}

  Figure~\ref{fig:exp:overlap} shows the overlap ratio distributions across three
  task families.
  For Entity Pair Matching, \bianque achieves full coverage (ratio~=~1.0) in
  \textbf{65\%} of correctly answered cases, indicating that the two-entity star
  structure is reliably recovered by KG retrieval.
  For Path-based Reasoning, full chain coverage reaches \textbf{42\%}, with a
  further 32\% showing partial coverage (ratio~$\in$~(0.5,~1.0)), reflecting the
  greater difficulty of recovering a complete 4-hop path from SPOKE.
  Intersection Reasoning shows the lowest full coverage at \textbf{18\%}, with most
  correct answers distributed across partial-coverage bins (76\% in (0,~1.0)),
  suggesting that multi-clue aggregation tasks rely more heavily on document and
  web evidence to bridge the missing KG edges.
  Across all three families, non-zero overlap accounts for over 94\% of Entity Pair,
  77\% of Path Reasoning, and 94\% of Intersection correct answers, confirming that
  \bianque's KG retrieval provides meaningful structural grounding even when full
  path coverage is not achieved.

    

\begin{figure}[t]
    \centering
\includegraphics[width=0.8\linewidth]{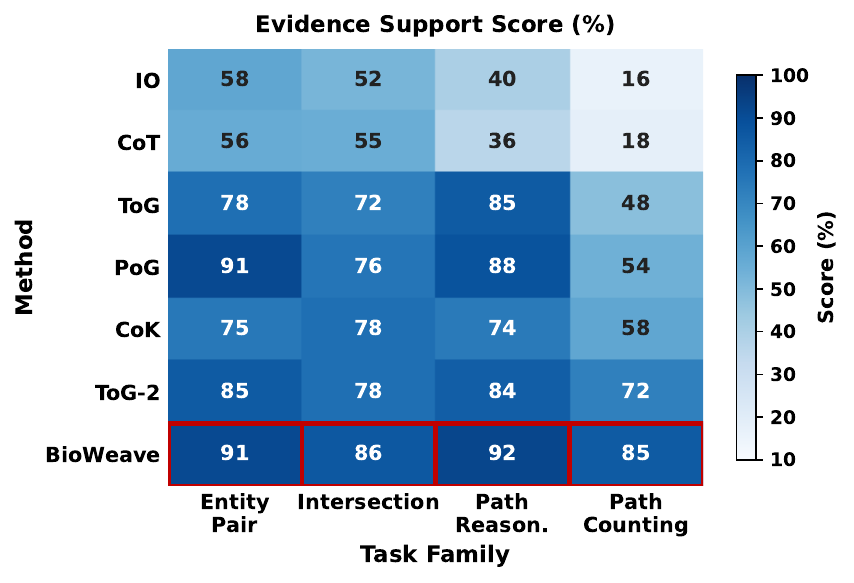}      
\caption{Evidence support scores across task families. Darker cells indicate higher verifier-rated evidence support.}
\label{fig:evidence-support-heatmap}
    
\end{figure}

\phantomsection\label{appendix:evidence-score-calibration}
\myparagraph{Evidence support across task families}
\label{exp:evidence-support-heatmap}
We further analyze the evidence support score across task families to evaluate whether each method can provide verifiable evidence for its answers. As shown in Figure~\ref{fig:evidence-support-heatmap}, LLM-only methods obtain low evidence support, especially on Path Reasoning and Path Counting. For example, IO only reaches 40\% evidence support on Path Reasoning and 16\% on Path Counting. This indicates that parametric knowledge alone often fails to provide sufficient source-grounded evidence for complex biomedical reasoning.
Retrieval-based methods improve evidence support, but their strengths vary by evidence type. KG-based methods perform well on entity-centric and path-based tasks. PoG reaches 91\% on Entity Pair and 88\% on Path Reasoning, showing that KG paths are useful for structured relation traversal. However, its evidence support drops to 54\% on Path Counting, suggesting that KG-only evidence is less reliable when the task requires complete entity-set recovery and deduplication. Hybrid methods further improve support on harder tasks, with ToG-2 reaching 72\% on Path Counting, but still showing weaker support than \bianque across all task families.

\bianque achieves the most stable evidence support, ranging from 85\% to 92\% across the four task families. Compared with ToG-2, \bianque improves evidence support by 6\% on Entity Pair, 8\% on Intersection, 8\% on Path Reasoning, and 13\% on Path Counting. This consistent improvement shows that \bianque can construct better supported evidence chains rather than only selecting plausible answers. The largest gain appears on Path Counting, confirming that unified evidence graph construction and source-aware verification are especially useful when the model must aggregate and verify multiple biomedical entities.

\subsection{Error and Judge Analysis}
\label{appendix:error-judge-analysis}
\label{appendix:exact-judge-disagreement}

This group studies how automatic scoring, LLM-based verification, and error categorization interact.
It is intended to make the evaluation protocol auditable, especially for open-ended biomedical answers.

\begin{figure}[t]
    \centering
    \includegraphics[width=1\linewidth]{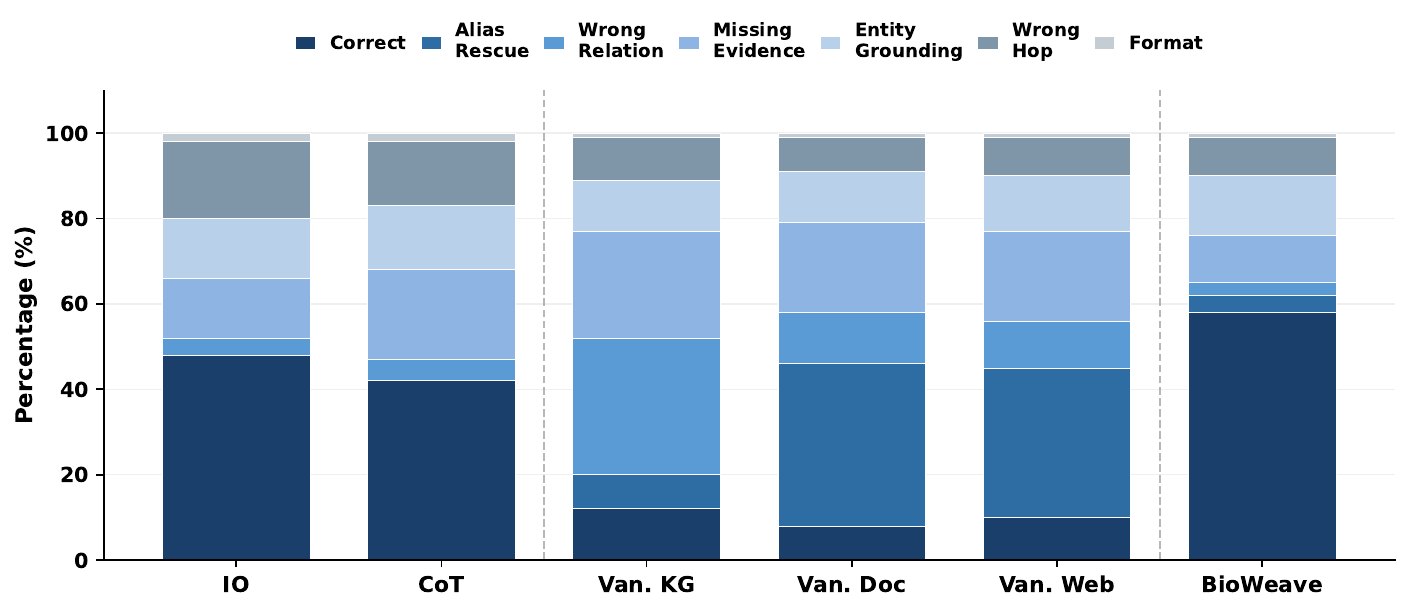}
    \caption{Distribution of error categories across different methods on BioMedHop. }
    \vspace{-3mm}
    \label{fig:exp:error_analysis}
\end{figure}

\myparagraph{Error Analysis}
\label{sec:exp:error_analysis}
To better understand the failure modes of different reasoning paradigms, we conduct a fine-grained error analysis on incorrect predictions. Based on the LLM-verifier error taxonomy in Table~\ref{tab:judge-error-types}, each failure is classified into one of seven categories. Figure~\ref{fig:exp:error_analysis} reports the normalized error distribution within the incorrect answers of each method.

The results show that different reasoning paradigms fail in different ways. LLM-only methods mainly suffer from unsupported generation: {unsupported answer} accounts for 48\% of IO errors and 42\% of CoT errors. This indicates that parametric knowledge alone often produces plausible but unsupported biomedical claims. Vanilla document and web RAG reduce this problem, but introduce a strong coverage bottleneck: {missing evidence} accounts for 38\% and 35\% of their errors, respectively, showing that isolated text retrieval often cannot recover a complete multi-hop evidence chain. In contrast, vanilla KG retrieval has fewer missing-evidence errors, but suffers from {entity grounding} and {counting/deduplication} errors, which account for 32\% and 25\% of its failures. This reflects the difficulty of aligning ontology-specific KG entities with biomedical aliases and avoiding duplicated entities in counting tasks.

Compared with these baselines, \bianque reduces the main structural errors. \text{Missing evidence} drops to 4\%, \text{entity grounding} to 3\%, wrong hop/relation to 9\%, and counting/deduplication to 11\%. In total, the proportion of structural errors decreases from 75--79\% in single-source RAG methods to 27\% in \bianque. The remaining errors are mainly \text{unsupported answer} and \text{distractor confusion}. Since the figure is normalized over incorrect answers, this does not mean \bianque produces more unsupported answers in absolute terms; rather, after retrieval, grounding, and deduplication errors are largely reduced, the residual failures are concentrated in harder ambiguous or weakly supported cases. Overall, the analysis shows that \bianque improves the reliability of biomedical reasoning by aligning and verifying heterogeneous evidence before answer generation.
\begin{table}[t]
  \centering
  \small
  \renewcommand{\arraystretch}{1.12}
  \caption{Verifier error taxonomy.}
  \label{tab:judge-error-types}
  \resizebox{0.9\columnwidth}{!}{%
  \begin{tabular}{@{}ll@{}}
    \toprule
    \textbf{Error Type} & \textbf{Interpretation} \\
    \midrule
    entity grounding & wrong entity, alias, or identifier resolution \\
    missing evidence & answer may be correct but evidence is insufficient \\
    wrong hop/relation & relation direction or metapath step is wrong \\
    distractor confusion & answer selects a type-compatible distractor \\
    counting/deduplication & count is affected by duplicates or missing items \\
    unsupported answer & answer is not supported by active evidence \\
    format error & output violates required answer format \\
    \bottomrule
  \end{tabular}
  }
\end{table}

\myparagraph{Judge reliability analysis}
\label{appendix:judge-reliability}
\label{appendix:evaluation-agreement}
To evaluate the reliability of the LLM-based verifier, we conduct a pairwise judge agreement analysis on 500 randomly sampled predictions. We compare GPT-4o-mini with two independent judge models, Qwen3-80B (Qwen/Qwen3-Next-80B-A3B-Instruct) and DeepSeek-V4, across four verifier dimensions: semantic answer correctness, evidence support, reasoning completeness, and error type assignment. As shown in Table~\ref{tab:judge-reliability}, the judges show high agreement on answer correctness, with 86.4\% agreement between GPT-4o-mini and Qwen3-80B and 89.2\% agreement between GPT-4o-mini and DeepSeek-V4. This suggests that semantic answer correctness is relatively stable across judge models.

Agreement decreases as the evaluation dimension becomes more fine-grained. Evidence support agreement remains moderate to high, reaching 74.2\% and 78.6\%, while reasoning completeness agreement drops to 68.8\% and 72.4\%. This is expected because evidence and reasoning judgments require checking whether the provided evidence covers the necessary biomedical entities, relations, and intermediate hops, which is more complex than judging answer equivalence alone. Error type agreement is the lowest, with 58.6\% and 63.5\% agreement, because error categories such as {missing evidence}, {wrong hop/relation}, and {entity grounding} can overlap in difficult cases.

Overall, the judge agreement analysis shows that the verifier is reliable for answer-level and evidence-level evaluation, while fine-grained error diagnosis is more challenging and should be interpreted as a diagnostic signal rather than an exact label. The higher agreement between GPT-4o-mini and DeepSeek-V4 across all dimensions further indicates that the observed trends are not specific to one judge pair.
\begin{table}[t]
  \centering
  \scriptsize
  \renewcommand{\arraystretch}{1.10}
  \caption{Pairwise judge agreement on 500 sampled predictions.}
  \label{tab:judge-reliability}
  \resizebox{0.92\linewidth}{!}{%
  \begin{tabular}{@{}lccccc@{}}
    \toprule
    \textbf{Judge Pair} & \textbf{N} & \textbf{Answer} & \textbf{Evidence} & \textbf{Reasoning} & \textbf{Error Type} \\
    \midrule
    \makecell[l]{GPT-4o-mini vs. Qwen3-80B\\{\scriptsize(Qwen/Qwen3-Next-80B-A3B-Instruct)}} & 500 & 86.4\% & 74.2\% & 68.8\% & 58.6\% \\
    GPT-4o-mini vs. DeepSeek-V4 & 500 & 89.2\% & 78.6\% & 72.4\% & 63.5\% \\
    \bottomrule
  \end{tabular}
  }
\end{table}

\subsection{Efficiency Analysis}
\label{appendix:effiency_analysis}
\label{appendix:cost-analysis}
\label{appendix:efficiency-cost}

\begin{figure}[t]
    \centering
    \includegraphics[width=1\linewidth]{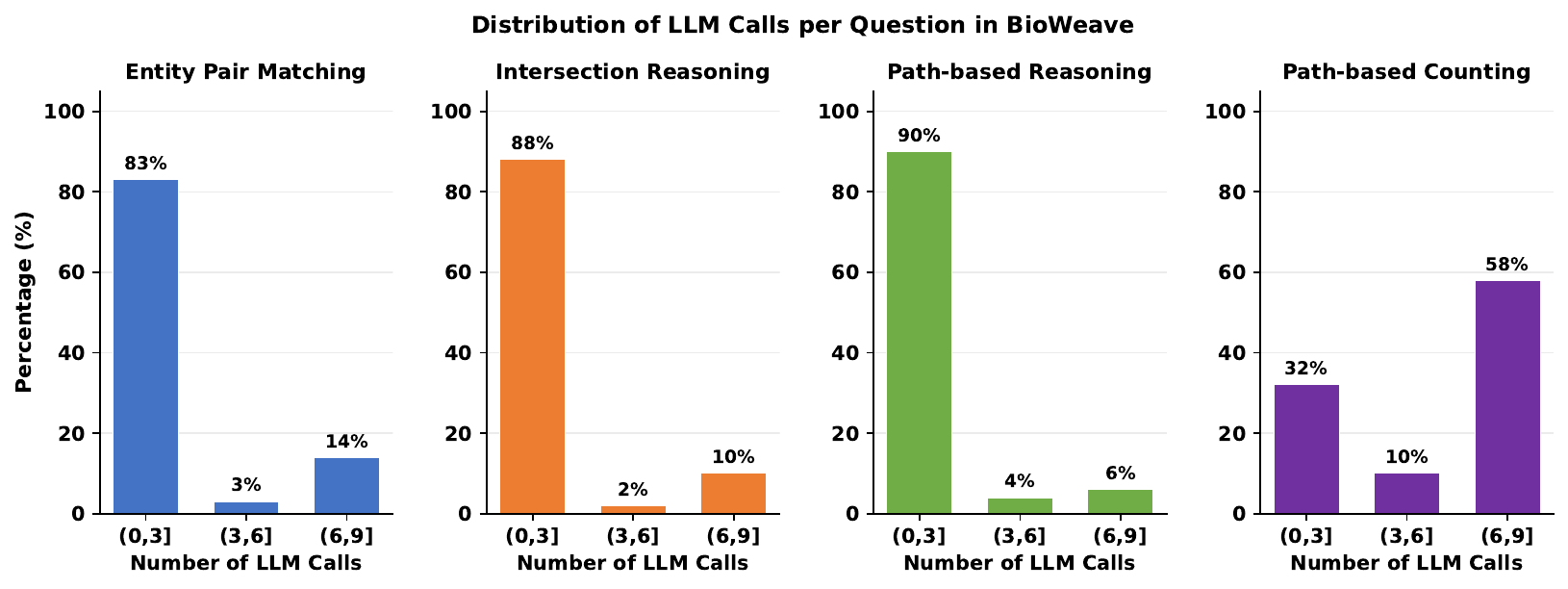}
\caption{Distribution of LLM calls per question in \bianque across four task families. Most entity-answer and path-reasoning questions are completed within three calls, while Path-based Counting requires more calls due to entity-set aggregation and deduplication.}
\label{fig:llm-call-distribution}
\end{figure}

\myparagraph{LLM calls cost analysis}
\label{exp:llm-calls-cost}
To evaluate the inference cost of \bianque, we analyze the distribution of LLM calls per question across the four task families, as shown in Figure~\ref{fig:llm-call-distribution}. The results show that most non-counting questions can be answered with a small number of LLM calls. Specifically, 83\% of Entity Pair questions, 88\% of Intersection questions, and 90\% of Path-based Reasoning questions are completed within three LLM calls. This indicates that when the required evidence path can be found and verified early, \bianque can stop the reasoning process without repeatedly invoking the LLM.
The main exception is Path-based Counting. Only 32\% of counting questions are completed within three calls, while 58\% require six to nine calls. This is expected because counting requires the model to recover all valid answer entities, merge aliases, remove duplicates, and verify whether each entity satisfies the required path pattern. Compared with single-answer tasks, this process needs more evidence checking and often triggers additional refinement steps. Even so, all task families are completed within nine LLM calls, showing that the iterative evidence construction process remains bounded.

Overall, the LLM-call distribution suggests that \bianque is efficient for standard entity-answer and path-reasoning questions, while allocating more computation to harder counting cases. This behavior is desirable for \benchmark, since the most expensive cases are also those that require more complete evidence aggregation and verification.

\myparagraph{Efficiency analysis on \benchmark}
\label{exp:efficiency-analysis-biomedhop}
We compare the efficiency and effectiveness of different reasoning methods on \benchmark by reporting average total processing time, API calls per question, and Overall Avg., as shown in Table~\ref{tab:efficiency_biomedhop}. Among all methods, \bianque achieves the highest Overall Avg. of 59.8 while requiring only 4.5 API calls per question, the lowest among all compared methods. Compared with ToG-2, \bianque improves Overall Avg. by 10.5\% and further reduces API calls from 5.2 to 4.5, while only increasing average processing time from 26.5s to 29.8s. This shows that \bianque gains substantial answer quality without introducing heavy LLM-call overhead.

Compared with CoK, \bianque improves accuracy by 17.7\%, reduces API calls by 5.3, and also lowers average processing time by 4.4s. Compared with ToG, \bianque is both more accurate and more efficient, improving accuracy by 26.3\% while reducing average processing time by 28.6s and API calls by 10.0. These results indicate that the proposed evidence graph construction and verification strategy does not simply increase computation; instead, it helps prune noisy evidence and reduce unnecessary LLM interactions.

Overall, \bianque achieves the best balance between efficiency and answer quality. Although ToG-2 has the lowest processing time, its accuracy is much lower than \bianque. In contrast, \bianque maintains a moderate runtime, uses the fewest API calls, and achieves the strongest reasoning performance, demonstrating that source-aware evidence integration can improve biomedical reasoning while controlling inference cost.

\begin{table}[t]
\centering
\scriptsize
\renewcommand{\arraystretch}{1.10}
\caption{Efficiency analysis of different methods on \benchmark. Average total time is measured in seconds per query.}
\label{tab:efficiency_biomedhop}
\resizebox{0.92\linewidth}{!}{
\begin{tabular}{@{}lccc@{}}
\toprule
\textbf{Method} & \textbf{Avg. Time (s)} $\downarrow$ & \textbf{API Calls} $\downarrow$ & \textbf{Overall Acc.} $\uparrow$ \\
\midrule
ToG & 58.4 & 14.5 & 33.5 \\
CoK & 34.2 & 9.8 & 42.1 \\
ToG-2 & \textbf{26.5} & 5.2 & 49.3 \\
\rowcolor{gray!10}
\textbf{\bianque} & 29.8 & \textbf{4.5} & \textbf{59.8} \\
\bottomrule
\end{tabular}
}
\end{table}

\clearpage
\section{Benchmark Details}
\label{sec:appendix}



\subsection{Task Construction Details}
\label{appendix:benchmark-task-construction}

This appendix provides additional details for the benchmark construction pipeline described in Section~\ref{sec:benchmark-construction}. 
Each \benchmark{} instance is generated from an answer-determining biomedical graph motif.
The motif defines the latent reasoning target, while the source-conditioned evidence layer is attached afterward.
This separation allows the same gold answer structure to be evaluated under KG-only, document-only, web-only, and hybrid evidence settings.

\myparagraph{Construction pipeline}
Each instance is constructed in four steps.
First, we instantiate a typed biomedical graph pattern $\mathcal{G}^{gold}_q$, which contains either a unique answer node or a closed answer set.
Second, we validate entity types, relation directions, answer uniqueness, and answer-set closure for counting tasks.
Third, we render the same latent graph into MCQ and open-answer formats.
For MCQ questions, distractors are sampled from biomedical entities with compatible types and comparable lexical length; for open-answer questions, the gold answer is evaluated through normalized names, aliases, and identifiers.
Finally, the source-conditioned evidence layer is attached to the fixed graph; source condition changes only the exposed evidence channel rather than gold answer structure.


\begin{table}[t]
\centering
\small
\renewcommand{\arraystretch}{1.10}
\caption{Entity pair type distribution in Entity Pair Matching.}
\label{tab:entity-pair-statistics}
\begin{tabular}{lc}
\hline
\textbf{Entity Pair Type} & \textbf{Percentage} \\ \hline
Disease Pair - Gene & 13.4\% \\
Disease Pair - Compound & 13.4\% \\
Disease Pair - Symptom & 12.8\% \\
Disease Pair - Anatomy & 13.6\% \\
Compound Pair - Side Effect & 13.6\% \\
Compound Pair - Disease & 13.4\% \\
Compound Pair - Food & 4.4\% \\
Protein Pair - Protein Domain & 4.6\% \\
Protein Pair - Cell Type & 10.9\% \\
\textbf{Total} & \textbf{100.0\%} \\ \hline
\end{tabular}
\end{table}

\myparagraph{Task families}
We instantiate four task from the above construction pipeline, covering shared-neighbor matching, multi-clue aggregation, typed path traversal, and set-level counting. Only the first three families are rendered as both multiple-choice and open-answer questions; Path-based Counting is evaluated with numeric count answers.

\smallskip
\noindent\raisebox{0.2ex}{\scriptsize$\bullet$}\enspace
\myparagraph{Entity Pair Matching}
Entity Pair Matching asks the model to identify the shared-neighbor biomedical node jointly connected to two anchor entities. For example, two diseases may share a gene, compound, symptom, or anatomical entity, and the answer is the shared neighbor rather than either anchor. Entity pairs and gold shared-neighbor nodes are retrieved from the Monarch Initiative through the BioLink API. The task covers nine relation families, including disease--gene, disease--compound, compound--side-effect, compound--food, protein--protein-domain, and protein--cell-type associations. This task provides a controlled test of shallow graph-grounded shared-neighbor matching. Table~\ref{tab:entity-pair-statistics} summarizes the entity-pair distribution.

\begin{figure}[h]
  \centering
  \includegraphics[width=\columnwidth]{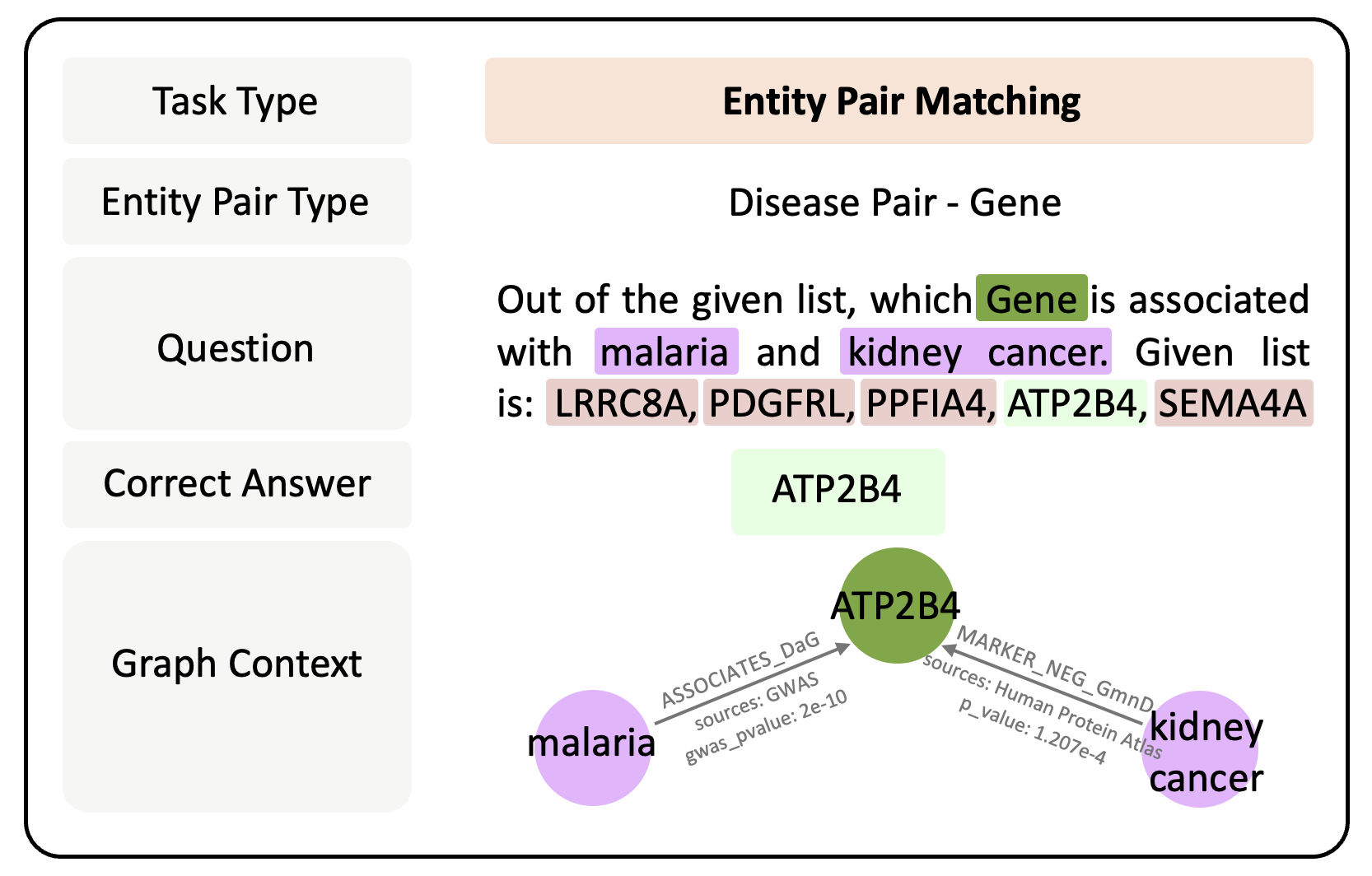}
  \caption{Example from Entity Pair Matching. The model must identify the shared biomedical entity connecting two query entities.}
  \label{fig:entity-pair-example}
\end{figure}

\myparagraphunderline{Example}
Figure~\ref{fig:entity-pair-example} shows a shallow multi-hop question where two query entities share a target biomedical entity.
The model must ground both query entities and identify the shared neighbor, rather than selecting an answer from lexical similarity alone.

\smallskip
\noindent\raisebox{0.2ex}{\scriptsize$\bullet$}\enspace
\myparagraph{Intersection Reasoning}
Intersection Reasoning asks for the disease that satisfies several typed biomedical clues. Each question provides three neighboring entities, such as genes, symptoms, compounds, anatomy terms, organisms, proteins, or locations, and the target answer is the disease connected to all of them. We collect disease-centered neighborhoods and group neighbors into seven entity types: \textit{Symptom, Gene, Compound, Anatomy, Organism, Protein, Location}. Each rendered question forms an answer-centered \emph{3-edge intersection graph} rather than a linear multi-hop chain. This task tests conjunctive entity aggregation, since the model must identify the disease supported by the intersection of multiple biomedical conditions.

\begin{figure}[h]
  \centering
  \includegraphics[width=\columnwidth]{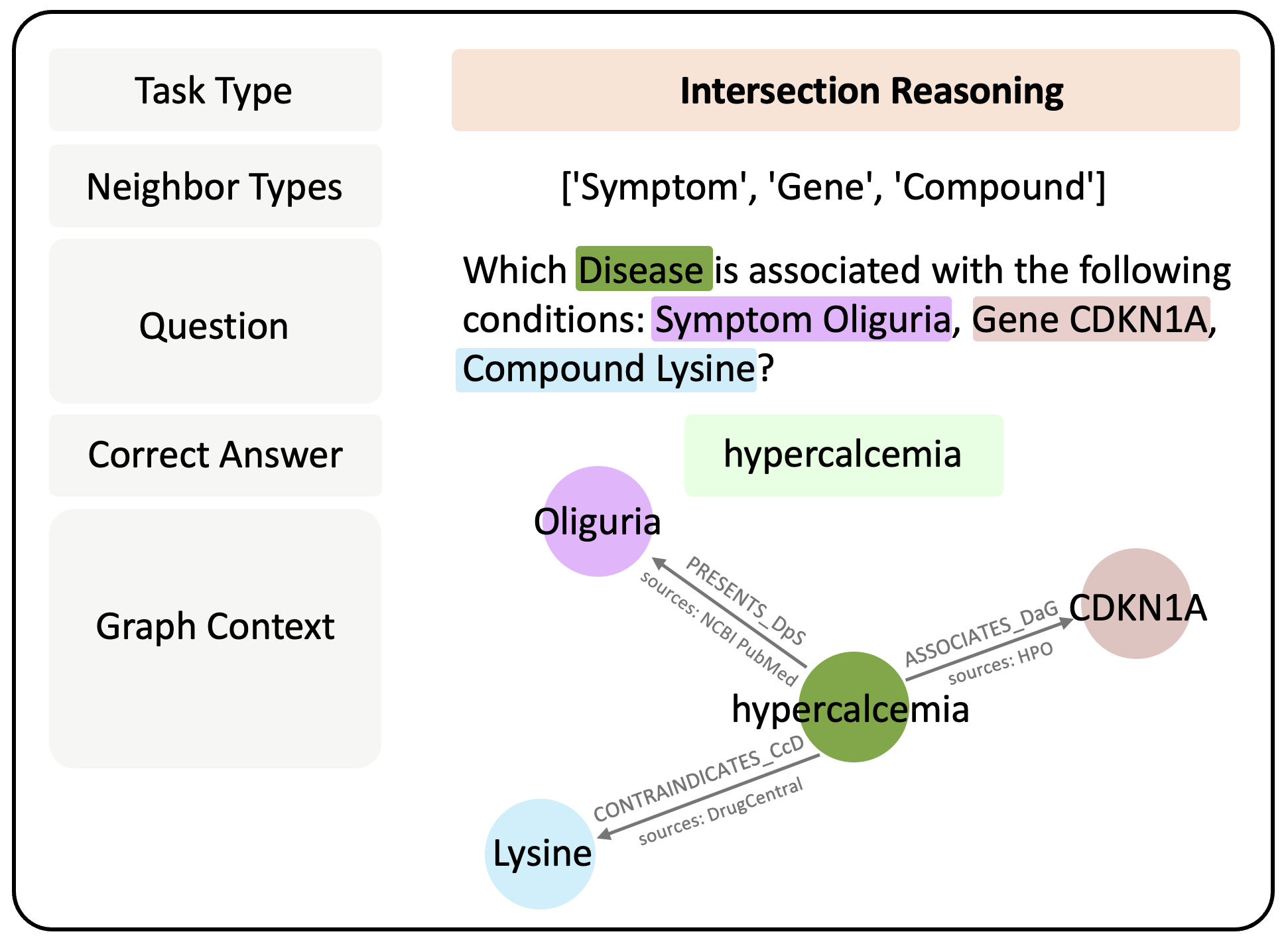}
  \caption{Example from Intersection Reasoning. The target disease must satisfy multiple entity-level clues.}
  \label{fig:intersection-reasoning-example}
\end{figure}

\myparagraphunderline{Example}
Figure~\ref{fig:intersection-reasoning-example} illustrates an Intersection question where several biomedical clues jointly constrain the target disease. This setting tests whether a model can aggregate multiple entity-level conditions and recover the disease satisfying all of them.

\smallskip
\noindent\raisebox{0.2ex}{\scriptsize$\bullet$}\enspace
\myparagraph{Path-based Reasoning}
Path-based Reasoning evaluates typed metapath traversal. The disease is hidden as an internal node, while the question exposes endpoint entities and natural-language relation descriptions. The model must preserve relation direction and follow the typed biomedical path to recover the missing disease. We instantiate ten five-node metapaths from SPOKE, such as \textit{Anatomy $\rightarrow$ Gene $\rightarrow$ Disease $\rightarrow$ Compound $\rightarrow$ Blend}. The complete template list is provided in Table~\ref{tab:path-templates}.

\begin{figure}[h]
  \centering
  \includegraphics[width=\columnwidth]{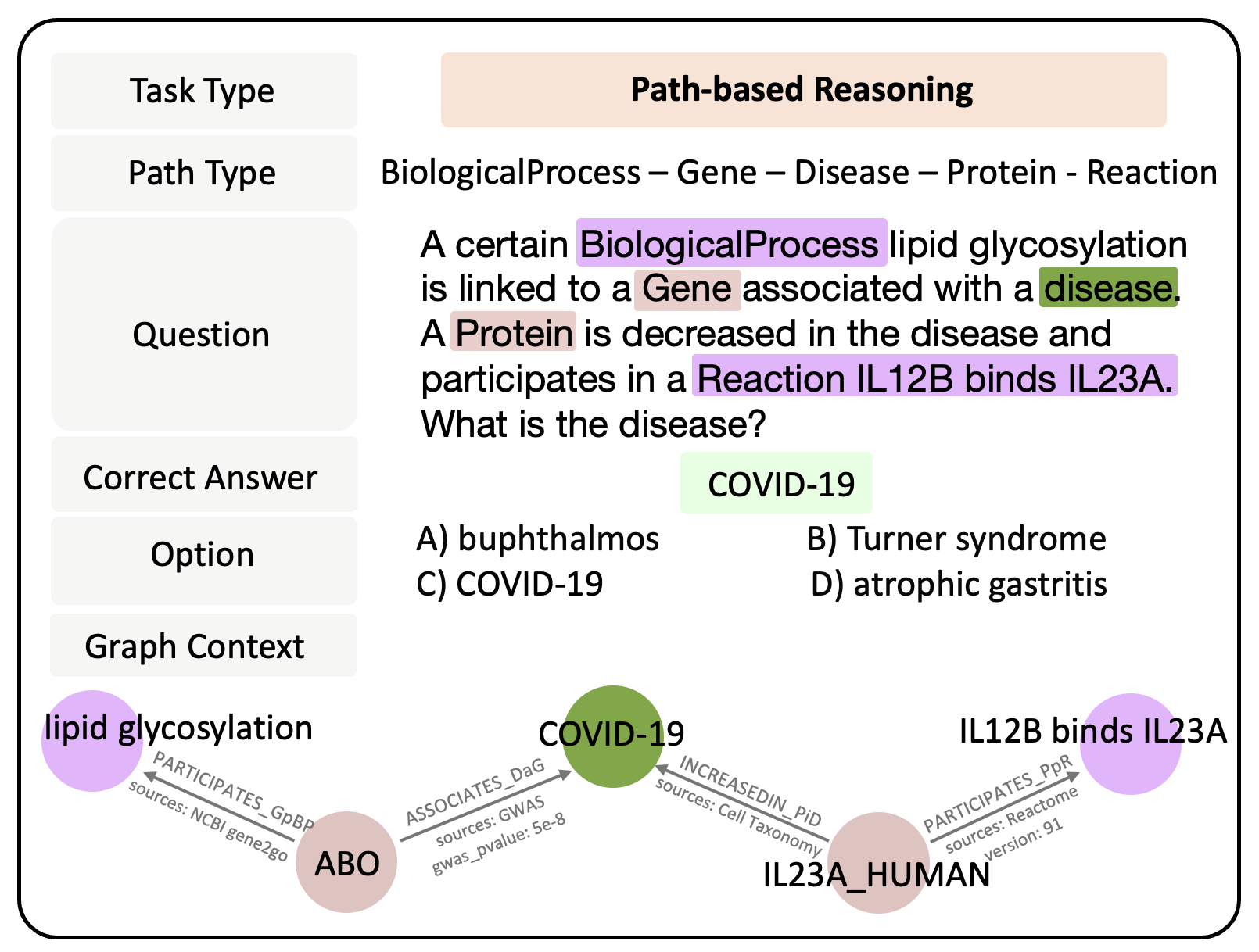}
  \caption{Example from Path-based Reasoning. The model must infer the disease by connecting multiple typed biomedical relations.}
  \label{fig:path-reasoning-example}
\end{figure}

\myparagraphunderline{Example}
Figure~\ref{fig:path-reasoning-example} presents a Path-based Reasoning question built from a biomedical metapath. The answer is a hidden disease node, and solving the question requires following typed relations across heterogeneous entity types.

\newpage
\smallskip
\noindent\raisebox{0.2ex}{\scriptsize$\bullet$}\enspace
\myparagraph{Path-based Counting}
Path-based Counting uses the same typed metapath family as Path-based Reasoning, but changes the output from entity identification to set-valued aggregation. Instead of naming one disease, the model must count all distinct diseases satisfying the relation pattern. This task is harder than single-answer path reasoning because the model must recover the full answer set, merge aliases, remove duplicates, and return the exact number.

\begin{figure}[h]
  \centering
  \includegraphics[width=\columnwidth]{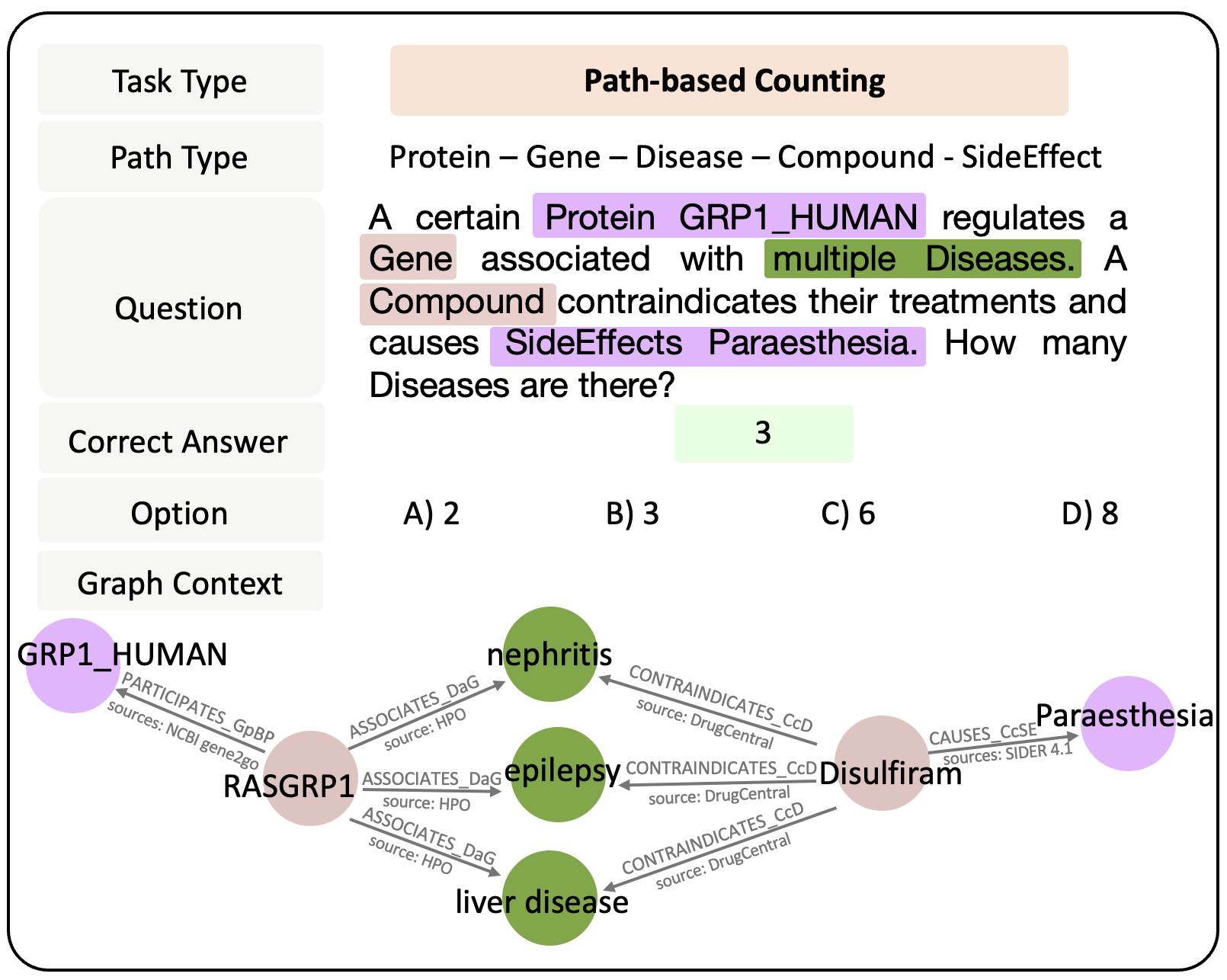}
  \caption{Example from Path-based Counting. The model must aggregate all valid disease nodes satisfying the multi-hop relation pattern.}
  \label{fig:path-counting-example}
\end{figure}

\myparagraphunderline{Example}
Figure~\ref{fig:path-counting-example} shows a set-valued variant of path reasoning. The model must identify all valid disease nodes satisfying the relation pattern and return the number of distinct candidates after entity-level deduplication.

\myparagraph{Dataset statistics}
\label{appendix:benchmark-statistics}
After source expansion, the above graph motifs form the 10,045 source-conditioned instances used in our experiments. 
Table~\ref{tab:dataset-statistics} reports the task composition by construction family. 
The main text reports the answer-centered evidence-graph distribution in Table~\ref{tab:structure-distribution}; here we provide the underlying task counts and construction details.


\myparagraph{Path-based reasoning templates}
\label{appendix:path-templates}
Table~\ref{tab:path-templates} lists the typed metapaths used for deep path construction before document or web evidence is attached. 
Each template specifies an entity-type chain and its natural-language rendering. 
For Path-based Reasoning, the hidden answer is the disease node inside the path. 
For Path-based Counting, the answer is the number of distinct diseases satisfying the same relation pattern.
\begin{table*}[h]
\centering
\renewcommand{\arraystretch}{1.35}
\setlength{\tabcolsep}{5pt}
\caption{Path structures and corresponding question templates for Path-based Reasoning and Path-based Counting tasks in \benchmark.}
\label{tab:path-templates}
\begin{tabular}{|c|p{13.8cm}|}
\hline
\textbf{Path ID} & \textbf{Details} \\ \hline
\multirow{2}{*}{1} 
  & \textbf{Path Structure:} Anatomy $\rightarrow$ Gene $\rightarrow$ Disease $\rightarrow$ Compound $\rightarrow$ Blend \\ 
  & \textbf{Question Template:} A certain anatomical entity [Anatomy] upregulates a gene associated with a disease. A compound contraindicates its treatment and contains a blend [Blend]. What is the disease? \\ \hline
\multirow{2}{*}{2} 
  & \textbf{Path Structure:} BiologicalProcess $\rightarrow$ Gene $\rightarrow$ Disease $\rightarrow$ Compound $\rightarrow$ Food \\ 
  & \textbf{Question Template:} A certain biological process [BiologicalProcess] is linked to a gene associated with a disease. A compound contraindicates its treatment and contains a food [Food]. What is the disease? \\ \hline
\multirow{2}{*}{\centering 3} 
  & \textbf{Path Structure:} BiologicalProcess $\rightarrow$ Gene $\rightarrow$ Disease $\rightarrow$ Compound $\rightarrow$ Protein \\ 
  & \textbf{Question Template:} A certain biological process [BiologicalProcess] is linked to a gene associated with a disease. A compound contraindicates its treatment and binds to a protein [Protein]. What is the disease? \\ \hline
\multirow{2}{*}{\centering 4} 
  & \textbf{Path Structure:} BiologicalProcess $\rightarrow$ Gene $\rightarrow$ Disease $\rightarrow$ Protein $\rightarrow$ Reaction \\ 
  & \textbf{Question Template:} A certain biological process [BiologicalProcess] is linked to a gene associated with a disease. A protein [Protein] is decreased in the disease and participates in a reaction [Reaction]. What is the disease? \\ \hline
\multirow{2}{*}{\centering 5} 
  & \textbf{Path Structure:} CellType $\rightarrow$ Gene $\rightarrow$ Disease $\rightarrow$ Compound $\rightarrow$ Food \\ 
  & \textbf{Question Template:} A certain cell type [CellType] expresses a gene associated with a disease. A compound contraindicates its treatment and contains a food [Food]. What is the disease? \\ \hline
\multirow{2}{*}{\centering 6} 
  & \textbf{Path Structure:} CellularComponent $\rightarrow$ Gene $\rightarrow$ Disease $\rightarrow$ Anatomy $\rightarrow$ CellType \\ 
  & \textbf{Question Template:} A certain cellular component [CellularComponent] is linked to a gene associated with a disease. The disease localizes to an anatomical structure [Anatomy], which is part of a cell type [CellType]. What is the disease? \\ \hline
\multirow{2}{*}{\centering 7} 
  & \textbf{Path Structure:} Location $\rightarrow$ Compound $\rightarrow$ Disease $\rightarrow$ Gene $\rightarrow$ MiRNA \\ 
  & \textbf{Question Template:} A certain location [Location] contains a compound [Compound] that contraindicates the treatment of a disease. The disease is associated with a gene targeted by an miRNA [MiRNA]. What is the disease? \\ \hline
\multirow{2}{*}{\centering 8} 
  & \textbf{Path Structure:} MiRNA $\rightarrow$ Gene $\rightarrow$ Disease $\rightarrow$ Protein $\rightarrow$ ProteinDomain \\ 
  & \textbf{Question Template:} An miRNA [MiRNA] targets a gene associated with a disease. A protein [Protein] is decreased in the disease and is part of a protein domain [ProteinDomain]. What is the disease? \\ \hline
\multirow{2}{*}{\centering 9} 
  & \textbf{Path Structure:} Protein $\rightarrow$ Gene $\rightarrow$ Disease $\rightarrow$ Compound $\rightarrow$ ProteinDomain \\ 
  & \textbf{Question Template:} A protein [Protein] regulates a gene associated with a disease. A compound contraindicates its treatment and binds to a protein domain [ProteinDomain]. What is the disease? \\ \hline
\multirow{2}{*}{\centering 10} 
  & \textbf{Path Structure:} SideEffect $\rightarrow$ Compound $\rightarrow$ Disease $\rightarrow$ Anatomy $\rightarrow$ CellType \\ 
  & \textbf{Question Template:} A side effect [SideEffect] is caused by a compound [Compound] that contraindicates the treatment of a disease. The disease localizes to an anatomical structure [Anatomy], which is part of a cell type [CellType]. What is the disease? \\ \hline
\end{tabular}
\end{table*}

\FloatBarrier
\begin{table*}[t]
\centering
\renewcommand{\arraystretch}{1.08}
\caption{Task counts in the source-conditioned split of \benchmark.}
\label{tab:dataset-statistics}
\resizebox{0.8\textwidth}{!}{%
\begin{tabular}{llcl}
\hline
\textbf{Category} & \textbf{Task Type} & \textbf{Count} & \textbf{Details} \\\hline
\multirow{2}{*}{Answer-centered graph}
& Entity Pair Matching & 2,142 & 9 entity pair types \\
& Intersection Reasoning & 3,682 & 3-clue combinations over 14 neighbor-type patterns \\ \hline
\multirow{2}{*}{Deep path reasoning}
& Path-based Reasoning & 2,653 & 10 predefined metapaths \\
& Path-based Counting & 1,568 & multi-hop set aggregation \\ \cline{2-4}
& \textbf{Total} & \textbf{10,045} & \\ \hline
\end{tabular}
}
\end{table*}

\subsection{Source-conditioned Evidence Construction}
\label{appendix:benchmark-source-evidence}

The source-conditioned evidence layer is attached after the answer graph is fixed.
For each question, the KG path or answer-centered neighborhood defines the gold topology; source conditioning changes only which evidence channels are exposed to the model.
Following the source space in Section~\ref{sec:prelim}, \benchmark{} supports KG, document, web, and hybrid evidence conditions.
This design intentionally trades some ecological breadth for controllability: the latent topology fixes the answer and reasoning operator, while source-conditioned renderings test whether systems can recover the same target from structured, textual, or hybrid evidence.

\myparagraph{KG evidence}
KG evidence provides the structured answer topology.
It serializes answer-centered neighborhoods and typed metapaths from Monarch/BioLink and SPOKE.
For Entity Pair Matching and Intersection Reasoning, the evidence contains the anchor entities, the candidate answer node, and the typed edges connecting them.
For Path-based Reasoning and Path-based Counting, it contains the ordered metapath and intermediate entities.
This structured view serves as the reference topology for aligning document and web witnesses.

\myparagraph{Document evidence}
Document evidence provides PubMed-centered multi-document witnesses for the same topology.
We retrieve literature snippets using \WMTool{BuildHopQueries} and \WMTool{RetrieveMultiDoc}.
For a KG path $e_1 \stackrel{r_1}{\rightarrow} e_2 \cdots \stackrel{r_k}{\rightarrow} e_{k+1}$, \benchmark builds hop-level queries $(e_i,r_i,e_{i+1})$ and partial-chain queries $(e_i,r_i,\ldots,e_j)$.
Entity queries include normalized names, aliases, database identifiers, and ontology-specific labels, while relation queries include biomedical predicates such as \textit{associates with}, \textit{contraindicates}, \textit{localizes to}, \textit{expresses}, \textit{binds}, and \textit{causes}.
The design encourages evidence to be distributed across documents: one paper may support a gene--disease association, another may support a drug--target relation, and another may provide phenotype or anatomical context.

\myparagraph{Web evidence}
Web evidence provides source-conditioned witnesses for missing-hop repair and freshness-sensitive information.
It is included only when the source condition enables Web.
We first prioritize relevant clinical-trial records and then use targeted Google Search API results through \AgentTool{SearchWeb} when additional coverage is needed.
All web snippets are normalized back to graph entities and relation phrases before evaluation, so they act as external witnesses rather than unconstrained context.
For reproducibility, each web evidence unit stores the search query, retrieval timestamp, URL or record identifier, title/snippet, source type, and selected evidence span.
Experiments use the recorded evidence bundle rather than unlogged live search during scoring.

\myparagraph{Entity grounding and metadata}
Each evidence unit stores provenance and grounding metadata.
Snippets are linked back to KG entities through exact names, aliases, identifiers, and type-compatible mentions.
The evidence layer stores both the textual mention and the normalized entity, since biomedical concepts may appear as abbreviations, MONDO/UMLS labels, database identifiers, or clinical synonyms.
It also records source type, retrieval query, supporting hop, linked entities, relation phrases, source prior, and provenance score when available.
These fields support KG-only, document-only, web-only, and hybrid evaluation, and enable verifier checks that separate answer correctness from evidence support.

\subsection{Benchmark Metrics}
\label{appendix:benchmark-metrics}

This subsection defines the metrics used throughout the experiments. 
Automatic metrics are the primary evaluation signal for format-specific answer correctness. 
LLM-verifier metrics are reported separately as grounding diagnostics for open-ended and evidence-grounded cases. 
Because \benchmark{} contains MCQ, open-answer, and counting formats, we report metrics by format rather than using a single unqualified answer accuracy.

\myparagraph{Automatic answer metrics}
\textbf{MCQ Acc.} is option exact-match accuracy on multiple-choice questions. 
\textbf{Open Acc.} is alias-normalized accuracy on open-ended entity questions. 
We report Open Acc. separately for Entity Pair Matching, Intersection Reasoning, and Path-based Reasoning because these task families stress different grounding and generation behavior. 
\textbf{Count Exact} (CountEx) is used only for Path-based Counting. 
CountEx checks whether the extracted number exactly matches the gold count. 
Before CountEx is computed, predicted entity lists are canonicalized with the same alias and identifier normalization used for Open Acc., and duplicate aliases of the same biomedical entity are collapsed.
For example, if the gold set is \{BRCA1, BRCA2\}, an output that repeats BRCA1 and omits BRCA2 is not counted as exact, while accepted aliases for both entities are counted as exact.
\textbf{Avg.} in compact tables is the macro average of MCQ Acc. and Open Acc. within the same entity-answer task family; it is not an aggregate over all task families.
Following prior KG-augmented QA evaluations~\cite{tog1.0sun2023think,tog2.0ma2024think,plan-on-graph}, we do not use Recall or F1 as primary metrics because \benchmark{} is not restricted to a fixed document database; its evidence can come from KG paths, PubMed-centered documents, web results, or their source-conditioned combinations.





\myparagraph{Verifier-based grounding metrics}
For source-conditioned evaluation, each verifier call returns three grounding dimensions together with an error type and a short reason.
We do not merge these dimensions into a single grounded-accuracy score.
Instead, we report them separately as verifier-based diagnostics, since semantic answer correctness, evidence support, and reasoning completeness reflect different failure modes.

\myparagraphunderline{Semantic answer correctness}
This metric (\textbf{Ans.}) checks whether the predicted biomedical answer is semantically equivalent to the gold answer.
For open-ended questions, the verifier accepts aliases, abbreviations, database identifiers, and spelling variants when they clearly refer to the same entity.

\myparagraphunderline{Evidence support}
This metric  (\textbf{Evid.}) measures whether supplied KG paths, document snippets, or web evidence support the predicted answer.
A binary evidence-supported label is derived by thresholding this score, allowing evaluation to separate answer plausibility from explicit evidence support.

\myparagraphunderline{Reasoning completeness}
This metric (\textbf{Reason.}) checks whether supplied trace or selected evidence covers required hops, entities, or counting set.
This dimension is especially important for multi-hop and counting cases, where a correct final answer may still omit an intermediate shared neighbor, duplicate an entity, or miss an entity in the counted set.

\myparagraph{Diagnostic use}
The verifier metrics distinguish answer equivalence, evidence support, and reasoning coverage. 
They do not prove that evidence causally produced the answer; causal evidence use is instead tested through source ablations and evidence-removal controls. 
We additionally use cross-judge consistency and \texttt{error\_type} to diagnose failures from entity grounding, missing evidence, wrong hops, distractor confusion, counting errors, unsupported generation, formatting, or judge instability. 
The complete verifier prompt is provided in Appendix~\ref{prompt:llm-verifier}.

\section{Experiment Details}
\label{sec:experiment-details}
\phantomsection
\myparagraph{Experiment datasets}
\label{appendix:evaluation-setup}
To evaluate whether biomedical reasoning systems can solve source-conditioned biomedical reasoning, we evaluate all methods on \benchmark.
\benchmark contains 10,045 QA instances generated from latent biomedical graph motifs and rendered under KG, document, web, and hybrid evidence conditions.
The four task families are Entity Pair Matching, Intersection Reasoning, Path-based Reasoning, and Path-based Counting.
For the first three task families, we evaluate both multiple-choice and open-answer formats; for Path-based Counting, we evaluate exact numeric counting.
The source condition changes the exposed evidence channel, but does not change the gold answer.
This allows us to compare whether a method can recover the same biomedical target when the evidence is structured, unstructured, or distributed across heterogeneous sources.

Unless a table explicitly sweeps multiple backbones, the main experiments use GPT-4-Turbo as the answer-generation backbone, matching the main automatic evaluation in Table~\ref{tab:main-results}.
The LLM-verifier experiments in Table~\ref{tab:main-llm-verifier} evaluate the same task families with verifier-based grounding dimensions.
The source-conditioned and component ablations in Appendix~\ref{appendix:additional-ablation-analysis} use the same task taxonomy and report format-specific automatic metrics together with verifier diagnostics.
We use stratified evaluation subsets by task family, question format, source condition, and evidence-graph structure.
If a source-specific bucket is smaller than the requested evaluation budget, we use the full available bucket and record the cap in the split manifest.
All methods are evaluated on the same question IDs and scoring scripts.
For each source condition, a method only receives the evidence channels allowed by its retrieval design and by the benchmark split; retrieval caps, output schemas, and parsing rules are fixed before evaluation.

\vspace{4mm}
\phantomsection
\myparagraph{Experiment baselines}
\label{appendix:compared-methods}
We compare \bianque with four categories of baselines under an unsupervised setting with GPT-4-Turbo as the LLM:
\begin{itemize}
    \item LLM-only methods, including input-output prompting (IO) and Chain-of-Thought prompting (CoT)~\cite{wei2022cot};
    \item Vanilla retrieval methods, including Naive Doc and Naive Web, which expose only document or web evidence to a retrieval-augmented prompt~\cite{lewis2020rag};
    \item KG-based RAG methods, including Think-on-Graph (ToG)~\cite{tog1.0sun2023think} and Plan-on-Graph (PoG)~\cite{plan-on-graph};
    \item Hybrid RAG methods, including Chain-of-Knowledge (CoK)~\cite{li2023chaincok} and Think-on-Graph-2.0 (ToG-2)~\cite{tog2.0ma2024think}.
\end{itemize}

\noindent IO and CoT receive only the question, options when available, and the required answer schema.
Naive Doc and Naive Web test whether a single unstructured evidence channel is sufficient.
ToG and PoG test structured KG-path reasoning without document/web witnesses.
CoK and ToG-2 test prior hybrid RAG designs that combine structured and textual evidence, but do not use \bianque's unified evidence graph conversion, entity normalization, and verification-aware pruning.
The proposed \bianque system retrieves source-conditioned KG paths, PubMed-centered document evidence, and optional web evidence, converts them into a unified evidence graph, and answers from the verified evidence topology.
This keeps the answer-generation backbone, test split, and evaluation protocol constant while varying the retrieval and reasoning strategy.

\vspace{2mm}
\phantomsection
\myparagraph{Backbone models}
\label{appendix:backbone-models}
The main method comparison uses GPT-4-Turbo to isolate differences among retrieval and reasoning frameworks.
To test whether \benchmark remains diagnostic across different model families, we additionally run a backbone sweep over nine LLMs: Qwen3-4B, GPT-4o-mini, Claude-3.5-Haiku, Qwen3-80B (Qwen/Qwen3-Next-80B-A3B-Instruct), MiMo-V2.5-Pro, DeepSeek-V4, Gemini-2.5-Pro, Claude-3.7-Sonnet, and GPT-4-Turbo.
In this sweep, each backbone is evaluated with IO, PoG, ToG-2, and \bianque, and Figure~\ref{fig:backbone-robustness} reports Overall Avg. while Table~\ref{tab:full-backbone-sweep} reports the full task-family breakdown.
The backbone sweep replaces only the answer-generation model; task splits, evidence conditions, prompts, parsing, and scoring remain fixed.

\vspace{4mm}
\phantomsection
\myparagraph{Experiment implementation}
\label{appendix:implementation-details}
All prediction records are stored in a unified JSON-style schema with \texttt{answer}, \texttt{supporting\_sources}, \texttt{evidence\_ids}, and \texttt{brief\_reason}; evidence fields may be empty for LLM-only baselines.
Answer generation uses temperature 0 unless otherwise stated.
For relevance scoring in retrieval and pruning, we use SentenceBERT~\cite{reimers-gurevych-2019-sentence} as the embedding model for $\mathbf{h}(\cdot)$ in Section~\ref{sec:method:pruning}.
Unless otherwise specified, BioWeave uses $D_{\max}=3$, a raw KG expansion cap of 200 paths, $W_{doc}=15$, $W_{web}=10$, and source-wise embedding pruning to 10 units per source.
The multi-source candidate pool is capped at $W_1=40$, the cross-score pool at $W_2=15$, and the final LLM-aware selector keeps $K=5$ evidence units.
For evidence pruning, we set $\lambda_{\mathrm{sem}}=0.7$, $\lambda_{\mathrm{ent}}=0.3$, $\beta=0.7$, $\gamma=0.75$, $W_{\mathrm{src}}=3$, $(\rho_{\mathrm{KG}},\rho_{\mathrm{Doc}},\rho_{\mathrm{Web}})=(1.0,0.8,0.7)$, and $\alpha_{\mathrm{prior}},\alpha_{\mathrm{sup}},\alpha_{\mathrm{kg}}=0.33$.
These hyperparameters are fixed across task families, source conditions, LLM backbones, and baselines; they are not tuned separately for any method or task.
Prediction outputs use a maximum of 1024 tokens, while verifier outputs use a maximum of 256 tokens.
For MCQ questions, we parse the selected option and compute exact-match accuracy.
For open-answer questions, we normalize biomedical names, aliases, and identifiers before computing Open Acc.
For counting questions, we extract the numeric answer and report CountEx.
Overall Avg. averages accuracy-style metrics across task families.
Following prior KG-augmented QA evaluations~\cite{tog1.0sun2023think,tog2.0ma2024think,plan-on-graph}, we do not use Recall or F1 as primary scores because the available knowledge sources are not limited to a fixed document database; \benchmark{} exposes KG paths, PubMed-centered documents, and source-conditioned web evidence.

Verifier-based metrics are used as diagnostics rather than replacements for automatic metrics.
The default LLM verifier is GPT-4o-mini.
It returns semantic answer correctness, evidence support, reasoning completeness, \texttt{error\_type}, and \texttt{short\_reason}; these correspond to Ans., Evid., and Reason. in Table~\ref{tab:main-llm-verifier} and the additional experiments.
Ans. and Reason. are percentages of cases judged correct or complete; Evid. is the percentage of cases with \texttt{evidence\_support\_score}$\geq 3$ on the 0--5 verifier scale.
To check judge stability, Appendix~\ref{appendix:judge-reliability} compares GPT-4o-mini with Qwen3-80B and DeepSeek-V4 on 500 sampled predictions.
All benchmark construction, source planning, retrieval query construction, evidence graph conversion, source repair, answer generation, and verifier prompt templates are provided in Appendix~\ref{appendix:prompts}.


\section{Detailed Related Work}
\label{sec:detailed-related-work}
\phantomsection\label{relwork:rag}
\myparagraph{Retrieval-augmented LLMs}
Recent studies have highlighted the reasoning abilities of LLMs through prompting techniques.
Although LLMs show strong language understanding and generation capability, they can produce fluent but unsupported statements when the answer requires external, up-to-date, or domain-specific evidence.
Retrieval-augmented generation addresses this limitation by coupling parametric language models with non-parametric memory, allowing the model to condition generation on retrieved evidence~\cite{lewis2020rag,gao2023retrieval}.
In biomedical settings, retrieval sources may include scientific literature, clinical knowledge bases, ontologies, or biomedical KGs.
However, standard text-based RAG generally treats retrieval as passage selection.
Iterative retrieval-generation methods add context over multiple rounds~\cite{shao2023enhancing}, and prompting methods such as chain-of-thought and ReAct encourage models to expose intermediate reasoning or retrieval actions~\cite{wei2022cot,yao2022react}.
These methods improve the interaction between retrieval and generation, but they still largely depend on semantic relevance and LLM-side reconciliation of retrieved evidence.
This is insufficient when the question requires entity normalization, relation typing, provenance checking, and reasoning over multiple interacting biomedical entities.
\bianque is motivated by this gap: instead of appending retrieved passages alone, it aligns document evidence with structured biomedical relations and verifies the final answer against cross-source evidence.

\phantomsection\label{relwork:kg-rag}
\myparagraph{KG-based RAG}
KG-based RAG methods use structured triples or paths to reduce the ambiguity of pure text retrieval.
Think-on-Graph asks an LLM to walk through KG neighbors step by step~\cite{tog1.0sun2023think}, while StructGPT reformulates reasoning over structured data as repeated read-and-reason operations~\cite{jiang2023structgpt}.
Planning and debate variants further use LLM calls to rank, correct, or verify graph traversal decisions~\cite{plan-on-graph,debated-on-graph}.
Paths-over-Graph focuses on retrieving explicit multi-hop KG paths, showing that path-structured evidence can improve LLM reasoning when the KG contains the required relations~\cite{pogtan2025paths}.
However, KG-only reasoning inherits the incompleteness, freshness, and ontology-boundary limitations of the underlying graph.

\phantomsection\label{relwork:hybrid-rag}
\myparagraph{Hybrid RAG}
Hybrid RAG methods aim to combine structured and unstructured knowledge.
Chain-of-Knowledge dynamically adapts over heterogeneous sources~\cite{li2023chaincok}, GraphRAG constructs document-level graph structure to support query-focused summarization~\cite{graphragmicrosoft}, and HybridRAG combines KG and vector retrieval for information extraction~\cite{sarmah2024hybridrag}.
Think-on-Graph 2.0 further uses KG-guided retrieval to support interpretable reasoning over multiple sources~\cite{tog2.0ma2024think}.
These methods motivate our hybrid evidence setting, but they are usually designed as reasoning systems rather than benchmarks and often leave source reconciliation to the generator.
Verifier-first and reflection-centric RAG systems are complementary to this direction: they decide whether retrieved evidence is sufficient, whereas \bianque further asks whether evidence units can be normalized to the same typed biomedical topology with provenance and source priors.
\benchmark and \bianque instead make source conditions explicit: the benchmark tests whether KG, document, web, and hybrid evidence are used correctly, while the method aligns heterogeneous biomedical evidence into a shared reasoning graph before answer generation.

\phantomsection\label{relwork:biomedical-multihop-benchmarks}
\myparagraph{Biomedical Multi-hop Benchmarks}
BioHopR and MedHopQA are the closest recent efforts to \benchmark.
BioHopR evaluates multi-hop, multi-answer reasoning over PrimeKG, focusing on one-to-many and many-to-many biomedical relations among drugs, diseases, proteins, and phenotypes~\cite{kim2025biohopr,chandak2023primekg}.
Its formulation highlights the importance of bridge entities: even when a final answer is plausible, the model must identify the intermediate entity that connects the query and target nodes.
However, BioHopR is restricted to one KG source and relatively short reasoning chains, which may underrepresent heterogeneous biomedical evidence and broader multi-entity reasoning scenarios.
MedHopQA instead evaluates disease-centered multi-hop reasoning across two Wikipedia articles and uses open-ended answers with ontology-grounded synonym sets from MONDO, NCBI Gene, and NCBI Taxonomy for lexical and concept-level evaluation~\cite{islamaj2026medhopqa}.
This design is valuable because it tests cross-document synthesis and reduces reliance on answer-option elimination.
At the same time, MedHopQA is document-centered and does not explicitly evaluate KG-grounded multi-entity reasoning over structured biomedical relations.
\benchmark combines these two directions: it retains KG-grounded complex reasoning, supports multiple-choice and open-ended forms, and is designed to measure KG-only, document-only, hybrid graph-document, and multi-source reasoning systems.

\phantomsection\label{relwork:biomedical-kg-reasoning}
\myparagraph{Biomedical Knowledge Graph Reasoning}
Biomedical KGs provide structured representations for heterogeneous biomedical relations.
Hetionet integrates compounds, diseases, genes, symptoms, side effects, pathways, biological processes, and anatomical entities for drug repurposing~\cite{himmelstein2017hetionet}, while PrimeKG extends precision-medicine-oriented graph construction over diverse biomedical concepts~\cite{chandak2023primekg}.
Resources such as Monarch, SPOKE, UMLS, HPO, and DisGeNET further illustrate the need to normalize entities, preserve provenance, and integrate phenotype, genotype, disease, pharmacological, and clinical evidence~\cite{mungall2017monarch,morris2023spoke,bodenreider2004umls,robinson2008hpo,pinero2015disgenet}.
Prior KGQA work shows that structured graphs are useful for interpretable reasoning, but biomedical graph reasoning remains difficult because entity aliases are noisy, path spaces are large, and many valid answers may share intermediate nodes.
Our work builds on this line by pairing KG retrieval with multi-document evidence and unified evidence graph conversion, enabling cross-source verification rather than relying on a single knowledge source.

\newpage
\phantomsection\label{relwork:biomedical-qa}
\myparagraph{Biomedical question answering}
Biomedical QA is essential for knowledge acquisition in the biomedical field. 
LLMs have been successfully applied to various biomedical QA tasks that are knowledge-intensive \cite{singhal2023large, lievin2024can, thirunavukarasu2023large}.
Existing benchmarks such as PubMedQA, BioASQ-QA, MedQA, and MedMCQA evaluate biomedical literature comprehension or medical exam knowledge~\cite{jin2019pubmedqa,jin2021disease,pal2022medmcqa}.
These datasets have driven progress, but many questions can be treated as single-hop comprehension, answer classification, or multiple-choice selection.
Recent work argues that multiple-choice formats may allow answer elimination rather than genuine inference, and that widely circulated exam-style benchmarks are increasingly vulnerable to saturation and contamination~\cite{islamaj2026medhopqa}.
MedExQA further introduces explanation-oriented evaluation, but it is still centered on answer justification rather than systematic multi-hop, multi-entity biomedical reasoning~\cite{kim2024medexqa}.
Our benchmark complements these efforts by explicitly testing complex biomedical reasoning across interacting entities, heterogeneous evidence, and open-ended answer generation.

\clearpage
\onecolumn
\section{Case Study: Multi-Source Cross-Verified Interpretable Reasoning}
\label{appendix:case-study}
\definecolor{kgblue}{RGB}{30,90,180}
\definecolor{docgreen}{RGB}{0,140,60}
\definecolor{weborange}{RGB}{210,110,20}

\newcommand{\KGCase}[1]{\textcolor{kgblue}{#1}}
\newcommand{\DocCase}[1]{\textcolor{docgreen}{#1}}
\newcommand{\WebCase}[1]{\textcolor{weborange}{#1}}
\newcommand{\CaseRel}[1]{\xrightarrow{\text{\scriptsize #1}}}
\newcommand{\CaseBackRel}[1]{\xleftarrow{\text{\scriptsize #1}}}

This section presents representative qualitative case studies to illustrate how
\bianque constructs multi-source evidence topologies and verifies answers across
heterogeneous sources. These cases are intended to visualize retrieval,
evidence graph conversion, pruning, and verification; they are not additional
aggregate evaluation results. Paths from different sources are color-coded:
\KGCase{KG}, \DocCase{Doc}, and \WebCase{Web}. Text-derived paths are treated as
candidate graph-compatible evidence units rather than newly asserted KG facts.

\begin{table}[!htbp]
\small
\centering
\caption{Multi-source interpretable reasoning for ``Which gene is associated with both psoriasis and Takayasu arteritis?''. Paths from KG and Doc are color-coded: \KGCase{KG}, \DocCase{Doc}.}
\label{tab:case-study-kg-doc-hlab}
\renewcommand{\arraystretch}{1.08}
\begin{tabularx}{0.98\textwidth}{@{}p{3.0cm}X@{}}
\toprule
\textbf{Field} & \textbf{Content} \\
\midrule
\textbf{Question} & Which gene is associated with both psoriasis and Takayasu arteritis? \\
\textbf{Answer} & HLA-B \\
\textbf{Source Condition} & KG+Doc \\
\textbf{Topic Entities} & \{psoriasis, Takayasu arteritis\} \\
\midrule
\textbf{Reasoning Operator} & Shared-node matching over two disease--gene associations. \\
\textbf{LLM Indicator} & ``psoriasis'' -- associated gene -- answer(gene) -- associated disease -- ``Takayasu arteritis'' \\
\textbf{Split Questions} &
split\_question 1: Which genes are associated with psoriasis? \newline
split\_question 2: Which of these genes are also associated with Takayasu arteritis? \\
\midrule
\textbf{Source Evidence} &
\textbf{\KGCase{KG Path:}} \newline
\KGCase{$\{\mathrm{psoriasis}\} \CaseRel{associates with}
\{\mathbf{HLA\mbox{-}B}\} \CaseBackRel{associates with}
\{\mathrm{Takayasu~arteritis}\}$} \newline
\textbf{\DocCase{Doc Witness:}} \newline
\textit{PubMed literature reports \DocCase{\textbf{HLA-B haplotypes}} in
\DocCase{\textbf{psoriasis}} genetic association studies and
\DocCase{\textbf{HLA-B*52 susceptibility}} in
\DocCase{\textbf{Takayasu arteritis}}.} \newline
\textbf{\DocCase{Doc (converted KG-path):}} \newline
\DocCase{$\{\mathrm{psoriasis}\} \CaseRel{HLA haplotype signal}
\{\mathrm{HLA\mbox{-}B}\}$} \newline
\DocCase{$\{\mathrm{Takayasu~arteritis}\} \CaseRel{susceptibility locus}
\{\mathrm{HLA\mbox{-}B*52}\}$} \newline
\textbf{\WebCase{Web Witness:}} Not enabled by the KG+Doc source condition. \\
\textbf{Source Notes} &
\href{https://pubmed.ncbi.nlm.nih.gov/17101473/}{PMID:17101473};
\href{https://pubmed.ncbi.nlm.nih.gov/30498034/}{PMID:30498034}. \\
\midrule
\textbf{\bianque Answer} &
\textbf{answer:} \{\KGCase{\textbf{HLA-B}}\} \newline
\textbf{reason:} The KG path identifies HLA-B as the shared bridge gene between the two diseases. The document-derived paths corroborate both disease--HLA-B links, giving cross-source support for the same answer. \\
\bottomrule
\end{tabularx}
\end{table}

\begin{table}[!htbp]
\small
\centering
\caption{Multi-source interpretable reasoning for ``Which cancer is connected to EGFR-mutant tumors treated with osimertinib in a registered clinical-trial record?''. Paths from KG and Web are color-coded: \KGCase{KG}, \WebCase{Web}.}
\label{tab:case-study-kg-web-egfr}
\renewcommand{\arraystretch}{1.08}
\begin{tabularx}{0.98\textwidth}{@{}p{3.0cm}X@{}}
\toprule
\textbf{Field} & \textbf{Content} \\
\midrule
\textbf{Question} & Which cancer is connected to EGFR-mutant tumors treated with osimertinib in a registered clinical-trial record? \\
\textbf{Answer} & Non-small cell lung cancer \\
\textbf{Source Condition} & KG+Web \\
\textbf{Topic Entities} & \{EGFR, osimertinib\} \\
\midrule
\textbf{Reasoning Operator} & Drug--target--disease linking with online evidence repair. \\
\textbf{LLM Indicator} & ``osimertinib'' -- targets -- ``EGFR'' -- disease context -- answer(cancer) \\
\textbf{Split Questions} &
split\_question 1: What molecular target is linked to osimertinib? \newline
split\_question 2: What cancer context is supported by clinical-trial evidence for EGFR-mutant tumors? \\
\midrule
\textbf{Source Evidence} &
\textbf{\KGCase{KG Path:}} \newline
\KGCase{$\{\mathrm{osimertinib}\} \CaseRel{targets}
\{\mathrm{mutant~EGFR}\} \CaseRel{associated with}
\{\mathbf{non\mbox{-}small~cell~lung~cancer}\}$} \newline
\textbf{\DocCase{Doc Witness:}} Not enabled by the KG+Web source condition. \newline
\textbf{\WebCase{Web Witness:}} \newline
\textit{ClinicalTrials.gov describes a phase-II study of neoadjuvant
\WebCase{\textbf{osimertinib}} for stage I--IIIA
\WebCase{\textbf{EGFR-mutant non-small cell lung cancer}}.} \newline
\textbf{\WebCase{Web (converted KG-path):}} \newline
\WebCase{$\{\mathrm{osimertinib}\} \CaseRel{neoadjuvant therapy for}
\{\mathrm{EGFR\mbox{-}mutant~NSCLC}\}
\CaseRel{disease alias}
\{\mathbf{non\mbox{-}small~cell~lung~cancer}\}$} \\
\textbf{Source Notes} &
\href{https://clinicaltrials.gov/study/NCT03433469}{ClinicalTrials.gov: NCT03433469}. \\
\midrule
\textbf{\bianque Answer} &
\textbf{answer:} \{\KGCase{\textbf{Non-small cell lung cancer}}\} \newline
\textbf{reason:} The KG supplies the target--disease structure, while the clinical-trial record repairs the treatment-specific hop. The web-derived path normalizes NSCLC to non-small cell lung cancer and verifies the same disease answer. \\
\bottomrule
\end{tabularx}
\end{table}

\begin{table}[!htbp]
\small
\centering
\caption{Multi-source interpretable reasoning for ``Which disease is supported by PubMed literature and clinical-trial records as a BRCA-mutated cancer treated with olaparib maintenance therapy?''. Paths from Doc and Web are color-coded: \DocCase{Doc}, \WebCase{Web}.}
\label{tab:case-study-doc-web-olaparib}
\renewcommand{\arraystretch}{1.08}
\begin{tabularx}{0.98\textwidth}{@{}p{3.0cm}X@{}}
\toprule
\textbf{Field} & \textbf{Content} \\
\midrule
\textbf{Question} & Which disease is supported by PubMed literature and clinical-trial records as a BRCA-mutated cancer treated with olaparib maintenance therapy? \\
\textbf{Answer} & Ovarian cancer \\
\textbf{Source Condition} & Doc+Web \\
\textbf{Topic Entities} & \{BRCA1/2 mutation, olaparib\} \\
\midrule
\textbf{Reasoning Operator} & Unstructured source agreement with graph-compatible claim conversion. \\
\textbf{LLM Indicator} & ``BRCA1/2 mutation'' -- molecular subtype -- disease -- maintenance therapy -- ``olaparib'' \\
\textbf{Split Questions} &
split\_question 1: Which disease is described as BRCA-mutated in PubMed literature? \newline
split\_question 2: Which disease setting uses olaparib as maintenance therapy in clinical-trial evidence? \\
\midrule
\textbf{Source Evidence} &
\textbf{\KGCase{KG Path:}} Not exposed to the answerer under the Doc+Web source condition. \newline
\textbf{\DocCase{Doc Witness:}} \newline
\textit{PubMed reports maintenance \DocCase{\textbf{olaparib}} trials for
patients with newly diagnosed advanced \DocCase{\textbf{ovarian cancer}}
carrying \DocCase{\textbf{BRCA mutations}}.} \newline
\textbf{\DocCase{Doc (converted KG-path):}} \newline
\DocCase{$\{\mathrm{BRCA1/2~mutation}\} \CaseRel{molecular subtype of}
\{\mathbf{ovarian~cancer}\} \CaseBackRel{maintenance therapy}
\{\mathrm{olaparib}\}$} \newline
\textbf{\WebCase{Web Witness:}} \newline
\textit{Clinical-trial records describe \WebCase{\textbf{olaparib maintenance}}
in \WebCase{\textbf{BRCA-mutated ovarian cancer}} settings.} \newline
\textbf{\WebCase{Web (converted KG-path):}} \newline
\WebCase{$\{\mathrm{olaparib}\} \CaseRel{maintenance therapy for}
\{\mathrm{BRCA\mbox{-}mutated~ovarian~cancer}\}
\CaseRel{disease}
\{\mathbf{ovarian~cancer}\}$} \\
\textbf{Source Notes} &
\href{https://pubmed.ncbi.nlm.nih.gov/34715071/}{PMID:34715071};
\href{https://clinicaltrials.gov/study/NCT01844986}{ClinicalTrials.gov: NCT01844986}. \\
\midrule
\textbf{\bianque Answer} &
\textbf{answer:} \{\DocCase{\textbf{Ovarian cancer}}\} \newline
\textbf{reason:} Without exposing KG facts under the Doc+Web condition, \bianque converts two unstructured witnesses into graph-compatible paths. Both paths connect BRCA mutation and olaparib maintenance to ovarian cancer. \\
\bottomrule
\end{tabularx}
\end{table}

\begin{table}[!htbp]
\small
\centering
\caption{Multi-source interpretable reasoning for ``Which disease is characterized by the BCR-ABL fusion and has imatinib as a targeted therapy in literature and clinical-trial records?''. Paths from KG, Doc, and Web are color-coded: \KGCase{KG}, \DocCase{Doc}, \WebCase{Web}.}
\label{tab:case-study-all-cml}
\renewcommand{\arraystretch}{1.08}
\begin{tabularx}{0.98\textwidth}{@{}p{3.0cm}X@{}}
\toprule
\textbf{Field} & \textbf{Content} \\
\midrule
\textbf{Question} & Which disease is characterized by the BCR-ABL fusion and has imatinib as a targeted therapy in literature and clinical-trial records? \\
\textbf{Answer} & Chronic myeloid leukemia \\
\textbf{Source Condition} & KG+Doc+Web \\
\textbf{Topic Entities} & \{BCR-ABL, imatinib\} \\
\midrule
\textbf{Reasoning Operator} & Multi-source entity normalization and therapy--disease verification. \\
\textbf{LLM Indicator} & ``imatinib'' -- inhibits -- ``BCR-ABL'' -- molecular marker of -- answer(disease) \\
\textbf{Split Questions} &
split\_question 1: What disease is characterized by the BCR-ABL fusion? \newline
split\_question 2: What disease has imatinib as a targeted therapy in literature and clinical-trial records? \\
\midrule
\textbf{Source Evidence} &
\textbf{\KGCase{KG Path:}} \newline
\KGCase{$\{\mathrm{imatinib}\} \CaseRel{inhibits}
\{\mathrm{BCR\mbox{-}ABL1}\} \CaseRel{associated with}
\{\mathbf{chronic~myeloid~leukemia}\}$} \newline
\textbf{\DocCase{Doc Witness:}} \newline
\textit{Biomedical literature describes \DocCase{\textbf{chronic myeloid leukemia}}
as a \DocCase{\textbf{BCR-ABL-positive}} disease and reports
\DocCase{\textbf{imatinib}} monitoring/treatment in chronic-phase CML.} \newline
\textbf{\DocCase{Doc (converted KG-path):}} \newline
\DocCase{$\{\mathrm{BCR\mbox{-}ABL}\} \CaseRel{molecular marker of}
\{\mathbf{chronic~myeloid~leukemia}\} \CaseBackRel{treated with}
\{\mathrm{imatinib}\}$} \newline
\textbf{\WebCase{Web Witness:}} \newline
\textit{ClinicalTrials.gov records compare \WebCase{\textbf{imatinib-based}}
treatment in \WebCase{\textbf{BCR-ABL-positive chronic myeloid leukemia}}.} \newline
\textbf{\WebCase{Web (converted KG-path):}} \newline
\WebCase{$\{\mathrm{imatinib}\} \CaseRel{trial therapy for}
\{\mathrm{BCR\mbox{-}ABL\mbox{-}positive~CML}\}
\CaseRel{disease alias}
\{\mathbf{chronic~myeloid~leukemia}\}$} \\
\textbf{Source Notes} &
\href{https://pmc.ncbi.nlm.nih.gov/articles/PMC3415739/}{PMC:3415739};
\href{https://clinicaltrials.gov/study/NCT00070499}{ClinicalTrials.gov: NCT00070499}. \\
\midrule
\textbf{\bianque Answer} &
\textbf{answer:} \{\KGCase{\textbf{Chronic myeloid leukemia}}\} \newline
\textbf{reason:} All three sources converge on chronic myeloid leukemia. The KG path gives a direct therapy--fusion--disease trace, while the document and clinical-trial paths normalize CML/chronic myeloid leukemia and BCR-ABL/BCR-ABL1 aliases. \\
\bottomrule
\end{tabularx}
\end{table}

\FloatBarrier

\clearpage
\onecolumn
\section{Prompts}
\label{appendix:prompts}
This section lists the reusable prompt templates used by \benchmark construction and the \bianque reasoning pipeline.
The templates cover benchmark question rewriting, question initialization and source planning, source repair, retrieval query construction, unified evidence graph conversion, evidence graph refinement, LLM-aware evidence selection, source-conditioned answering, verified answer generation, and LLM-based verification.
Curly-braced fields denote runtime inputs filled by the benchmark record, retrieval module, answer generator, or evaluation script.
The \bianque-specific templates include source planning, repair control, evidence-graph refinement, and verified answer generation.

\begin{center}
\begin{minipage}{0.9\columnwidth}
    \vspace{2mm}
    \centering
    \begin{tcolorbox}[title=Benchmark Question Rewriting Prompt Template]
    \label{prompt:question-rewriting}
        \small
You generate biomedical benchmark questions.
Preserve the gold answer, answer type, source condition, and answer options.
The \texttt{generated\_question} field must contain only the question stem; never include the option list because options are stored separately.
Do not leak the gold answer unless it is already one of the options.
Return only JSON.

\vspace{5pt}
\texttt{In-Context Few-shot}
\vspace{5pt}

Task: Rewrite this BioMedHop item into a polished benchmark question for the active evidence source setting.

Source set: \texttt{\{Source\_Set\}}

Required source cue: \texttt{\{Source\_Cue\}}

Question generation source: \texttt{\{Question\_Generation\_Source\}}

Task family: \texttt{\{Task\_Family\}}

Answer type: \texttt{\{Answer\_Type\}}

Reasoning operator: \texttt{\{Reasoning\_Operator\}}

Evidence graph structure: \texttt{\{Hop\_Structure\}}

Original question: \texttt{\{Original\_Question\}}

Source-conditioned question: \texttt{\{Source\_Conditioned\_Question\}}

Options: \texttt{\{Options\}}

Output schema: \texttt{\{generated\_question, question\_type, source\_set, reasoning\_operator, answer\_type\}}

Constraints: keep the same biomedical meaning and same answer; begin with the required evidence cue or a close paraphrase; do not include options inside the question stem; use fluent biomedical wording; return compact JSON only.

A:
    \end{tcolorbox}
    \vspace{2mm}
\end{minipage}
\end{center}

\begin{center}
\begin{minipage}{0.9\columnwidth}
    \vspace{2mm}
    \begin{tcolorbox}[title=Question Initialization and Source Planning Prompt Template]
    \label{prompt:question-initialization}
        \small
You are a biomedical reasoning planner.
Given a BioMedHop question, identify the biomedical mentions, candidate entity types, expected answer type, reasoning operator, and source plan.
Prefer KG evidence for typed biomedical relations, document evidence for distributed literature support, and web evidence for freshness or missing local coverage.
If \texttt{\{Allowed\_Sources\}} is supplied, do not select sources outside it.

\vspace{5pt}
\texttt{In-Context Few-shot}
\vspace{5pt}

Q: \texttt{\{Question\}}

Options: \texttt{\{Options\}}

Allowed sources: \texttt{\{Allowed\_Sources\}}

Task family hint: \texttt{\{Task\_Family\}}

Evidence graph structure hint: \texttt{\{Hop\_Structure\}}

Return JSON with:
\texttt{mentions}, \texttt{answer\_type}, \texttt{reasoning\_operator}, \texttt{source\_plan}, \texttt{missing\_hops}, and \texttt{brief\_reason}.

A:
    \end{tcolorbox}
    \vspace{1mm}
\end{minipage}
\end{center}

\vspace{1mm}
\begin{center}
\begin{minipage}{0.9\columnwidth}
    \vspace{2mm}
    \begin{tcolorbox}[title=\bianque Source Repair Prompt Template]
    \label{prompt:bioweave-source-repair}
        \small
You are the \bianque controller.
Given a question, the active source condition, the current evidence graph, and the failed reasoning check, decide which hop or source should be repaired.
Do not add sources outside the active source condition.
Prefer targeted repair: re-query only the missing entity, relation, or hop instead of restarting retrieval.
If the current evidence is already sufficient, return \texttt{repair\_needed=false}.

\vspace{5pt}
\texttt{In-Context Few-shot}
\vspace{5pt}

Q: \texttt{\{Question\}}

Active source condition: \texttt{\{Source\_Set\}}

Reasoning operator: \texttt{\{Reasoning\_Operator\}}

Expected answer type: \texttt{\{Answer\_Type\}}

Current source plan: \texttt{\{Current\_Source\_Plan\}}

Current evidence graph: \texttt{\{Evidence\_Graph\}}

Failed check: \texttt{\{Failed\_Reasoning\_Check\}}

Candidate missing hops: \texttt{\{Candidate\_Missing\_Hops\}}

Return JSON with \texttt{repair\_needed}, \texttt{missing\_hop}, \texttt{target\_entities}, \texttt{target\_relation}, \texttt{updated\_source\_plan}, \texttt{retrieval\_queries}, and \texttt{brief\_reason}.

A:
    \end{tcolorbox}
    \vspace{1mm}
\end{minipage}
\end{center}

\vspace{1mm}
\begin{center}
\begin{minipage}{0.9\columnwidth}
    \vspace{2mm}
    \begin{tcolorbox}[title=Retrieval Query Construction Prompt Template]
    \label{prompt:retrieval-query-construction}
        \small
You are constructing source-conditioned retrieval queries for BioWeave.
Generate compact document and web queries from normalized biomedical entities, aliases, identifiers, relation phrases, retrieved KG paths, and missing-hop signals.
Document queries should target PubMed-centered evidence for each required hop.
Web queries should be used only when the active source condition permits Web evidence or when missing/fresh evidence must be repaired; prioritize clinical-trial records, drug labels, guidelines, and high-quality biomedical sources before broad web search.
Do not create queries for sources outside the active source condition.

\vspace{5pt}
\texttt{In-Context Few-shot}
\vspace{5pt}

Q: \texttt{\{Question\}}

Active source condition: \texttt{\{Source\_Set\}}

Reasoning operator: \texttt{\{Reasoning\_Operator\}}

Expected answer type: \texttt{\{Answer\_Type\}}

Topic entities and aliases: \texttt{\{Topic\_Entities\}}

Candidate KG paths: \texttt{\{KG\_Paths\}}

Missing or uncertain hops: \texttt{\{Missing\_Hops\}}

Relation phrases: \texttt{\{Relation\_Phrases\}}

Return JSON with \texttt{doc\_queries}, \texttt{web\_queries}, \texttt{target\_entities}, \texttt{target\_relations}, and \texttt{brief\_reason}.

A:
    \end{tcolorbox}
    \vspace{1mm}
\end{minipage}
\end{center}

\vspace{1mm}
\begin{center}
\begin{minipage}{0.9\columnwidth}
    \vspace{2mm}
    \begin{tcolorbox}[title=Unified Evidence Graph Conversion Prompt Template]
    \label{prompt:unified-evidence-graph-conversion}
        \small
You are a biomedical evidence normalizer.
Convert each document or web snippet into graph-compatible evidence units for the unified evidence graph.
Use only claims stated in the snippet.
Normalize biomedical entities when possible using aliases, identifiers, and entity types.
Map relation phrases to concise biomedical predicates such as \textit{associated with}, \textit{treats}, \textit{causes}, \textit{expresses}, \textit{localizes to}, \textit{binds}, \textit{contraindicates}, or \textit{has phenotype}.
If a snippet does not support any hop, mark it as unused.

\vspace{5pt}
\texttt{In-Context Few-shot}
\vspace{5pt}

Q: \texttt{\{Question\}}

Topic entities: \texttt{\{Topic\_Entities\}}

Expected answer type: \texttt{\{Answer\_Type\}}

Reasoning operator: \texttt{\{Reasoning\_Operator\}}

Target hop or missing relation: \texttt{\{Target\_Hop\}}

Snippet source: \texttt{\{Doc\_or\_Web\}}

Snippets: \texttt{\{Snippets\}}

Return JSON with \texttt{evidence\_units}, \texttt{unused\_snippets}, and \texttt{normalization\_notes}.

A:
    \end{tcolorbox}
    \vspace{1mm}
\end{minipage}
\end{center}

\vspace{1mm}
\begin{center}
\begin{minipage}{0.9\columnwidth}
    \vspace{2mm}
    \begin{tcolorbox}[title=\bianque Evidence Graph Refinement Prompt Template]
    \label{prompt:bioweave-graph-refinement}
        \small
You are refining a \bianque source-aware evidence graph.
Merge evidence paths that refer to the same biomedical entity or relation.
Preserve provenance identifiers, source types, normalized entity IDs, aliases, relation labels, and hop-level support.
Remove duplicate paths, unsupported claims, and evidence that violates the active source condition.
For hybrid evidence, explicitly mark which KG paths are supported by compatible document or web paths.

\vspace{5pt}
\texttt{In-Context Few-shot}
\vspace{5pt}

Q: \texttt{\{Question\}}

Active source condition: \texttt{\{Source\_Set\}}

Reasoning operator: \texttt{\{Reasoning\_Operator\}}

Expected answer type: \texttt{\{Answer\_Type\}}

Candidate KG paths: \texttt{\{KG\_Paths\}}

Document-derived evidence units: \texttt{\{Document\_Evidence\_Units\}}

Web-derived evidence units: \texttt{\{Web\_Evidence\_Units\}}

Selected evidence units: \texttt{\{Selected\_Evidence\_Units\}}

Return JSON with \texttt{nodes}, \texttt{edges}, \texttt{merged\_aliases}, \texttt{source\_support}, \texttt{removed\_evidence}, and \texttt{reasoning\_coverage}.

A:
    \end{tcolorbox}
    \vspace{1mm}
\end{minipage}
\end{center}

\vspace{1mm}
\begin{center}
\begin{minipage}{0.9\columnwidth}
    \vspace{2mm}
    \begin{tcolorbox}[title=Source-conditioned Answering Prompt Template]
    \label{prompt:source-conditioned-answering}
        \small
You are answering a BioMedHop biomedical reasoning question.
Use only evidence from the active source set.
Align surface mentions to normalized biomedical entities before answering.
For hybrid evidence, prefer claims supported by both graph structure and textual evidence.
For counting questions, deduplicate aliases before counting.
If evidence is insufficient, give the best supported answer and identify the missing hop or weak evidence channel in \texttt{brief\_reason}.

\vspace{5pt}
\texttt{In-Context Few-shot}
\vspace{5pt}

Active source set: \texttt{\{Source\_Set\}}

Q: \texttt{\{Question\}}

Answer type: \texttt{\{Answer\_Type\}}

Reasoning operator: \texttt{\{Reasoning\_Operator\}}

Evidence graph structure: \texttt{\{Hop\_Structure\}}

Options: \texttt{\{Options\}}

KG-grounded topic entities: \texttt{\{KG\_Topic\_Entities\}}

Relation phrases: \texttt{\{Relation\_Phrases\}}

KG paths: \texttt{\{KG\_Paths\}}

Multi-document evidence: \texttt{\{Document\_Snippets\}}

Hop-aware document queries: \texttt{\{Document\_Queries\}}

Web evidence: \texttt{\{Web\_Snippets\}}

Targeted online search queries: \texttt{\{Web\_Queries\}}

Unified evidence graph: \texttt{\{Evidence\_Graph\}}

Return JSON with keys \texttt{answer}, \texttt{normalized\_answer}, \texttt{supporting\_sources}, \texttt{evidence\_ids}, and \texttt{brief\_reason}.

A:
    \end{tcolorbox}
    \vspace{1mm}
\end{minipage}
\end{center}

\vspace{1mm}
\begin{center}
\begin{minipage}{0.9\columnwidth}
    \vspace{2mm}
    \begin{tcolorbox}[title=\bianque Verified Answer Generation Prompt Template]
    \label{prompt:bioweave-verified-answer}
        \small
You are the final \bianque answer generator.
Answer only from the refined source-aware evidence graph.
Before producing the answer, verify that the selected evidence satisfies the reasoning operator, expected answer type, and active source condition.
For multiple-choice questions, select the option supported by the evidence graph.
For open-ended questions, output the normalized biomedical entity name or identifier.
For counting questions, count distinct normalized answer entities after alias merging.
If the evidence graph does not support a valid answer, return \texttt{needs\_repair=true} and identify the missing hop instead of hallucinating.

\vspace{5pt}
\texttt{In-Context Few-shot}
\vspace{5pt}

Q: \texttt{\{Question\}}

Options: \texttt{\{Options\}}

Active source condition: \texttt{\{Source\_Set\}}

Reasoning operator: \texttt{\{Reasoning\_Operator\}}

Expected answer type: \texttt{\{Answer\_Type\}}

Refined evidence graph: \texttt{\{Refined\_Evidence\_Graph\}}

Required source support: \texttt{\{Required\_Source\_Support\}}

Return JSON with \texttt{answer}, \texttt{normalized\_answer}, \texttt{supporting\_sources}, \texttt{evidence\_ids}, \texttt{needs\_repair}, \texttt{missing\_hop}, and \texttt{brief\_reason}.

A:
    \end{tcolorbox}
    \vspace{1mm}
\end{minipage}
\end{center}

\vspace{1mm}
\begin{center}
\begin{minipage}{0.9\columnwidth}
    \vspace{2mm}
    \begin{tcolorbox}[title=LLM-aware Evidence Selection Prompt Template]
    \label{prompt:path-selection}
        \small
You are a biomedical evidence selector.
Given the question, reasoning operator, active source condition, and candidate evidence units, select the top-$K$ units most likely to support the correct answer.
Do not select an evidence unit only because it shares a surface word with the question.
Prefer units that contain normalized topic entities or aliases, satisfy the required hop or relation pattern, match the expected answer type, are supported across active sources when hybrid evidence is required, and preserve provenance identifiers.

\vspace{5pt}
\texttt{In-Context Few-shot}
\vspace{5pt}

Q: \texttt{\{Question\}}

Reasoning operator: \texttt{\{Reasoning\_Operator\}}

Expected answer type: \texttt{\{Answer\_Type\}}

Active source condition: \texttt{\{Source\_Set\}}

Candidate evidence units with relevance and verification scores: \texttt{\{Candidate\_Evidence\_Units\}}

$K$: \texttt{\{K\}}

Return JSON with \texttt{selected\_evidence}, \texttt{rejected\_evidence}, and \texttt{coverage\_notes}.

A:
    \end{tcolorbox}
    \vspace{1mm}
\end{minipage}
\end{center}

\vspace{1mm}
\begin{center}
\begin{minipage}{0.9\columnwidth}
    \vspace{2mm}
    \begin{tcolorbox}[title=LLM Verifier Prompt Template]
    \label{prompt:llm-verifier}
        \small
You are a strict biomedical QA evaluator.
Judge only from the gold answer, the active source condition, and the supplied evidence.
Return only valid JSON with exactly these keys:
\texttt{answer\_correct}, \texttt{evidence\_support\_score}, \texttt{reasoning\_complete}, \texttt{error\_type}, and \texttt{short\_reason}.

\vspace{5pt}
\texttt{In-Context Few-shot}
\vspace{5pt}

Definitions:
\texttt{answer\_correct} is true if the prediction is semantically equivalent to the gold answer.
Accept aliases, abbreviations, and biomedical identifiers when clearly equivalent.
\texttt{evidence\_support\_score} is an integer from 0--5 measuring how well the supplied evidence supports the predicted answer.
\texttt{reasoning\_complete} is true if the prediction or evidence trace covers the required hops, entities, relation directions, or counted entity set.
\texttt{error\_type} should be one of \texttt{none}, \texttt{entity\_grounding}, \texttt{missing\_evidence}, \texttt{wrong\_hop\_or\_relation}, \texttt{distractor\_confusion}, \texttt{counting\_or\_deduplication}, \texttt{unsupported\_answer}, \texttt{format\_error}, or \texttt{other}.

\vspace{5pt}

Q: \texttt{\{Question\}}

Gold answer: \texttt{\{Gold\_Answer\}}

Prediction: \texttt{\{Prediction\}}

Options: \texttt{\{Options\}}

Question format: \texttt{\{Question\_Format\}}

Task family: \texttt{\{Task\_Family\}}

Evidence graph structure: \texttt{\{Hop\_Structure\}}

Active source condition: \texttt{\{Source\_Set\}}

Automatic exact/alias correctness: \texttt{\{Auto\_Correct\}}

Model-provided support: \texttt{\{Model\_Support\}}

Supplied evidence: \texttt{\{Evidence\}}

A:
    \end{tcolorbox}
    \vspace{2mm}
\end{minipage}
\end{center}

\end{document}